\documentclass[runningheads]{llncs}

\usepackage[mobile]{eccv}

\usepackage{eccvabbrv}

\usepackage{graphicx}
\usepackage{booktabs}

\usepackage[accsupp]{axessibility}  

\usepackage[pagebackref,breaklinks,colorlinks,citecolor=eccvblue]{hyperref}

\usepackage{orcidlink}

\usepackage{enumerate}
\usepackage{enumitem}
\usepackage{bbm}
\usepackage{colortbl}
\usepackage{booktabs}
\usepackage{multirow}

\newcommand{\subpara}[1]{{\textbf{#1.}}}
\newcommand{\para}[1]{\vspace{0.4em} \noindent \textbf{#1.}}
\newcommand{\supp}{Supplementary Material\xspace}
\newcommand{\firstone}[1]{\colorbox{red!15}{#1}}
\newcommand{\secondone}[1]{\colorbox{blue!15}{#1}}

\begin{document}

\title{ScribblePrompt: Fast and Flexible Interactive Segmentation for Any Biomedical Image} 

\titlerunning{ScribblePrompt}

\author{Hallee E. Wong\inst{1,2}\orcidlink{0000-0003-1343-9672} \and
Marianne Rakic\inst{1,2}\orcidlink{0000-0003-2376-9448} \and
John Guttag\inst{1}\orcidlink{0000-0003-0992-0906} \and
{Adrian~V.~Dalca}\inst{1,2,3}\orcidlink{0000-0002-8422-0136}
}

\authorrunning{H.E.~Wong et al.}

\institute{MIT CSAIL, Cambridge, MA, USA 
\and 
Martinos Center, Massachusetts General Hospital, Charlestown, MA, USA
\and 
Harvard Medical School, Boston, MA, USA 
\\
\email{\{hallee,mrakic,guttag,adalca\}@mit.edu}
}
\maketitle

\begin{abstract}
Biomedical image segmentation is a crucial part of both scientific research and clinical care. With enough labelled data, deep learning models can be trained to accurately automate specific biomedical image segmentation tasks. However, manually segmenting images to create training data is highly labor intensive and requires domain expertise. We present \emph{ScribblePrompt}, a flexible neural network based interactive segmentation tool for biomedical imaging that enables human annotators to segment previously unseen structures using scribbles, clicks, and bounding boxes. Through rigorous quantitative experiments, we demonstrate that given comparable amounts of interaction, ScribblePrompt produces more accurate segmentations than previous methods on datasets unseen during training. In a user study with domain experts, ScribblePrompt reduced annotation time by 28\% while improving Dice by 15\% compared to the next best method. ScribblePrompt's success rests on a set of careful design decisions. These include a training strategy that incorporates both a highly diverse set of images and tasks, novel algorithms for simulated user interactions and labels, and a network that enables fast inference. We showcase ScribblePrompt in an interactive demo, provide code, and release a dataset of scribble annotations at \url{https://scribbleprompt.csail.mit.edu}
\end{abstract}

\section{Introduction}

Biomedical image segmentation is an essential step in a wide range of biomedical research and clinical care pipelines. Deep learning has become the predominant method to automate existing segmentation tasks~\cite{UNet, isensee_nnunet_2021, tang_selfsupervised_2022, milletari_v-net_2016}. Biomedical researchers and clinicians often encounter novel segmentation tasks involving either new regions of interest \cite{zhang_covid_2022} or new image modalities \cite{sati_new_flair_2012, men_oct_2015}. Unfortunately, supervised training of accurate models for new domains requires diverse images with careful annotations by skilled experts.  

\begin{figure}
    \centering
    \includegraphics[width=\linewidth]{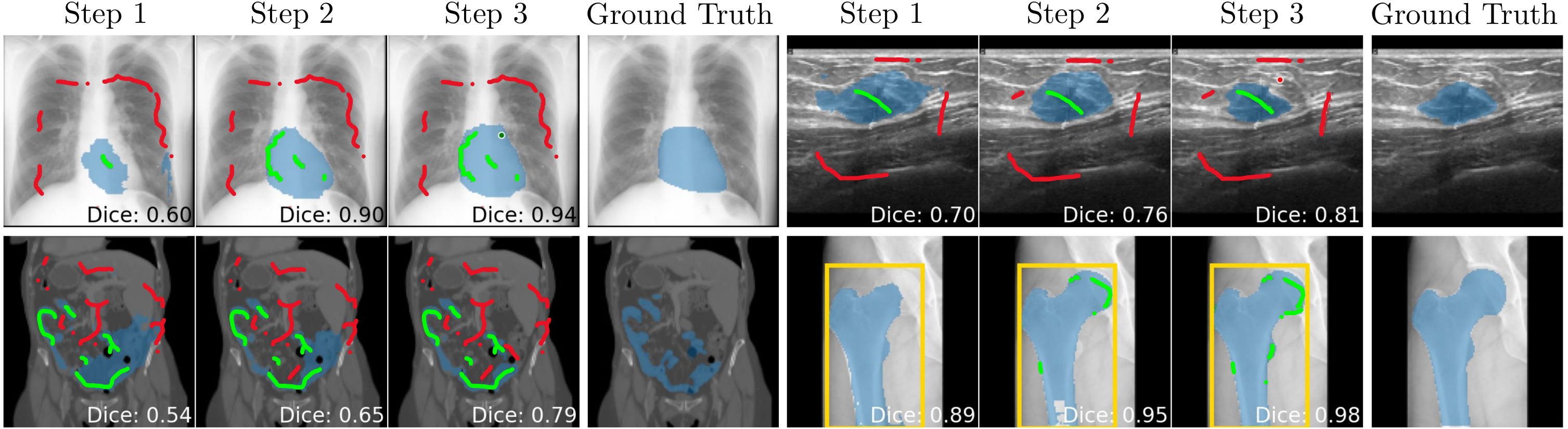}
    \caption{\textbf{ScribblePrompt enables rapid iterative interactive segmentation of \emph{unseen} tasks using bounding boxes, clicks, and scribbles.} We show predictions from ScribblePrompt with iterative interaction steps on examples from datasets unseen during training. At each step, we visualize positive scribble and click inputs in \textcolor{ForestGreen}{green}, negative scribble and click inputs in \textcolor{red}{red}, bounding box inputs in \textcolor{Dandelion}{yellow}, and the predicted segmentation in \textcolor{eccvblue}{blue}. Scribble thickness is enlarged for visual clarity. See \supp for more examples.}
    \label{fig:teaser}
\end{figure}

Most widely used interactive segmentation systems for biomedical imaging provide minimal, intensity-based, algorithmic assistance~\cite{ITK-SNAP}. Despite a growing literature, learning-based systems are less widely used in practice, perhaps because they are generally focused on specific tasks or modalities, which limits their applicability~\cite{MIDeepSeg, sakinis_IF-Seg_2019, wang_deepigeos_2019, RIL-contour, wang_bifseg_2018, zhou_volumetric_2023}. Recent vision foundation models target broad applicability, but require fine-tuning to the medical domain~\cite{SAM, zou_SEEM_2023, wang_painter_2022, wang_seggpt_2023}. Most interactive segmentation models require specific interactions, like carefully placed clicks, making them easy to develop but difficult to use in practice. The result of these shortcomings is that existing learning-based models are not widely used in practice.

In this work, we present \emph{ScribblePrompt}, a general model for interactive biomedical image segmentation that 
\begin{enumerate}
    \item enables users to rapidly and accurately accomplish any biomedical image segmentation task, outperforming state-of-the-art models, particularly for \emph{unseen} labels and image types;
    \item is flexible to different annotation styles, including bounding boxes, clicks, \emph{and} scribbles,
    \item is computationally efficient, enabling fast inference, even on a single CPU.
\end{enumerate}

To achieve this, we focused on design decisions aimed at ease of use and realistic interactions. To train our model, we started by gathering a large corpus of biomedical imaging datasets. We then designed task augmentation through synthesis strategies to encourage generalization to unseen tasks. Finally, we designed \emph{new} interaction simulation strategies for training. The model itself uses an architecture optimized for efficient inference.

To evaluate ScribblePrompt, we compared its performance to that of existing interactive segmentation methods on datasets unseen during training. These experiments involved manually-collected scribbles, simulated interactions (\cref{fig:teaser}), and a user study in which experienced annotators were asked to segment images. In the user study, ScribblePrompt reduced annotation time by 28\% while increasing Dice by 15\% compared to the next best method, the Segment Anything Model (SAM)~\cite{SAM}. Given similar amounts of interaction, ScribblePrompt achieved consistently better Dice scores than other methods. 
We release an interactive tool, code, model weights, and our dataset of manually-collected scribbles at \href{https://scribbleprompt.csail.mit.edu}{https://scribbleprompt.csail.mit.edu}.

\section{Related Works}

\subpara{Interactive Biomedical Image Segmentation}
Early research into interactive segmentation of biomedical images focused on traditional intensity-based methods~\cite{ITK-SNAP, grady_random_2006, boykov_graphcuts_2001, criminisi_geos_2008, vezhnevets_growcut_2005}. Recent deep-learning based techniques ~\cite{MIDeepSeg, RIL-contour, wang_bifseg_2018, wang_deepigeos_2019} use human interaction to improve the accuracy of segmentation tasks where labelled data exists, but fully-automatic segmentation methods fail to produce accurate segmentations. This approach assumes that the model will be used to perform the same segmentation task(s) it was trained on, leading to domain-specific solutions~\cite{wang_deepigeos_2019, atzeni_deep_2022, liu_isegformer_2022, ooi_interactive_2021, rajchl_deepcut_2017, wang_slicseg_2016}. 
A few methods generalize to new, but similar, classes \cite{sakinis_IF-Seg_2019} and modalities \cite{MIDeepSeg, wang_bifseg_2018} to those seen in training. In contrast, ScribblePrompt is designed to help annotators segment a wide range of biomedical images and (potentially unseen) tasks at inference time.

\subpara{Foundation Models}
Recent vision foundation models employ prompting to enable generalization to new tasks. These models, trained on large collections of (natural) image data, segment potentially-unseen structures specified by spatial prompts \cite{SAM,zou_SEEM_2023}, text \cite{radford_clip_2021, SAM, zou_SEEM_2023}, or examples \cite{wang_painter_2022, wang_seggpt_2023, universeg, czolbe_neuralizer_2023}. Some of these models can be used for interactive segmentation with spatial prompts. 

Fine-tuning models initially developed using natural images is often unhelpful for \emph{specific} tasks in the biomedical imaging domain, compared to training from scratch~\cite{raghu_transfusion_2019}. The Segment Anything Model (SAM)~\cite{SAM}, a foundation model trained on natural images, performs well at biomedical image segmentation tasks with clear boundaries (\eg organs in abdominal CT), but performs poorly on tasks and modalities involving more subtle delineations (\eg deep structures in Brain MRI) \cite{huang_segment_2023, mazurowski_segment_2023, shi_generalist_2023, he_accuracy_2023}. Several recent or concurrent papers fine-tune SAM for specific biomedical tasks or modalities \cite{paranjape_adaptivesam_2023, lin_samus_2023, wang_samocta_2023, kim_evaluation_2023, hu_efficiently_2023, zhang_customized_2023}. A few fine-tune SAM to segment medical images from multiple-modalities with bounding boxes \cite{MedSAM}, clicks \cite{wu_medsamadapter_2023} or both \cite{cheng_sam-med2d_2023}. We show in our experiments that these models don't perform well on many \textit{unseen} tasks and are limiting in their required interactions.

Several natural image methods simulate \emph{iterative} user interactions and condition the model on previous predictions to train a single network to make the initial prediction and perform refinement \cite{RITM, SAM, sofiiuk_fbrs_2020, bredell_iterative_2018, forte_getting_2020, simpleclick}. This approach is appropriate for natural images where there is ample data and typically one modality. However, such methods have not previously been developed for the biomedical imaging domain, which is more fragmented because of specialized regions of interest and different modalities. We use iterative ideas in ScribblePrompt, but we facilitate iterative interactive segmentation of unseen biomedical imaging tasks.

\subpara{User Interaction}
Users can prompt interactive segmentation systems in many different ways, such as bounding boxes \cite{rother_grabcut_2004, xu_deepgrabcut_2017, SAM, benenson_largescale_2019, wang_bifseg_2018, rajchl_deepcut_2017}, scribbles \cite{wang_deepigeos_2019, chen_scribbleseg_2023, wang_slicseg_2016}, or clicks \cite{wang_deepigeos_2019, MIDeepSeg, benenson_largescale_2019, simpleclick, RITM}. 
Few interactive segmentation models incorporate multiple types of inputs. Some works have explored training with more specialized types of clicks \cite{zhang_iog_2020}, such as extreme points \cite{maninis_dextr_2018, roth_going_2021}, interior margin points \cite{MIDeepSeg}, and center clicks \cite{RITM, xu_dios_2016}, to maximize the information per interaction. Specialized interactions often require more user time, and models trained only with such interactions are less robust to deviations from the interaction protocol \cite{xu_deepgrabcut_2017, RITM, benenson_largescale_2019}. In contrast, we focus on a user-friendly model that performs well under a variety of interaction scenarios. 

Scribbles are an intuitive form of interaction, however few interactive segmentation works employ them, because acquiring \emph{realistic} scribbles for training and evaluation is challenging. Previous deep-learning models trained for interactive scribble-based segmentation either relied on collecting large datasets of manually drawn scribbles \cite{zhou_volumetric_2023, lin2016scribblesup, lee_scribble2label_2020} or simplistic approaches such as sampling random points \cite{bai_errortolerant_2014, asad_econet_2022, wang_deepigeos_2019} or Bezier curves \cite{agustsson_interactive_2019}. One work has explored simulating more complex scribbles for interactive segmentation of natural images~\cite{chen_scribbleseg_2023}. We build on these concepts to create a new scribble simulation engine for training and evaluation. 

Scribble-supervised learning methods use scribble annotations as \emph{supervision} to train automatic segmentation models for predicting segmentation given only an input image~\cite{lin2016scribblesup, li_scribblevc_2023, zhang2022cyclemix, luo_scribble_2022, gotkowski2024embarrassingly}. However, these methods require the manual scribble-annotation of many training images and retraining for each new task. In contrast, ScribblePrompt is trained on a large corpus of datasets once and then can be used to perform new segmentation tasks at inference time without retraining, using scribbles as \emph{input}.

\section{ScribblePrompt~Approach}

\begin{figure}
    \centering
    \includegraphics[width=\linewidth]{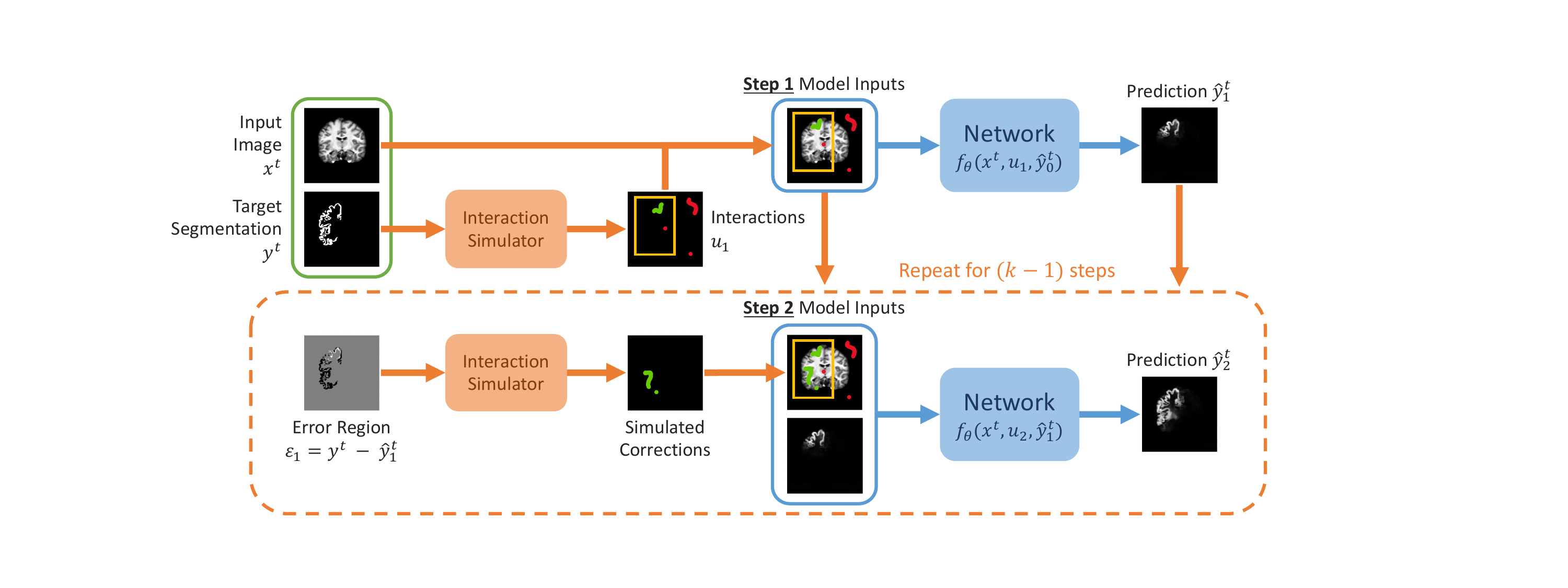}
    \caption{\textbf{Training}. We simulate $k$ consecutive steps of interactive segmentation. Given an image segmentation pair $(x^t,y^t)$, we first simulate a set of initial interactions $u_1$, which may contain bounding boxes, clicks, and/or scribbles. We predict segmentation $\hat{y}_1^t := f_\theta(x^t, u_1, \hat{y}_0^t)$ where the initial prediction $\hat{y}^t_0$ is set to zeros. In the second step, we simulate corrections using the error region $\varepsilon_1^t$ between the previous prediction $\hat{y}_1^t$ and ground truth $y^t$, and add them to the set of initial interactions $u_1$ to get $u_2$. We predict segmentation $\hat{y}_2^t := f_\theta(x^t, u_2, \hat{y}_1^t)$ and repeat to produce a series of predictions, $\hat{y}_1^t, \dots, \hat{y}_k^t$. We learn $\theta$ to minimize $\sum_{i=1}^k \mathcal{L}_{seg}(y^t, \hat{y}_i^t)$, the sum of losses between the target segmentation $y^t$ and iterative predictions $\hat{y}_1^t, \dots, \hat{y}_k^t$. 
    }
    \label{fig:training_overview}
\end{figure}

We present an interactive segmentation method that can generalize to new biomedical imaging modalities and regions of interest, while staying focused on practical usability. We describe the problem formulation, and present the important aspects of the ScribblePrompt framework: (i) simulation of realistic interactions during training, (ii) augmentation with synthetic labels during training to encourage generalization, and (iii) an efficient architecture for fast inference. We show an overview of training in \cref{fig:training_overview} and \ref{fig:schematic_data}.

\subsection{Problem Formulation}

Let $t$ be a segmentation task consisting of image and segmentation pairs,\linebreak$\{(x^t, y^t)_j\}_{j=1}^N$. At step $i$, given an image $x^t$, a set of user interactions $u_i$, and previous prediction $\hat{y}^t_{i-1}$, we learn function $f_\theta(x^t, u_i, \hat{y}^t_{i-1})$ with parameters $\theta$ that produces a segmentation $\hat{y}_i$. The set of interactions $u_i$, which may include positive or negative scribbles, positive or negative clicks, and bounding boxes, is provided by a user who has access to the image $x^t$ and previous prediction $\hat{y}_{i-1}^t$.

We minimize the difference between the true segmentation $y^t$ and each of the $k$ iterative predictions $\hat{y}_1, \dots, \hat{y}_k$, 
\begin{equation}\label{eq:loss}
    \mathcal{L}(\theta; \mathcal{T}) = \mathbb{E}_{ t \in \mathcal{T} } \left[ \mathbb{E}_{(x^t,y^t) \in t} \left[ \sum_{i=1}^k \mathcal{L}_{Seg} \left(y^t, f_\theta(x^t, u_i, \hat{y}^t_{i-1}) \right) \right] \right] ,
\end{equation}
where $\mathcal{L}_{Seg}$ is a supervised segmentation loss. 

During training, we sample a task $t$ from training task collection $\mathcal{T}$, from which we sample an image $x^t$, and segmentation map $y^t$. We simulate a possible set of interactions $u_i$ based on $y^t$, and predict $\hat{y}_{i}^t = f_\theta(x^t, u_i, \hat{y}^t_{i-1})$. We simulate the next set of interactions $u_{i+1}$ based on the error region $\varepsilon_i^t = y^t - \hat{y}_{i}^t$, and repeat for $k$ iterations. In the following sections, we describe the core aspects of the framework: strategies for simulating $u_i$, collecting $\mathcal{T}$, sampling $(x^t,y^t)$, building $f_\theta$, and optimizing $\mathcal{L}(\theta; \mathcal{T})$.

\subsection{Prompt Simulation}

\begin{figure}[t]
  \centering
  \includegraphics[width=\linewidth]{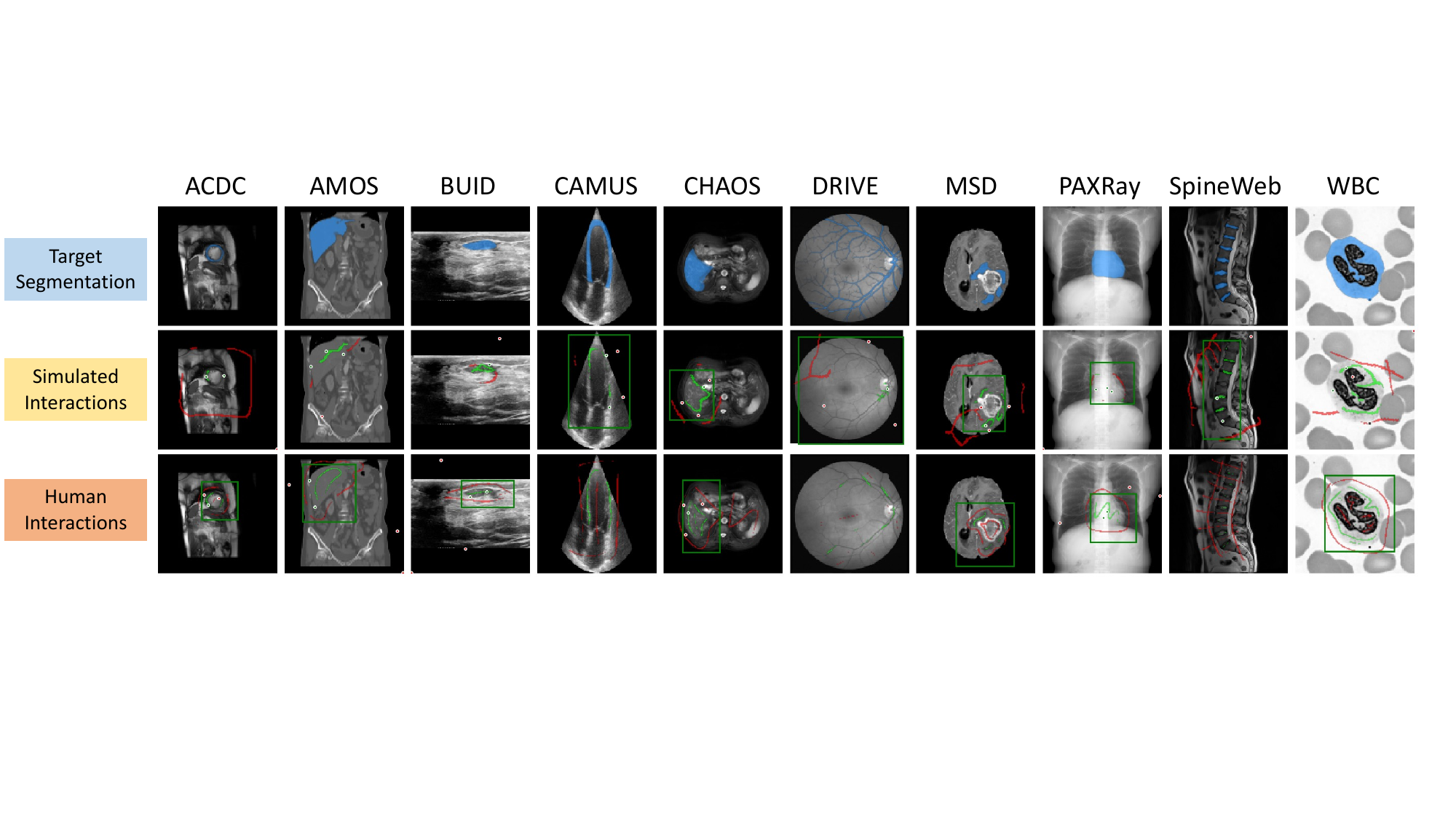}
  \caption{\textbf{Simulated scribbles and clicks}. Positive interactions (\textcolor{ForestGreen}{green}) are simulated on the segmentation label $y^t$ (\textcolor{eccvblue}{blue}), while negative interactions (\textcolor{red}{red}) are simulated on the background $1 - y^t$. Scribble thickness is enlarged for visual clarity.}
  \label{fig:click_scribble_strategies}
\end{figure}

To enable a practical, easy-to-use model, we encourage robustness to different types of user interactions $u_i$. We introduce algorithms for simulating scribbles, clicks, and bounding box inputs during training. 
Each scribble and click strategy (\cref{fig:click_scribble_strategies}) can be applied to a ground truth segmentation label $y^t$ to simulate positive interactions, or to the background $1-y^t$ to simulate negative interactions. We simulate positive and negative correction scribbles or clicks by applying the same strategies to the false negative region $\varepsilon_i^t>0$ and false positive region $\varepsilon_i^t<0$ for error region $\varepsilon_i^t$.

\subpara{Scribbles}
Given binary mask $y \in\{0,1\}^{h \times w}$, we simulate a scribble mask \linebreak$s \in [0,1]^{h \times w}$ by first generating clean scribbles, and then corrupting them to account for user behavior and variability. We illustrate this process in \supp Sec. \ref{appendix:scribble_simulation}.
To generate the clean scribbles, we start with one of these strategies:
\begin{enumerate}[label=(\roman*)]
    \item \textbf{Line Scribbles}: We draw random lines by connecting two end points sampled from $\{(u,v)| y_{uv}=1\}$. 
    \item \textbf{Centerline Scribbles}: We simulate scribbles in the center of label $y$ using a thinning algorithm~\cite{zhangsuen_thinning_1984} that reduces the label to a 1-pixel wide skeleton. 
    \item \textbf{Contour Scribbles}:
    We simulate a rough contour of the desired segmentation within the boundaries of the mask. We first blur the mask to reduce the size of the label such that $\Tilde{y} = \min(y, y \circ G_k)$, where $G_k$ is a Gaussian blur kernel. Then we apply a threshold $\Tilde{y} < h$ sampled in some intensity range $h \sim U[\Tilde{y}_{min}, \Tilde{y}_{max}]$ and extract a contour inside the boundary of the mask.
\end{enumerate}

\noindent To limit the size and complexity of the centerline and contour scribbles, we apply a random mask that breaks the scribbles into smaller parts. We generate the random mask by sampling a smooth noise image $p$, where each pixel is sampled independently from $\mathcal{N}(\mu_p, \sigma_p^2)$, applying Gaussian blur, and then thresholding it at $\mu_p$. 
We warp the resulting scribble mask $s$ using a random deformation field $\phi$ to vary the scribble shape and thickness. We ensure the resulting scribble is consistent with mask $y$, by multiplying the warped scribble mask $s \circ \phi$ by $y$.

\subpara{Clicks}
Given a binary mask $y \in \{0,1\}^{h \times w}$, we simulate $n \sim U[n_{min}, n_{max}]$ clicks at a time using one of three strategies, illustrated in \cref{fig:click_scribble_strategies}:
\begin{enumerate}[label=(\roman*)]
    \item \textbf{Random clicks}: We randomly sample clicks from all pixels in the given region $\{(u,v) | y_{uv}= 1\}$. 
    
    \item \textbf{Center clicks}: We sample clicks from the set of points at the center of each disconnected component of the label. First, we create a multi-label mask $m \in \{1,\dots,C\}^{h \times w}$ identifying the $C$ components of label $y$. We identify the center of each components using the euclidean distance transform \cite{RITM, xu_dios_2016}.

    \item \textbf{Interior border region clicks}: We sample clicks from a border region inside the boundary of the mask. We first blur the mask $y$ to reduce the size of the label such that $\Tilde{y} = \min(y, y \circ G_k)$ where $G_k$ is a Gaussian blur kernel. We then sample click coordinates from $\{(u,v) | \Tilde{y}_{uv} \in [a,b] \}$, where $a,b \sim U[\Tilde{y}_{min}, \Tilde{y}_{max})$ are thresholds sampled in some intensity range. 
    
\end{enumerate}
Since users are inclined to spread out their clicks, we impose a minimum separation of a few pixels between random clicks and border region clicks. 

\subpara{Bounding Boxes}
We compute the minimum bounding box that encloses the label $y^t$, and enlarge each dimension by $r \sim U[0,20]$ pixels to account for human variability.

\subpara{Iterative Training}
During the first step ($i=1$), we sample the combination of interactions and the number of initial positive and negative interactions $n_{pos},n_{neg} \sim U[n_{min},n_{max}]$. The initial interactions $u_1$ are simulated using the ground truth label $y^t$. In subsequent steps, correction scribbles or clicks are sampled from the error region $\varepsilon_{i-1}^t$ between the last prediction $\hat{y}^t_{i-1}$ and the ground truth $y^t$. Since a user can make multiple corrections in each step, we sample $n_{cor} \sim U[n_{min},n_{max}]$ corrections (scribbles or clicks) per step.

\subsection{Data}
We build on large dataset gathering efforts like MegaMedical~\cite{universeg, tyche} to compile a collection of 77 open-access biomedical imaging datasets for training and evaluation, covering over 54k scans, 16 image types, and 711 labels. The collection includes a diverse array of biomedical domains, such as eyes~\cite{STARE, OCTA500, Rose, IDRID, DRIVE}, thorax~\cite{CheXplanation, LUNA, MSD, PAXRay, SCD}, spine~\cite{SpineWeb, VerSe, PAXRay, TotalSegmentator}, cells~\cite{WBC, BBBC003, BBBC038, ssTEM, ISBI_EM, PanNuke}, skin~\cite{ISIC}, abdominal~\cite{LiTS, NCI-ISBI, KiTS, AMOS, CHAOS_1, SegTHOR, BTCV, I2CVB, Promise12, Word, SCD, MSD, DukeLiver, CT2US, CT_ORG}, neck~\cite{SegThy, HaN-Seg, NerveUS, DDTI}, brain~\cite{BRATS, MCIC, ISLES, WMH, BrainDevelopment, OASIS-data, PPMI, LGGFlair, MSD, COBRE}, bones~\cite{PAXRay, HipXRay, TotalSegmentator}, teeth~\cite{PanDental, ToothSeg} and lesions \cite{BUID, BUSIS, MMOTU, MSD}. 

We define a 2D segmentation task as a combination of dataset, axis (for 3D modalities), and label. For datasets with multiple segmentation labels, we consider each label separately as a binary segmentation task and for 3D modalities we use the slice with maximum label area and the middle slice from each volume. We provide more details on the data in \supp Sec. \ref{appendix:data}.

\subpara{Task Diversity}
During training we sample image, segmentation pairs hierarchically -- by dataset and modality, axis, and then label -- to balance training on datasets of different sizes. To increase the diversity of segmentation tasks, we apply data augmentation~\cite{universeg} to both the input image and sampled segmentation prior to simulating the user interactions.

\begin{figure}
    \centering
    \includegraphics[width=\linewidth]{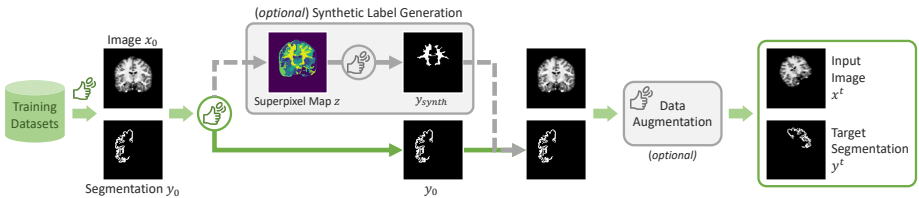}
    \caption{\textbf{Task sampling and augmentation}. During training, we sample an image and segmentation pair $(x_0, y_0)$. With probability $p_{synth}$, we replace $y_0$ with a synthetic label $y_{synth}$. We generate $y_{synth}$ by applying a superpixel algorithm to the image $x_0$ to generate a map $z$ of potential synthetic labels (superpixels), and then sampling one label. Finally, we apply random data augmentations to get $(x^t, y^t)$.}
    \label{fig:schematic_data}
\end{figure}

\subpara{Synthetic Labels}
To limit overfitting to specific segmentation tasks, we introduce a mechanism to generate synthetic labels (\cref{fig:schematic_data}). 
During training, for a given sample $(x_0, y_0)$, with probability $p_{synth}$, we replace $y$ with a synthetic label $y_{synth}$. 
Given an image $x_0$, we generate a synthetic label $y_{synth}$ by applying a superpixel algorithm \cite{felzenszwalb_superpixels} with scale parameter $\lambda \sim U[1,\lambda_{max}]$ to partition the image into a multi-label mask of $k$ superpixels $z \in \{1, \dots, k\}^{n \times n}$, and then randomly select a superpixel $y_{synth} = \mathbbm{1} (z = c)$. 
We conduct experiments varying $p_{synth}$ and show examples in \supp Sec. \ref{appendix:synth}.

\subsection{Network}

Motivated by producing a practical tool, we primarily demonstrate ScribblePrompt using an efficient fully-convolutional architecture similar to a UNet~\cite{UNet}. We also demonstrate ScribblePrompt using a vision transformer architecture~\cite{SAM}.

\textbf{ScribblePrompt-UNet:}
We use an 8 layer CNN following a decoder-encoder structure similar to the popular UNet architecture \cite{UNet} without Batch Norm. Each convolutional layer has 192 features and uses PReLu activation \cite{he_prelu_2015}.  

\textbf{ScribblePrompt-SAM:}
We take the smallest SAM model (ViT-b) ~\cite{SAM} and fine-tune the decoder. SAM takes bounding box and click inputs as lists of ${(x,y)}$ coordinates, and the low-resolution logits of the previous prediction $\hat{y}_{i-1}$. To adapt SAM for scribbles, we consider each scribbled pixel as a click. 

\subpara{Loss} 
We minimize eq. (\ref{eq:loss}) where $\mathcal{L}_{Seg}$ is a linear combination of soft Dice Loss \cite{dice1945measures} and Focal Loss \cite{focal_loss}. In preliminary experiments (\supp Sec. \ref{appendix:architecture_ablations}), training with Soft Dice Loss alone or a linear combination of Soft Dice Loss and Binary-Cross Entropy Loss resulted in slightly lower Dice scores.


\section{Experimental Setup}

We compare ScribblePrompt-UNet and ScribblePrompt-SAM to previous methods through experiments with manual scribbles, simulated interactions, and a user study with experienced annotators. Lastly, we report on inference runtime and ablation experiments.

\subpara{Data}
We use 65 (out of 77) datasets during training. We partition another nine datasets, ACDC~\cite{ACDC}, BUID~\cite{BUID}, BTCV Cervix~\cite{BTCV}, DRIVE~\cite{DRIVE}, HipXRay~\cite{HipXRay}, PanDental~\cite{PanDental}, SCD~\cite{SCD}, SpineWeb~\cite{SpineWeb}, and WBC~\cite{WBC}, each into validation (used in model selection, but not training) and test (used only for final evaluation). Three additional datasets, TotalSegmentator~\cite{TotalSegmentator}, SCR~\cite{SCR}, and COBRE~\cite{COBRE,fischl2012freesurfer,neurite}, were used only for final evaluation. 

We report final results on 12 sets of data (the test splits of the 9+3 evaluation datasets). The evaluation datasets were not seen during training, and cover 608 tasks and 8 modalities, including unseen image types and unseen labels. We selected these 12 evaluation datasets to cover a variety of modalities (MRI, CT, ultrasound, fundus photography, microscopy) and anatomical regions of interest (brain, teeth, bones, abdominal organs, muscles, heart, thorax, cells), including both healthy anatomy and lesions. 

\subpara{Training}
During training, we simulated five steps of interactive segmentation per example and set the maximum number of interactions per step to three. We set the minimum number of initial negative prompts to zero, and the minimum initial positive and correction prompts to one. During the first step, we sample the combination of prompt types (clicks, scribbles, bounding boxes) and then sample the number of positive and negative interactions for each type. In each subsequent steps, we simulate either scribble corrections or click corrections with equal probability. We selected these values based on what we believe to be reasonable interactions for a user to perform. 


\subpara{Baselines}
We compare to existing generalist methods for interactive segmentation, with a focus on methods developed for biomedical imaging. 

\textbf{SAM}~\cite{SAM}: We evaluate the smallest (ViT-b) and largest (ViT-h) versions of the Segment Anything Model (SAM) trained on natural images. SAM takes bounding boxes, clicks, and the logits of the previous prediction as input. 

\textbf{SAM-Med2D}~\cite{cheng_sam-med2d_2023}: SAM-Med2D is a SAM ViT-b model with additional adapter layers in the image encoder. SAM-Med2D was fine-tuned, using bounding boxes and iterative clicks as input, on a collection of biomedical imaging datasets containing 4.6M images and 19.7M segmentations. Following \cite{cheng_sam-med2d_2023}, we evaluate SAM-Med2D both with and without adapter layers.

\textbf{MedSAM}~\cite{MedSAM}: MedSAM is a SAM ViT-B model fine-tuned with bounding box prompts on a collection of biomedical imaging datasets containing 1.5M image segmentation pairs. We evaluate MedSAM with bounding boxes only, because we found it to perform poorly when given point or mask prompts.

\textbf{MIDeepSeg}~\cite{MIDeepSeg}: MIDeepSeg is an interactive segmentation framework designed to generalize to unseen tasks on medical images. MIDeepSeg takes interior margin points (positive clicks) as initial inputs, crops the image based on those points, and uses a CNN to make an initial prediction. Given additional clicks, the prediction can be refined using conditional random fields. We evaluate the pre-trained model, which was developed on placenta T2 MRI.

\section{Evaluation}

We evaluate all methods using both manual and simulated interactions, with a focus on scribbles. Simulated interactions enable us to test on a large volume of images and tasks. However, simulations do not always match user behavior. We address this by (1) collecting a diverse dataset of manual scribbles for evaluation, and (2) conducting an interactive user study.

\subsection{Manual Scribbles}
\label{sec:manual_scribbles_experiment}

\subpara{Setup}
We evaluate on two datasets of manual scribbles. First, we collected a dataset (MedScribble) of manual scribbles from three annotators for seven segmentation tasks from unseen datasets, with a total of 31 examples. The annotators were shown five training examples with the ground truth segmentation per task and instructed to draw positive and negative scribbles to indicate the region of interest on new images. We provide more details on MedScribble and release the data in \supp Sec. \ref{appendix:medscribble}. Second, we used 380 slices from the ACDC dataset~\cite{ACDC}, which has scribble annotations for three labels and background. 

In this evaluation, we make a prediction given a set of positive and negative scribbles, simulating a non-iterative scenario where the annotator draws several scribbles before running inference. We evaluate accuracy using Dice score \cite{dice1945measures} and 95th percentile Hausdorff Distance~\cite{huttenlocher_hd_1993}. Since MIDeepSeg's initial prediction network only takes positive inputs, we report results on both MIDeepseg's initial prediction from the positive scribbles, and after applying MIDeepSeg's refinement procedure using the negative scribbles as corrections. For SAM and variants fine-tuning SAM on clicks, we convert each scribble to a series of points. For MedSAM, we fit a bounding box to the positive scribbles. 

\begin{figure}
  \centering
  \includegraphics[width=\linewidth]{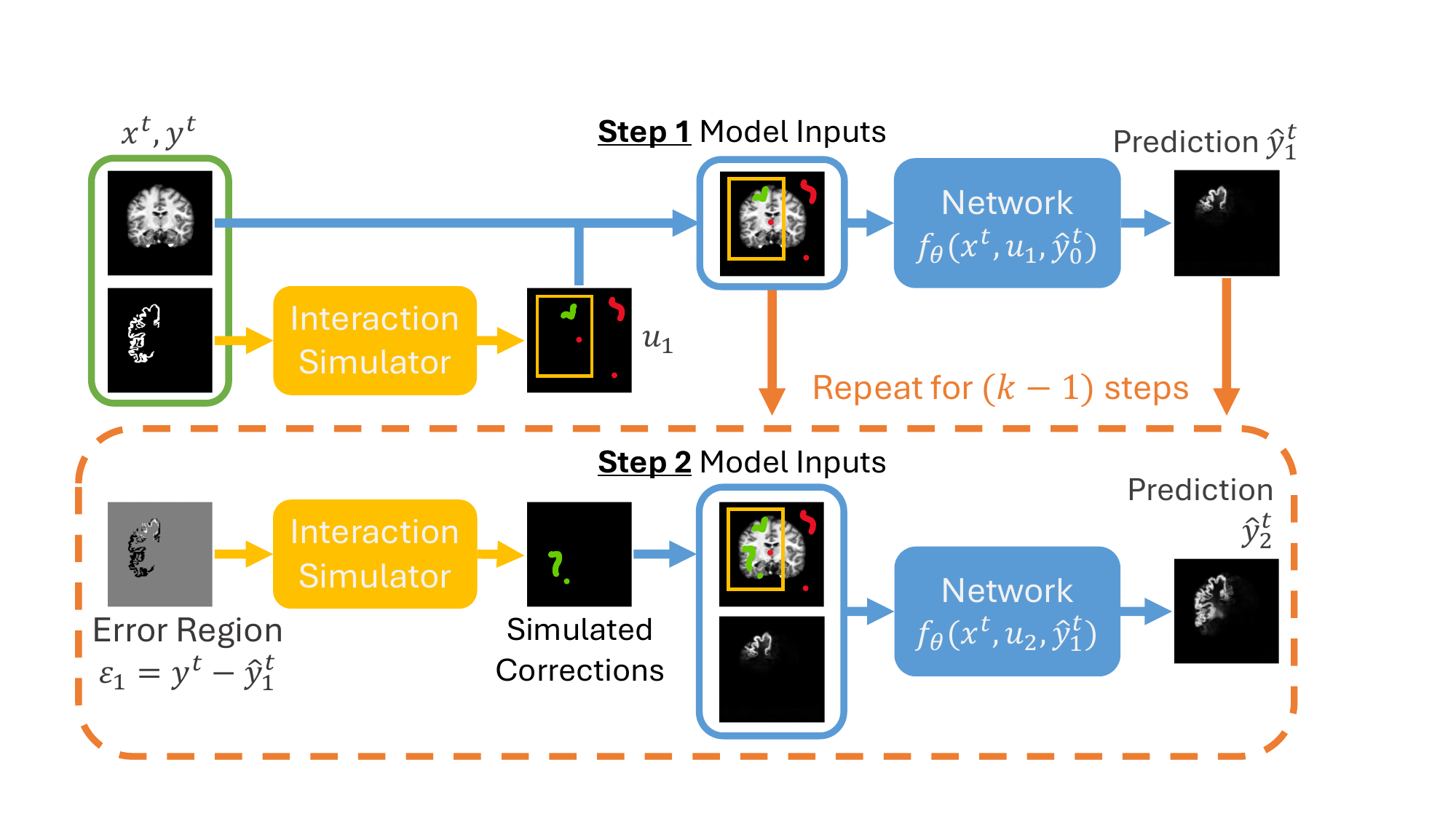}
  \caption{\textbf{Example predictions}. 
  SP = ScribblePrompt. Top: predictions after one step of manual scribbles. Bottom: predictions after five steps of simulated interactions (one center click followed by one center correction click per step). 
  }
  \label{fig:qualitative_examples}
\end{figure}

\subpara{Results}
ScribblePrompt-UNet and ScribblePrompt-SAM produce the most accurate segmentations in a single step of manual scribbles on both our manual scribbles dataset and the ACDC scribbles dataset (\cref{tab:manual_scribbles}). We show example predictions in \cref{fig:qualitative_examples} and additional comparisons to scribble-supervised learning in \supp Sec. \ref{appendix:scribblesup}.

SAM and SAM-Med2D do not generalize well to scribble inputs (which they were not trained for). MedSAM has better predictions than other SAM baselines using the SAM architecture, however it is not able to make use of the negative scribbles and thus often misses segmentations with holes in them (\cref{fig:qualitative_examples}). The initial predictions from MIDeepSeg's network are poor, but improve after applying the refinement procedure.

\begin{table}
    \caption{\textbf{Manual scribbles.} Mean Dice and HD95 with 95\% CI of segmentations predicted from manually-collected scribbles. \firstone{Best} and \secondone{second best} are highlighted. 
}
    \label{tab:manual_scribbles}
    \centering
    \rowcolors{2}{white}{gray!15}
    \resizebox{\textwidth}{!}{
    \setlength{\tabcolsep}{0.5em}
    \begin{tabular}{l|cc|cc}
    \toprule
    & \multicolumn{2}{c}{ MedScribble (n=93)} & \multicolumn{2}{|c}{ACDC (n=1,140) }
    \\
    \midrule
    Model & $\uparrow$ Dice Score & $\downarrow$ HD95 & $\uparrow$ Dice Score & $\downarrow$ HD95 
    \\
    \midrule
    SAM (ViT-b) & $0.40 \pm 0.05$ & $32.58 \pm 3.80$ & $0.20 \pm 0.01$ & $108.39 \pm 0.85$
    \\
    SAM (ViT-h) & $0.56 \pm 0.05$ & $14.61 \pm 3.18$ & $0.42 \pm 0.02$ & $59.78 \pm 2.02$
    \\
    SAM-Med2D w.o. adapter & $0.52 \pm 0.05$ & $30.34 \pm 3.29$ & $0.16 \pm 0.01$ & $107.25 \pm 1.12$
    \\
    SAM-Med2D w/ adapter & $0.30 \pm 0.06$ & $19.04 \pm 3.63$ & $0.17 \pm 0.02$ & $41.01 \pm 3.04$
    \\
    MIDeepSeg & $0.68 \pm 0.04$ & $6.94 \pm 1.06$ & $0.58 \pm 0.01$ & $5.78 \pm 0.26$
    \\
    MIDeepSeg w/ refinement & $0.81 \pm 0.03$ & $3.10 \pm 0.67$ & $0.73 \pm 0.01$ & \secondone{$2.71 \pm 0.18$}
    \\
    \midrule
    MedSAM (box) & $0.70 \pm 0.04$ & $7.54 \pm 1.35$ & $0.70 \pm 0.02$ & $3.53 \pm 0.28$
    \\
    \midrule
    \textbf{ScribblePrompt-UNet} & \secondone{$0.84 \pm 0.02$} & \secondone{$2.92 \pm 0.88$} & \firstone{$0.84 \pm 0.01$} & \firstone{$1.80 \pm 0.11$}
    \\
    \textbf{ScribblePrompt-SAM} & \firstone{$0.87 \pm 0.02$} & \firstone{$2.61 \pm 0.84$} & \secondone{$0.77 \pm 0.01$} & $4.17 \pm 0.32$
    \\
    \bottomrule
    \end{tabular}
    }
\end{table}

\subsection{Simulated Interactions}
\label{sec:simulated_experiment}

\subpara{Setup}
We use our interaction simulator to evaluate the performance of ScribblePrompt and baselines with iterative scribbles across the 12 evaluation sets.

For scribble-focused prompting procedures, we use:
\begin{itemize}[leftmargin=*]
    \item \textbf{Line Scribbles}: Three positive line scribbles and three negative line scribbles to start, followed by one correction line scribble at each step.
    \item \textbf{Centerline Scribbles}: One positive and one negative centerline scribble to start, followed by one correction centerline scribble at each step.
    \item \textbf{Contour Scribbles}: One positive and one negative contour scribble to start, followed by one correction contour scribble at each step. 
\end{itemize}
We limit each scribble to cover a maximum of $w$ pixels, where $w$ is the maximum dimension of the image. For centerline and contour scribbles, a scribble might contain multiple disconnected components. 

For click-focused prompting procedures, we use:
\begin{itemize}[leftmargin=*]
    \item \textbf{Center Clicks}: One positive click in the center of the largest component to start, followed by one (positive or negative) correction click per step in the center of the largest component of the error region.
    \item \textbf{Random Clicks}: One random positive click to start, followed by one (positive or negative) correction click per step, randomly sampled from the error region.  
    \item \textbf{Random Warm Start}: Three random positive clicks and three random negative clicks to start, followed by one (positive or negative) correction click per step in the center of the largest component of the error region.  
\end{itemize}
For each example and interaction procedure, we simulate a series of iterative interactions with five random seeds. Since MIDeepSeg cannot make accurate predictions with only a few clicks, we exclude it from simulations that start with a single click. For MedSAM, we show results for a bounding box prompt.

\subsubsection{Results.}
\cref{fig:simulated_results} shows that both versions of ScribblePrompt outperform baseline methods for all simulated interaction procedures at all numbers of interactions. We show examples with simulated interactions in \cref{fig:teaser} and \cref{fig:qualitative_examples} (bottom). We show results with similar trends using bounding boxes and by dataset with fully-supervised baselines in \supp Sec. F. 

\subsubsection{Discussion.}
\begin{figure}[t]
  \centering
  \includegraphics[width=\linewidth]{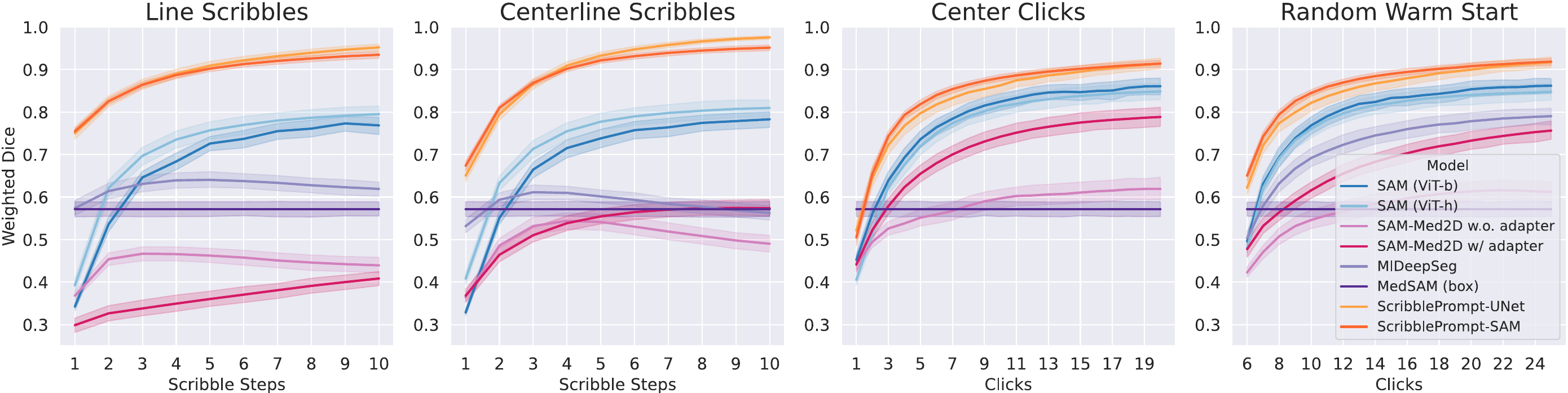}
  \caption{\textbf{Simulated clicks and scribbles}. We simulate interactions following three scribble protocols and three click protocols. We show more results and example predictions, with similar trends, in \supp Sec. \ref{appendix:simulated_clicks_scribbles}. 
  }
  \label{fig:simulated_results}
\end{figure}

Comparing the simulated scribble and click results (\cref{fig:simulated_results}) highlights the efficiency of scribble-based interactions. Both ScribblePrompt models reach an average Dice above $0.8$ in two scribble steps. It takes five steps of carefully-placed center clicks or eight random clicks to reach the same average Dice score with ScribblePrompt-SAM. 

Although SAM-Med2D was trained on a large biomedical imaging collection, it does not perform as well as SAM at refining its predictions on held-out evaluation sets. MedSAM only uses bounding box prompts and is not able to refine predictions, limiting its performance and usability. 
 
\subsection{User Study}

To assess the practical utility of ScribblePrompt, we conducted a user study with experienced annotators.
We compare ScribblePrompt-UNet to SAM (ViT-b). We selected ScribblePrompt-UNet for the user study because it had similar performance and lower latency compared to ScribblePrompt-SAM. We selected SAM (ViT-b) because it had the highest Dice score in our experiments with clicks and lower latency than the next closest baseline, SAM (ViT-h). 

\subpara{Setup}
Study participants ($n=16$) were neuroimaging researchers at an academic hospital. They used each model to segment a series of nine test images from different evaluation datasets. For each segmentation task, participants were shown the target segmentation and were asked to prompt the model until the predicted segmentation closely matched the target, or they could no longer improve the prediction. We provided participants with the target segmentation to disentangle the cognitive process of identifying the region of interest from prompting the model to achieve the desired segmentation. We provide additional details and visualizations in \supp Sec. \ref{appendix:user_study}. 

\subpara{Results}
Participants produced more accurate segmentations (0.84 vs. 0.73 Dice; $p = 0.001$ using a paired t-test) using ScribblePrompt (\cref{tab:userstudy}). Participants spent $\approx 1.5$ minutes per segmentation on average using ScribblePrompt, compared to over 2 minutes per segmentation with SAM ($p = 0.02$ using a paired t-test). While using ScribblePrompt, participants updated the prediction fewer times before being satisfied with the segmentation.

Upon completion, 15 out of 16 participants reported they preferred using ScribblePrompt to SAM and one participant had no preference. All participants reported it was easier to achieve the target segmentation using ScribblePrompt compared to SAM. 93.8\% of participants reported ScribblePrompt was better than SAM at refining its predictions in response to scribbles. 87.5\% of participants preferred using ScribblePrompt over SAM for clicks. 

\subpara{Discussion} 
The most common factors that influenced participant preference for ScribblePrompt were 1) being able get accurate predictions from multiple types of prompts, including scribbles and 2) responsiveness to their corrections, which enabled fine-grained control over the next prediction. For some tasks, such as retinal vein segmentation, SAM was unable to make accurate predictions even with many corrections. As a result, there was more variability in Dice score and time per segmentation, for SAM than ScribblePrompt. 

\begin{table}[t]
    \caption{\textbf{User study.} Mean $\pm$ std. for Dice score, HD95 and total time per segmentation. Median (and max) number of times the participants refreshed the prediction.
    }
    \label{tab:userstudy}
    \centering
    \resizebox{0.95\textwidth}{!}{
    \setlength{\tabcolsep}{0.5em}
    \begin{tabular}{lcccc}
    \toprule
    & Dice Score ($\uparrow$) & HD95 ($\downarrow$) & Sec./task ($\downarrow$) & Iter./task ($\downarrow$) 
    \\
    \midrule
    SAM (ViT-b) & $0.73 \pm 0.30$ & $16.1 \pm 37.1$ & $131.7 \pm 168.5$ & $6 ~ (47)$ 
    \\
    ScribblePrompt-UNet & \firstone{$0.84 \pm 0.13$} & \firstone{$4.5 \pm 9.1$} & \firstone{$94.2 \pm 89.0$} & \firstone{$3 ~(19)$}
    \\
    \bottomrule
    \end{tabular}
    }
\end{table}

\subsection{Inference Runtime}

\subpara{Setup}
We evaluate computational efficiency by measuring inference time on a single CPU for one prediction with a scribble input covering 128 pixels. Performance on a CPU reflects practical utility. Because of a variety of barriers such as protected health information, users may not be able to send their data to a server for computation, and must rely on local, most often CPU-only, computing.

\subpara{Results}
On a single CPU, ScribblePrompt-UNet requires $0.27 \pm 0.04$ sec per prediction, enabling the model to be used even in low-resource environments. The next most accurate model (\cref{fig:simulated_results}), SAM ViT-h, requires over 2 minutes per prediction on a CPU ($130.79 \pm 7.96$ sec). SAM ViT-b takes $13.59 \pm 0.77$ seconds per prediction. We report results for all baselines and on GPU hardware with similar trends in \supp Sec. \ref{appendix:gpu_runtime}.

\section{Ablations}\label{sec:ablations}

We conduct two ablations of important design decisions: (1) synthetic label inputs used during training, and (2) types of prompts simulated during training. We report results on the validation splits of nine datasets unseen during training. 

\begin{figure}[t]
  \centering
  \begin{subfigure}{0.4\linewidth}
    \includegraphics[width=\linewidth]{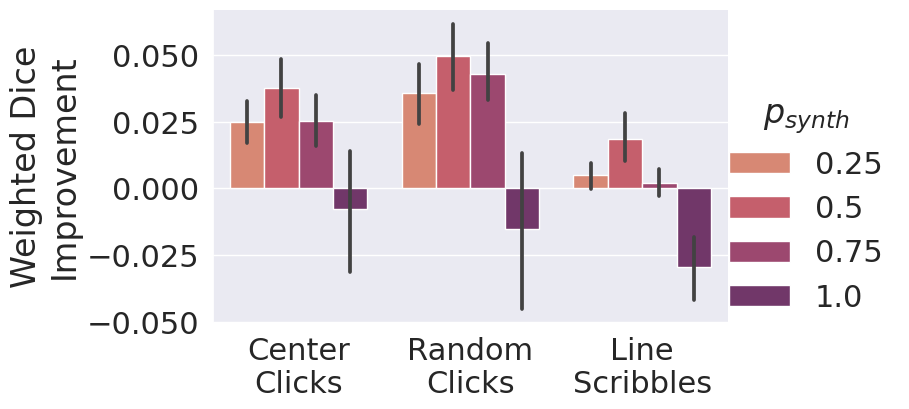}
    \caption{\textbf{Probability of synthetic labels.}} 
    \label{fig:superpixel_ablation_unet}
  \end{subfigure}
  \hfill
  \begin{subfigure}{0.57\linewidth}
      \includegraphics[width=\linewidth]{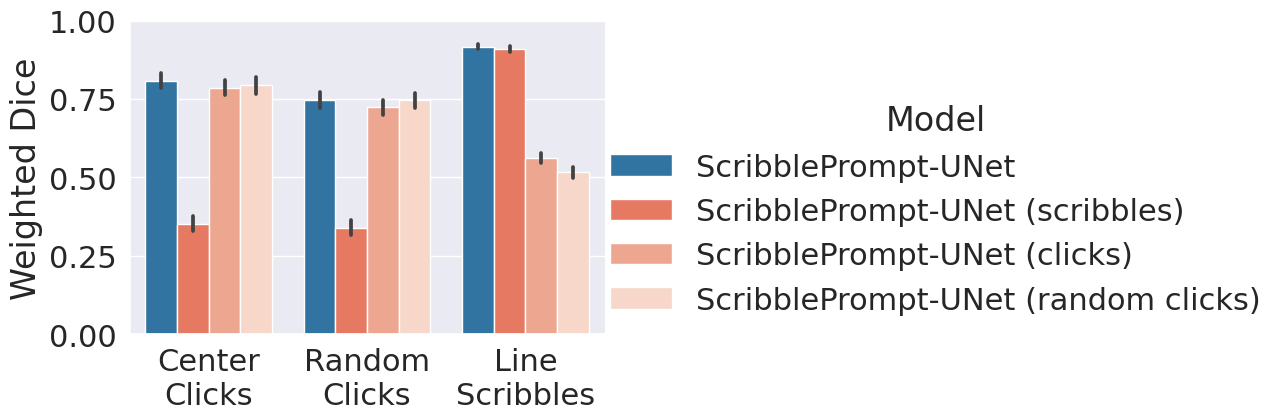}
    \caption{\textbf{Interactions during training.}}
    \label{fig:prompt_ablation}
  \end{subfigure}
  \caption{\textbf{Ablations}. We report Dice after five steps of simulated interactions following three inference-time interaction procedures. Error bars show 95\% CI. \textbf{(a)} shows mean change in Dice relative to ScribblePrompt-UNet trained without any synthetic labels. 
  }
  \label{fig:short}
\end{figure}

\subsection{Synthetic Labels}\label{sec:ablation_synth}

\cref{fig:superpixel_ablation_unet} shows the effect of varying the probability of sampling a synthetic label during training for ScribblePrompt-UNet. Training with both real and synthetic labels improves generalization to new datasets, compared to training with only real labels. Using $p_{synth}=0.5$ results in the highest Dice. Training with \emph{only} synthetic labels results in worse Dice scores. We show similar results for ScribblePrompt-SAM and other interactions in \supp Sec. \ref{appendix:ablation_synth}.

\subsection{Prompt Types}

\textbf{Setup.}
We evaluate ScribblePrompt-UNet models trained with different combinations of prompts, compared to the complete ScribblePrompt-UNet: 
\begin{itemize}[leftmargin=*]
    \item \textbf{ScribblePrompt-UNet~(scribbles)} trained on boxes and scribbles.
    \item \textbf{ScribblePrompt-UNet~(clicks)} trained on boxes and clicks.
    \item \textbf{ScribblePrompt-UNet~(random clicks)} trained on boxes and random clicks. 
\end{itemize}

\noindent\textbf{Results.}
\cref{fig:prompt_ablation} shows ScribblePrompt-UNet trained with scribbles, clicks and bounding boxes predicts segmentations more accurately than do ablated versions of ScribblePrompt-UNet. We show results for other interaction procedures with similar trends in \supp Sec. \ref{appendix:ablation_prompt}.

\section{Conclusion}

We present ScribblePrompt, a practical framework for interactive segmentation that enables users to segment diverse medical images with scribbles, clicks, and bounding boxes. We introduce methods for simulating realistic user interactions and generating synthetic labels. These methods enable us to train models that generalize to unseen segmentation tasks and datasets. ScribblePrompt is  more accurate than existing baselines, and ScribblePrompt-UNet is computationally efficient, even on a CPU. Our user study shows that nearly all users prefer ScribblePrompt and achieve segmentations with 15\% higher Dice with less effort than the next most accurate baseline. ScribblePrompt promises to significantly reduce the burden of manual segmentation in biomedical imaging. 

\section*{Acknowledgements}

 This work was supported in part by funding from Quanta Computer Inc., the Eric and Wendy Schmidt Center at the Broad Institute of MIT and Harvard and the Wistron Corporation. Research reported in this paper was supported by the National Institute of Biomedical Imaging and Bioengineering of the National Institutes of Health under award number R01EB033773. Much of the computation resources required for this research was performed on computational hardware generously provided by the Massachusetts Life Sciences Center.

%
%
\bibliographystyle{splncs04}
\bibliography{main}

\newpage

\title{ScribblePrompt: Fast and Flexible Interactive Segmentation for Any Biomedical Image} 

\titlerunning{ScribblePrompt}

\author{Hallee E. Wong\inst{1,2}\orcidlink{0000-0003-1343-9672} \and
Marianne Rakic\inst{1,2}\orcidlink{0000-0003-2376-9448} \and
John Guttag\inst{1}\orcidlink{0000-0003-0992-0906} \and
{Adrian~V.~Dalca}\inst{1,2,3}\orcidlink{0000-0002-8422-0136}
}

\authorrunning{H.E.~Wong et al.}

\institute{MIT CSAIL, Cambridge, MA, USA 
\and 
Martinos Center, Massachusetts General Hospital, Charlestown, MA, USA
\and 
Harvard Medical School, Boston, MA, USA 
\\
\email{\{hallee,mrakic,guttag,adalca\}@mit.edu}
}
\maketitle

\appendix

\section*{Table of Contents}

\cref{appendix:demo}:~Demo and Code
\\
\noindent\cref{appendix:implementation}:~ScribblePrompt Implementation 
\\
\noindent\cref{appendix:data}:~Data
\\
\noindent\cref{appendix:setup}:~Experimental Setup
\\
\noindent\cref{appendix:manual_scribbles}:~Evaluation with Manual Scribbles
\\
\noindent\cref{appendix:experiments}:~Evaluation with Simulated Interactions
\\
\noindent\cref{appendix:user_study}:~User Study 
\\
\noindent\cref{appendix:gpu_runtime}:~Inference Runtime
\\
\noindent\cref{appendix:ablations}:~Ablations 
\\

\section{Demo and Code}\label{appendix:demo}

An interactive demo, code, model weights, and the MedScribble dataset are  available at 
\begin{center}
    \url{https://scribbleprompt.csail.mit.edu}
\end{center}

\section{ScribblePrompt Implementation}
\label{appendix:implementation}

\subsection{Prompt Simulation}

In this section, we provide illustrations of the prompt simulation process. Each of these click and scribble simulation algorithms can be applied to the ground truth label (or false negative error region) to simulate positive clicks/scribbles and to the background (or false positive error region) to simulate negative clicks/scribbles. 

\subsubsection{Scribbles}
\label{appendix:scribble_simulation}

We simulate diverse and varied scribbles by first generating clean scribbles using one of three methods: (i) line scribbles, (ii) centerline scribbles or (iii) contour scribbles. Then, we break up and warp the scribbles to add more variability to account for human error. 

\begin{figure}
    \centering
    \includegraphics[width=\linewidth]{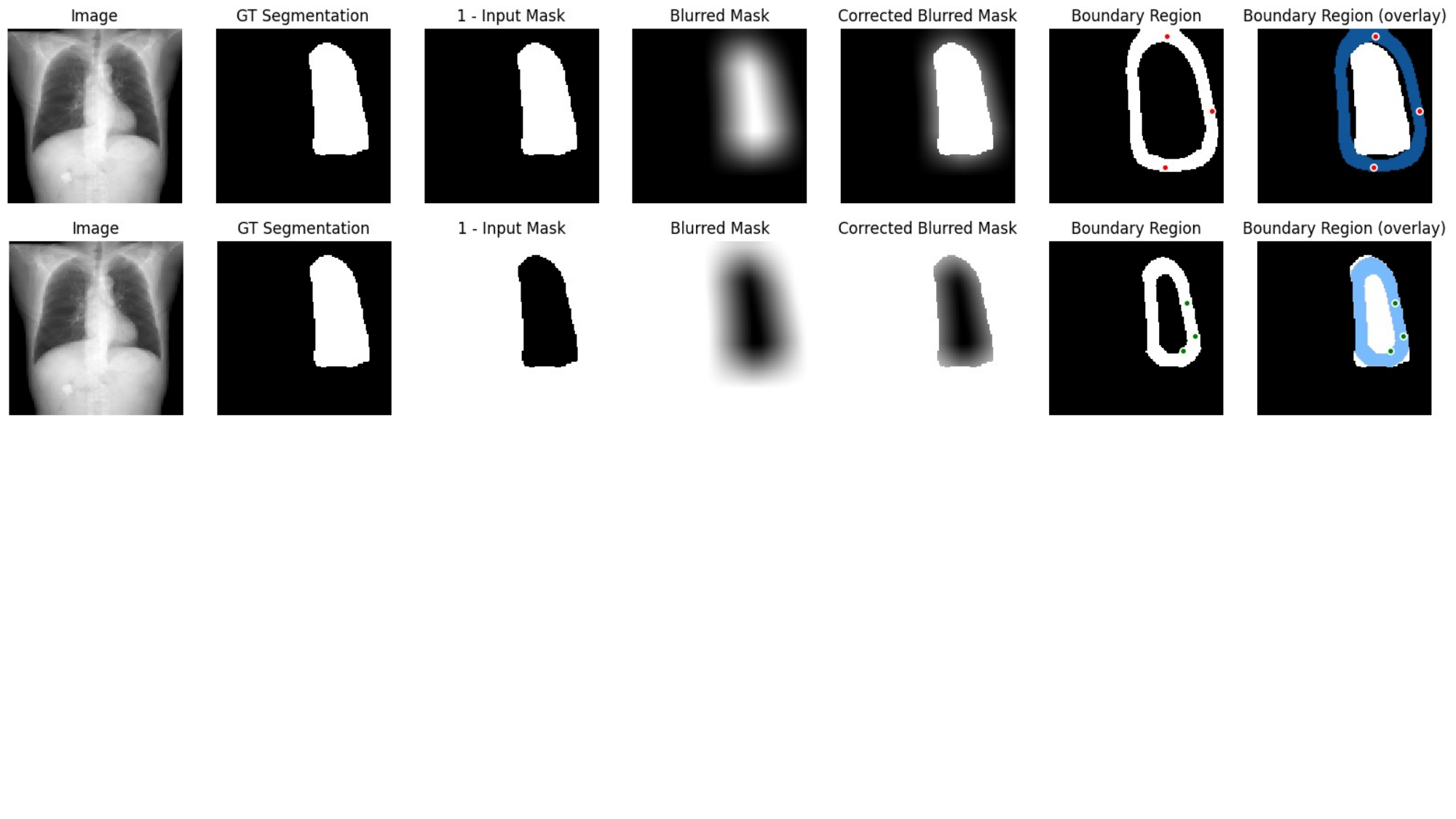}
    \caption{\textbf{Line scribbles}. Given an input mask $z$, we draw random lines by sampling two end points from $\{(u,v)| z_{uv}=1\}$. We use a random deformation field to warp the line scribbles and then multiply by the binary input mask $z$ to correct parts of the scribble that were warped outside the mask. We can simulate positive scribbles by applying the algorithm to the ground truth label $y$ (top) and negative scribbles by applying the algorithm to the background $1-y$ (bottom).}
    \label{fig:line_scribbles}
\end{figure}

\para{Line Scribbles}
\cref{fig:line_scribbles} illustrates the process of simulating line scribbles.

\begin{figure}
    \centering
    \includegraphics[width=\linewidth]{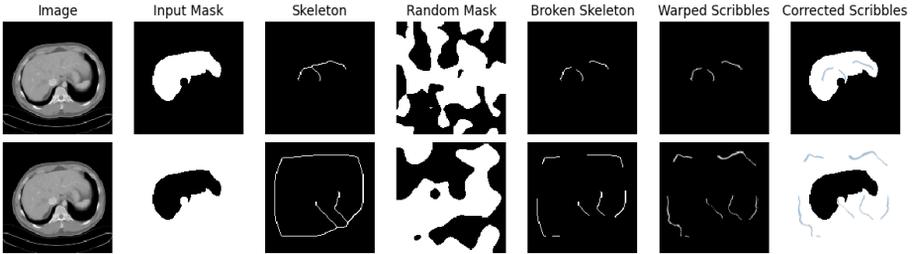}
    \caption{\textbf{Centerline scribbles}. Given an input mask, we apply a thinning algorithm \cite{zhangsuen_thinning_1984} to get a 1-pixel wide skeleton. We break up the skeleton using a random mask and use a random deformation field to warp the broken skeleton. Lastly, we multiply the scribble mask by the input binary mask to remove parts of the scribble that were warped outside the input mask. We can simulate positive scribbles by applying the algorithm to the label $y$ (top) and negative scribbles by applying the algorithm to the background $1-y$ (bottom).
    }
    \label{fig:centerline_scribbles}
\end{figure}

\para{Centerline Scribbles}
\cref{fig:centerline_scribbles} illustrates the process of simulating centerline scribbles. 

\begin{figure}
    \includegraphics[width=\linewidth]{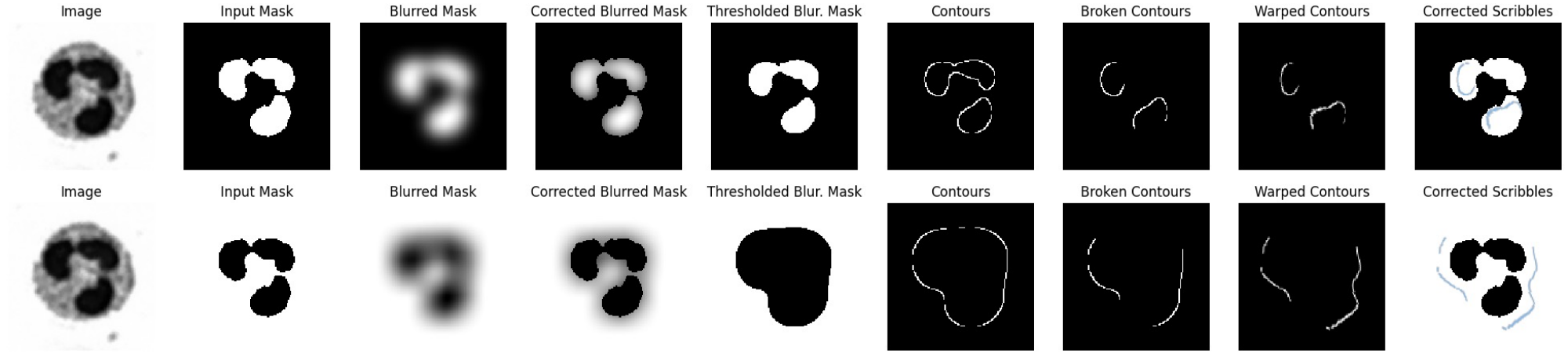}
    \caption{\textbf{Contour scribbles.} 
    We simulate a rough contour of the desired segmentation within the boundaries of the label. 
    Given a mask $z$, We first blur the mask to reduce the size of the label such that $\Tilde{z} = \min(z, z \circ G_k)$, where $G_k$ is a Gaussian blur kernel. Then we apply a threshold $\Tilde{z} < h$ sampled in some intensity range $h \sim U[\Tilde{z}_{min}, \Tilde{z}_{max}]$ and extract a contour inside the boundary of the mask. We break up the contour using a random mask and use a random deformation field to warp the broken contour. Lastly, we multiply the scribble mask by the input binary mask to correct parts of the scribble that were warped outside the mask. We can simulate positive scribbles by applying the algorithm to the label $y$ (bottom) and negative scribbles by applying the algorithm to the background $1-y$ (top).}
    \label{fig:contour_scribbles}
\end{figure}

\para{Contour Scribbles}
\cref{fig:contour_scribbles} illustrates the process of simulating contour scribbles. 

\begin{figure}
    \centering
    \includegraphics[width=\linewidth]{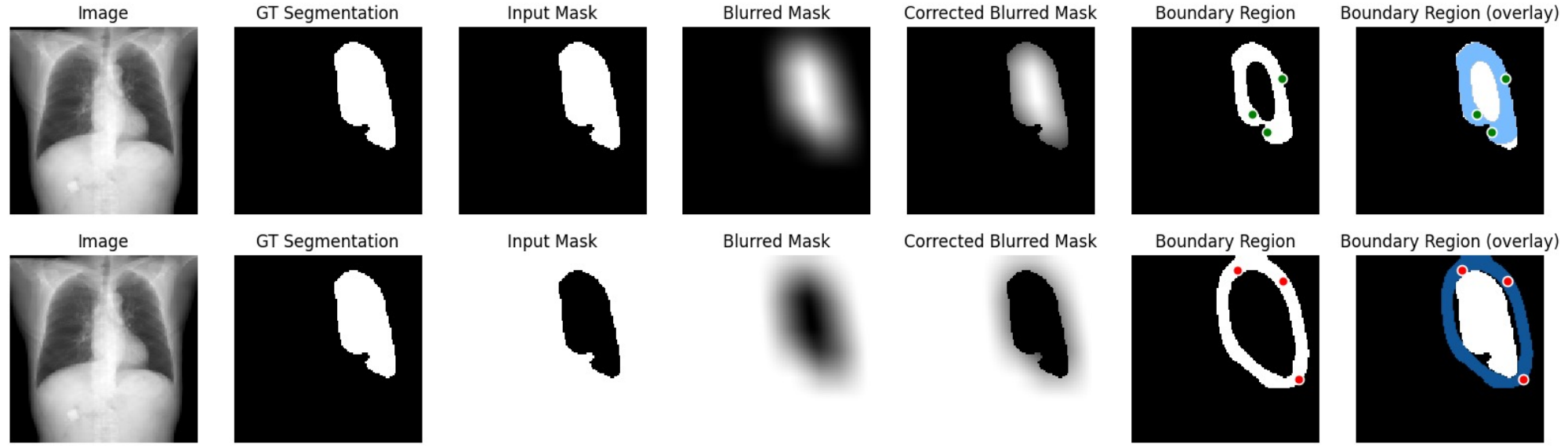}
    \caption{\textbf{Interior border region clicks.} We sample clicks from a border region inside the boundary of a given mask. 
    Given a mask $z$, we first blur the mask to reduce the size of the label such that $\Tilde{z} = \min(z, z \circ G_k)$ where $G_k$ is a Gaussian blur kernel. We then sample click coordinates from $\{(u,v) | \Tilde{z}_{uv} \in [a,b] \}$, where $a,b \sim U[\Tilde{z}_{min}, \Tilde{z}_{max})$ are thresholds sampled in some intensity range. We show the simulation process for negative border region clicks on the background $1-y$ (top) and positive border region clicks on the label $y$ (bottom).}
    \label{fig:interior_border_clicks}
\end{figure}

\para{Interior Border Region Clicks}
\cref{fig:interior_border_clicks} illustrates the process for simulating interior border region clicks. 

\subsection{Architecture and Training}
\label{appendix:architecture_ablations}

We discuss some of the modeling decisions in \mbox{ScribblePrompt-UNet} and \linebreak\mbox{ScribblePrompt-SAM}.

\begin{figure}
    \centering
    \includegraphics[width=\linewidth]{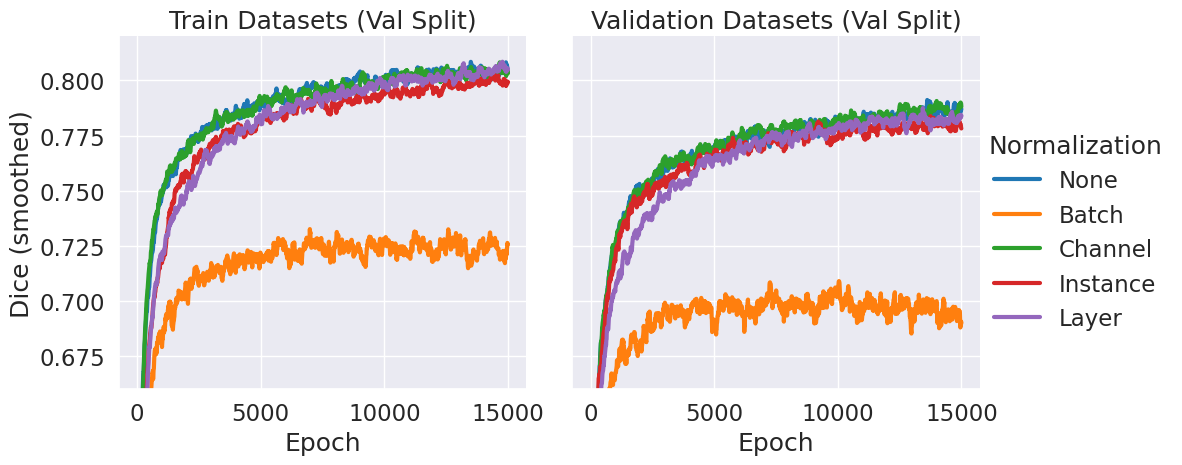}
    \caption{
    \textbf{Training ScribblePrompt-UNet with different normalization layers.}
    We show mean Dice averaged across five iterative predictions (using the training procedure for simulating interactions). At each epoch, we evaluate on 1,000 randomly sampled examples from the validation splits of the 65 training datasets and validation splits of the nine validation datasets. Dice was smoothed using Exponential Weighted Mean with $\alpha=0.1$.
    }
    \label{fig:varying_norm}
\end{figure}

\para{Normalization Layers}
In preliminary experiments, we evaluated normalization layers in the ScribblePrompt-UNet architecture such as Batch Norm~\cite{batchnorm}, Instance Norm~\cite{InstanceNorm}, Layer Norm~\cite{LayerNorm}, and Channel Norm~\cite{GroupNorm}. Including normalization did not improve the mean Dice on validation data compared to using no normalization layers (\cref{fig:varying_norm}).

\begin{figure}
    \centering
    \includegraphics[width=\linewidth]{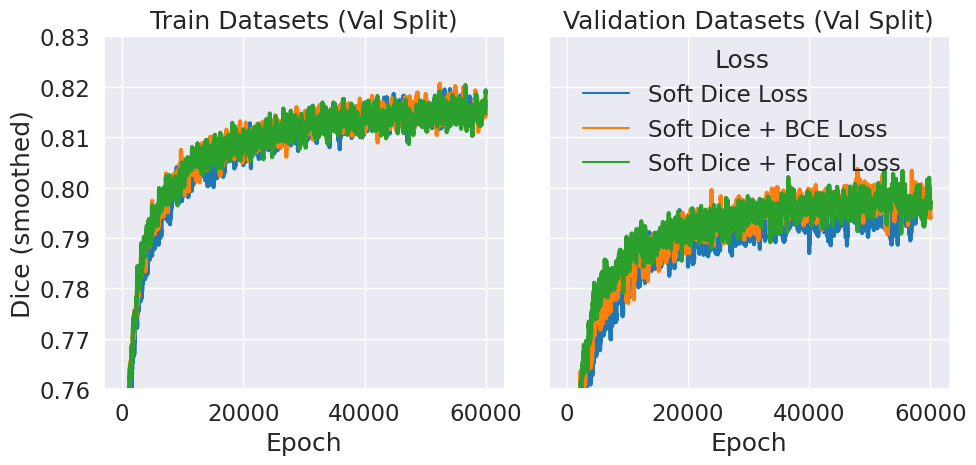}
    \caption{\textbf{Training ScribblePrompt-UNet with different loss functions.} We report Dice averaged across five iterative predictions (using the training procedure for simulating interactions). At each epoch, we evaluate on 1,000 randomly sampled examples from the validation splits of the 65 training datasets and validation splits of the nine validation datasets. Dice was smoothed using Exponential Weighted Mean with $\alpha=0.1$.}
    \label{fig:varying_loss}
\end{figure}

\para{Loss Function}
In preliminary experiments, we trained ScribblePrompt with Soft Dice Loss~\cite{dice1945measures}, a combination of Soft Dice Loss  and Binary Cross-Entropy Loss, and a combination of Soft Dice Loss and Focal Loss~\cite{focal_loss}, similar to \cite{SAM}. In the latter two losses, Dice Loss and BCE Loss or Focal Loss are weighted equally. We found that the combination of Soft Dice Loss and Focal Loss resulted in slightly higher mean Dice on the validation data for ScribblePrompt-UNet and ScribblePrompt-SAM. \cref{fig:varying_loss} shows Dice recorded during training in preliminary experiments with ScribblePrompt-UNet.

\para{ScribblePrompt-UNet Inputs}
We encode each prompt type in an input channel for ScribblePrompt-UNet. The input to ScribblePrompt-UNet has size $5 \times h \times w$ consisting of the input image $x^t$, bounding box encoding, positive click/scribble encoding, negative click/scribble encoding, and the logits of the previous prediction $\hat{y}^t_{i-1}$. For the first prediction, we set the previous prediction channel to zeros. We encode bounding boxes in a binary mask that is 1 inside the box(es) and 0 everywhere else. We encode positive and negative clicks using binary masks where a pixel is 1 if has been clicked and 0 otherwise. We encode positive and negative scribbles as masks on $[0,1]$ and combine them with the masks encoding clicks. Representing the interactions as masks is advantageous because inference time does not scale with the number of interactions. 

\para{ScribblePrompt-SAM Details}
To train ScribblePrompt-SAM, we took the pre-trained weights from SAM~\cite{SAM} with ViT-b backbone and froze all components of the network except for the decoder.

The SAM architecture can make predictions in single-mask mode or multi-mask mode. In \emph{single-mask mode}, the decoder outputs a single predicted segmentation given an input image and user interactions. In \emph{multi-mask mode}, the decoder predicts three possible segmentations and then outputs the segmentation with the highest predicted IoU by a MLP. We trained and evaluated ScribblePrompt-SAM in multi-mask mode to maximize the expressiveness of the architecture. During training we included a MSE term in the segmentation loss to train the MLP to predict the IoU of the predictions, as in \cite{SAM}.

\subsection{Synthetic Labels}
\label{appendix:synth}

To help reduce task overfitting -- memorizing the segmentation task for single-label datasets and thus ignoring user prompts -- we introduce a mechanism to generate synthetic labels. During training, for a given sample $(x_0, y_0)$, with probability $p_{synth}$ we replace $y_0$ with a synthetic label $y_{synth}$. 

\begin{figure}
    \centering
    \includegraphics[width=\linewidth]{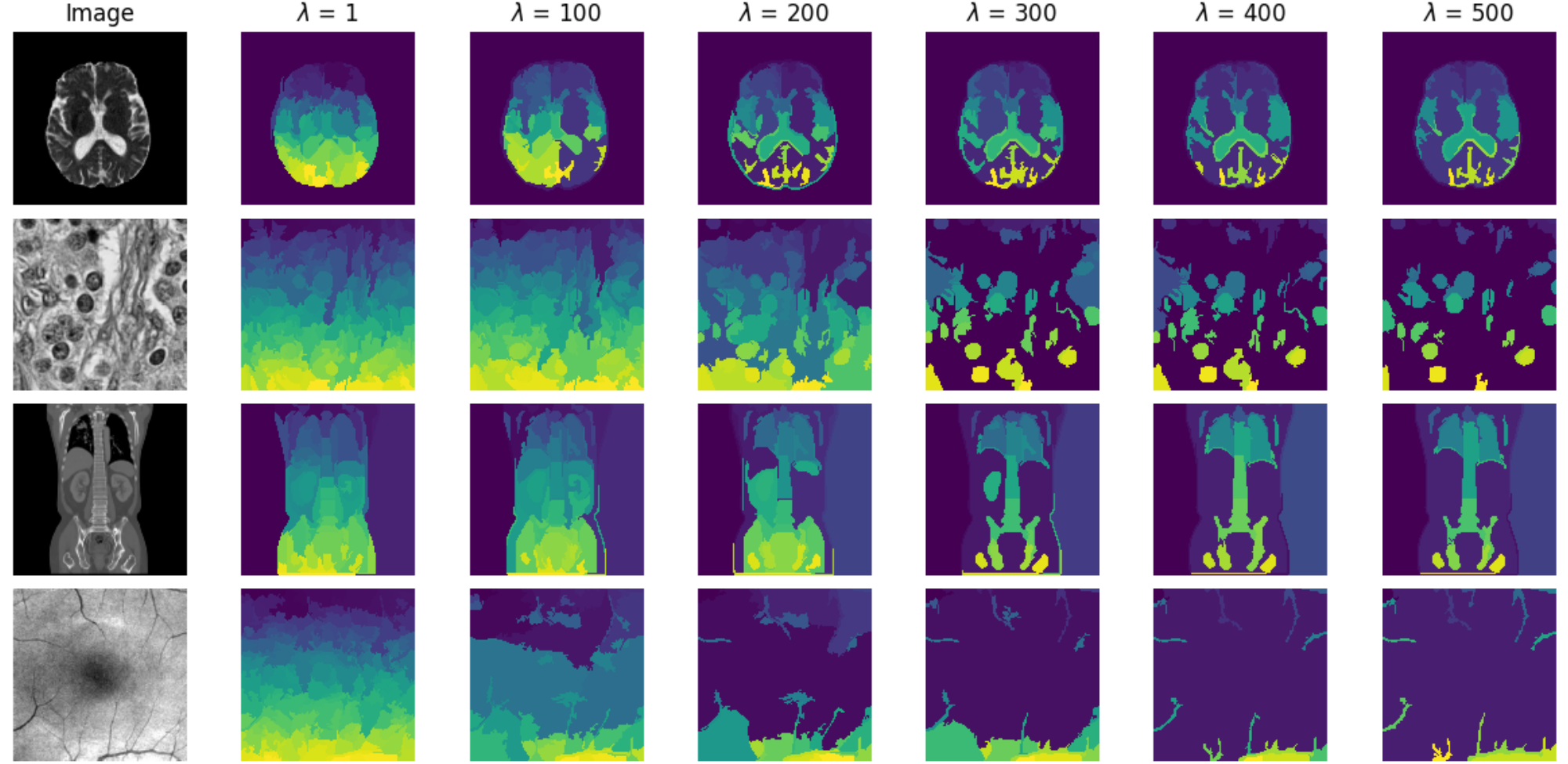}
    \caption{\textbf{Examples of possible synthetic labels}. Each color in the maps is a different synthetic label. During training, we replace a given label $y_0$ with a synthetic label $y_{synth}$ with probability $p_{synth}$. To generate $y_{synth}$, we apply a superpixel algorithm with randomly sampled scale parameter $\lambda$ to the image $x_0$ and then randomly select a superpixel as the synthetic label. We show examples of the synthetic label maps generated using a superpixel algorithm \cite{felzenszwalb_superpixels} with different $\lambda$.}
    \label{fig:superpixel_examples}
\end{figure}

We use a superpixel algorithm \cite{felzenszwalb_superpixels} with randomly sampled scale parameter $\lambda \sim U[1,500]$ to partition the image $x_0$ into a map of $k$ superpixels, $z \in \{1, \dots, k\}^{n \times n}$. Then, we randomly select a superpixel $c \sim \text{Cat}(\{1, \dots, k\}, 1/k)$ as the synthetic label $y_{synth} := \mathbbm{1}[z = c]$. \cref{fig:superpixel_examples} shows examples of training images and the corresponding maps of possible synthetic labels with different $\lambda$.

\clearpage
\section{Data}
\label{appendix:data}

We build on large dataset gathering efforts like MegaMedical~\cite{universeg, tyche} to compile a collection of 77 open-access biomedical imaging datasets for training and evaluation, covering over 54k scans, 16 image types, and 711 labels.
We gathered datasets with a particular focus on Microscopy, X-Ray, and Ultrasound modalities, which were not as well represented in the original MegaMedical~\cite{universeg}. The full list of datasets is provided in \cref{tab:eval_datasets} and \cref{tab:train_datasets}.

We define a 2D segmentation task as a combination of (sub)dataset, axis (for 3D modalities), and label. For datasets with multiple segmentation labels, we consider each label separately as a binary segmentation task. For datasets with sub-datasets (e.g., malignant vs. benign lesions) we consider each cohort as a separate task. For multi-annotator datasets, we treat each annotator as a separate label. For instance segmentation datasets, we sampled one instance at a time during training. 

For 3D modalities we use the slice with maximum label area (``maxslice'') and the middle slice (``midslice'') for each volume for training of ScribblePrompt. We report results evaluating on maxslices, but we observed similar trends evaluating on midslices.

\para{Division of Datasets}
The division of datasets and subjects for training, model selection, and evaluation is summarized in \cref{tab:data_split}. The 77 datasets were divided into 65 training datasets (Table \ref{tab:train_datasets}, 12 evaluation datasets. Data from 9 (out of 12) of the evaluation datasets was used in model development for model selection, and final evaluation. The other 3 evaluation datasets were completely held-out from model development and only used in the final evaluation. 

\para{Division of Subjects}
We split each of the 77 datasets into 60\% train, 20\% validation, and 20\% test by subject. We used the ``train'' splits from the 65 training datasets to train ScribblePrompt models. We use the ``validation'' splits from the 65 training datasets and 9 validation datasets for model selection. We report final evaluation results across 12 evaluation sets consisting of the ``test'' splits of the 9 validation datasets \emph{and} ``test'' splits of the 3 test datasets to maximize the diversity of tasks and modalities in our evaluation set (\cref{tab:data_split}). No data from the 9 validation datasets or 3 test datasets were seen by ScribblePrompt models during training. For TotalSegmentator~\cite{TotalSegmentator}, we only evaluated on 20 examples per task due to the large number of tasks in the dataset. In total, the evaluation data cover 608 segmentation tasks. 

\para{Image Processing}
We rescale image intensities to [0,1]. For methods using the SAM architecture, we convert the images to RGB and apply the pixel normalization scheme in \cite{SAM}. 

\para{Image Resolution}
We resized images to 128x128 for training of ScribblePrompt. We used this resolution to reduce training time during model development and to be able to conduct more thorough experiments. The ScribblePrompt approach is not tied to a particular resolution.

We conducted the experiments with MedScribble and simulated interactions with $128^2$ size images. For the ACDC scribbles dataset and the user study we evaluated $256^2$ size images to test ScribblePrompt's performance at higher resolutions. Although the ScribblePrompt-UNet architecture can take variable size inputs, we found downsizing the image to $128^2$ for inference then upsampling the prediction to the input image size produced the highest Dice predictions. 

For each method we resize the image to the method's training image size before running inference. Although the SAM architecture takes input images of size $1024^2$ (or $256^2$ in the case of SAM-Med2D), the the network outputs predictions of size $256^2$ that are up-sampled to the input image size. MIDeepSeg takes $96^2$ size images as inputs (after automatic cropping) and outputs predictions of size $96^2$.

\para{Interactive Baselines} SAM-Med2D used three of our evaluation datasets (ACDC~\cite{ACDC}, BTCV~\cite{BTCV} and TotalSegmentator~\cite{TotalSegmentator}) as training datasets \cite{sammed2d-data}. MedSAM used two of our evaluation datasets (TotalSegmentator~\cite{TotalSegmentator} and BUID~\cite{BUID}) as training datasets~\cite{MedSAM}.

\para{Supervised Baselines} We trained fully-supervised baselines for 10 of our evaluation datasets. For those datasets, We used the train and validation splits to train a fully-supervised nnUNet~\cite{isensee_nnunet_2021} for each 2D task (\cref{tab:data_split}). We report final results for all methods on the test splits of the evaluation datasets. 

\begin{table}
    \footnotesize
    \centering
    \rowcolors{2}{white}{gray!15}
    \caption{\textbf{Dataset split overview}. Each dataset was split into 60\% train, 20\% validation and 20\% test by subject. Data from the ``train'' splits of the 65 training datasets were used to train the models. The ScribblePrompt models did not see any data from the validation datasets or test datasets during training. Data from the ``validation'' split of the 9 validation datasets was used for ScribblePrompt (\firstone{SP}) model selection and baseline model selection (e.g., single-mask vs. multi-mask mode for SAM). We report final results on 12 ``evaluation sets'': data from the ``test'' splits of the 9 validation datasets and the ``test'' splits of the 3 test datasets. To train the fully-supervised \secondone{nnUNet} baselines, we used the training and validation splits of the 12 evaluation datasets.}
    \label{tab:data_split}
    \resizebox{\textwidth}{!}{
    \begin{tabular}{cc||p{3.cm}|p{4cm}|p{2.7cm}}
        & & \multicolumn{3}{c}{Split within each dataset by subject}
        \\
        \toprule
        Dataset Group & No. Datasets & {\centering Training Split (60\%)} & Validation Split (20\%) & Test Split (20\%)  \\
        \midrule
        Training Datasts & 65 & \firstone{SP} training & \firstone{SP} model selection & Not used  \\
        \hline
        Validation Datasets & 9 & \secondone{nnUNet} training & \firstone{SP} and baselines model selection, \secondone{nnUNet} training & Final evaluation 
        \\ 
        \hline
        Test Datasets & 3 & \secondone{nnUNet} training & \secondone{nnUNet} training & Final evaluation \\
        \bottomrule
    \end{tabular}
    }  
\end{table}

\begin{table}
\caption{
\textbf{Validation and test datasets}. We assembled the following set of datasets to evaluate ScribblePrompt and baseline methods. For the relative size of datasets, we include the number of unique scans (subject and modality pairs) that each dataset has. These datasets were unseen by ScribblePrompt during training. Three \firstone{test datasets} were completely held-out from model selection and development. The validation splits of the other 9 (validation) datasets were used for model selection. We report final results on the test splits of these 12 datasets.  
}
\label{tab:eval_datasets}
\centering
\resizebox{\textwidth}{!}{
\rowcolors{2}{white}{gray!15}
\begin{tabular}{p{3.7cm}p{7cm}ccp{3cm}}
    \toprule 
    \textbf{Dataset Name} & \textbf{Description} & \textbf{Scans} & \textbf{Labels} & \textbf{Modalities} 
    \\ 
    \midrule
    \ ACDC~\cite{ACDC} & {Left and right ventricular endocardium} & 99 & 3 & cine-MRI 
    \\
    \ BTCV Cervix~\cite{BTCV} & Bladder, uterus, rectum, small bowel & 30 & 4 & CT  
    \\
    \ BUID~\cite{BUID}& Breast tumors & 647 & 2 & Ultrasound 
    \\
    \firstone{COBRE~\cite{COBRE,fischl2012freesurfer,neurite}} & Brain anatomy & 258 & 45 & T1-weighted MRI
    \\
    \ DRIVE~\cite{DRIVE} & Blood vessels in retinal images & 20 & 1 & Optical camera
    \\
    \ HipXRay~\cite{HipXRay} & Ilium and femur & 140 & 2 & X-Ray 
    \\
    \ PanDental~\cite{PanDental} & Mandible and teeth & 215 & 2 & X-Ray 
    \\
    \ SCD~\cite{SCD} & Sunnybrook Cardiac Multi-Dataset Collection & 100 & 1 & cine-MRI 
    \\
    \firstone{SCR~\cite{SCR}} & Lungs, heart, and clavicles  & 247 & 5 & X-Ray
    \\
    \ SpineWeb~\cite{SpineWeb} & Vertebrae & 15 & 1 & T2-weighted MRI  
    \\
    \firstone{TotalSegmentator~\cite{TotalSegmentator}} & 104 anatomic structures (27 organs, 59 bones, 10 muscles, and 8 vessels) & 1,204 & 104 & CT 
    \\
    \ WBC~\cite{WBC} & White blood cell cytoplasm and nucleus & 400 & 2 & Microscopy 
    \\
    \bottomrule
\end{tabular}
}
\end{table}


\begin{table}

\caption{\textbf{Training datasets}. We assembled the following set of datasets to train ScribblePrompt. For the relative size of datasets, we have included the number of unique scans (subject and modality pairs) that each dataset has.}
\label{tab:train_datasets}
\centering
\rowcolors{2}{white}{gray!15}
\resizebox{\textwidth}{!}{
\begin{tabular}{p{3.7cm}p{7.5cm}cp{4cm}}
    \textbf{Dataset Name} & \textbf{Description} & \textbf{Scans} & \textbf{Modalities} 
    \\ 
    \toprule 
     AbdominalUS~\cite{AbdominalUS} & Abdominal organ segmentation & 1,543 & Ultrasound
     \\
     AMOS~\cite{AMOS} & Abdominal organ segmentation & 240 & CT, MRI 
     \\
     BBBC003~\cite{BBBC003} & Mouse embryos & 15 & Microscopy 
     \\
     BBBC038~\cite{BBBC038} & Nuclei instance segmentation & 670 & Microscopy 
     \\
     BrainDev~\cite{gousias2012magnetic, BrainDevFetal, BrainDevelopment, serag2012construction} & Adult and neonatal brain atlases & 53 & Multimodal MRI 
     \\
     BrainMetShare\cite{BrainMetShare} & Brain tumors & 420 & Multimodal MRI 
     \\
     BRATS~\cite{BRATS, bakas2017advancing, menze2014multimodal} & Brain tumors & 6,096 & Multimodal MRI 
     \\
     BTCV Abdominal~\cite{BTCV} & 13 abdominal organs & 30 & CT 
     \\  
     BUSIS~\cite{BUSIS} & Breast tumors & 163 & Ultrasound 
     \\
     CAMUS~\cite{CAMUS}  & Four-chamber and Apical two-chamber heart & 500 & Ultrasound
     \\
     CDemris~\cite{cDemris}  & Human left atrial wall & 60 & CMR 
     \\
     CHAOS~\cite{CHAOS_1, CHAOS_2}  & Abdominal organs (liver, kidneys, spleen) & 40 & CT, T2-weighted MRI 
     \\
     CheXplanation~\cite{CheXplanation} & Chest X-Ray observations & 170 & X-Ray
     \\
     CT2US~\cite{CT2US} & Liver segmentation in synthetic ultrasound & 4,586 & Ultrasound
     \\
     CT-ORG\cite{CT_ORG} & Abdominal organ segmentation (overlap with LiTS) & 140 & CT 
     \\
     DDTI~\cite{DDTI} & Thyroid segmentation & 472 & Ultrasound
     \\
     EOphtha~\cite{EOphtha} & Eye microaneurysms and diabetic retinopathy & 102 & Optical camera
     \\
     FeTA~\cite{FeTA} & Fetal brain structures & 80 & Fetal MRI 
     \\
     FetoPlac~\cite{FetoPlac} & Placenta vessel & 6 & Fetoscopic optical camera
     \\
     FLARE~\cite{FLARE21} & Abdominal organs (liver, kidney, spleen, pancreas) & 361 & CT 
     \\
     HaN-Seg~\cite{HaN-Seg} & Head and neck organs at risk & 84 & CT, T1-weighted MRI
     \\
     HMC-QU~\cite{HMC-QU, kiranyaz2020left} & 4-chamber (A4C) and apical 2-chamber (A2C) left  wall & 292 & Ultrasound 
     \\ 
     I2CVB~\cite{I2CVB} & Prostate (peripheral zone, central gland) & 19 & T2-weighted MRI 
     \\
     IDRID~\cite{IDRID} & Diabetic retinopathy & 54 & Optical camera 
     \\
     ISBI-EM~\cite{ISBI_EM} & Neuronal structures in electron microscopy & 30 & Microscopy
     \\
     ISIC~\cite{ISIC} & Demoscopic lesions & 2,000 & Dermatology
     \\
     ISLES~\cite{ISLES} & Ischemic stroke lesion & 180 & Multimodal MRI 
     \\
     KiTS~\cite{KiTS} & Kidney and kidney tumor & 210 & CT 
     \\
     LGGFlair~\cite{buda2019association, LGGFlair} & TCIA lower-grade glioma brain tumor & 110 & MRI 
     \\
     LiTS~\cite{LiTS} & Liver tumor & 131 & CT 
     \\
     LUNA~\cite{LUNA} & Lungs & 888 & CT 
     \\
     MCIC~\cite{MCIC} & Multi-site brain regions of schizophrenic patients & 390 & T1-weighted MRI
     \\
     MMOTU~\cite{MMOTU} & Ovarian tumors & 1,140 & Ultrasound
     \\
     MSD~\cite{MSD} & Large-scale collection of 10 medical segmentation datasets & 3,225 & CT, Multimodal MRI
     \\
     MuscleUS~\cite{MuscleUS} & Muscle segmentation (biceps and lower leg) & 8,169 & Ultrasound
     \\
     NCI-ISBI~\cite{NCI-ISBI} & Prostate & 30 & T2-weighted MRI 
     \\
     NerveUS~\cite{NerveUS} & Nerve segmentation & 5,635 & Ultrasound
     \\
     OASIS~\cite{OASIS-proc, OASIS-data} & Brain anatomy & 414 & T1-weighted MRI 
     \\
     OCTA500~\cite{OCTA500} & Retinal vascular & 500 & OCT/OCTA 
     \\
     PanNuke~\cite{PanNuke} & Nuclei instance segmentation & 7,901 & Microscopy
     \\
     PAXRay~\cite{PAXRay} & 92 labels covering lungs, mediastinum, bones, and sub-diaphram in Chest X-Ray & 852 & X-Ray 
     \\
    PROMISE12~\cite{Promise12} & Prostate & 37 & T2-weighted MRI
    \\
    PPMI~\cite{PPMI,dalca2018anatomical} & Brain regions of Parkinson patients & 1,130 & T1-weighted MRI
    \\
    QUBIQ~\cite{qubiq} & Collection of 4 multi-annotator datasets (brain, kidney, pancreas and prostate) & 209 & T1-weighted MRI, Multimodal MRI, CT
    \\
     ROSE~\cite{Rose} & Retinal vessel & 117 & OCT/OCTA 
     \\
     SegTHOR~\cite{SegTHOR} & Thoracic organs (heart, trachea, esophagus) & 40 & CT 
     \\
     SegThy~\cite{SegThy} & Thyroid and neck segmentation & 532 & MRI, Ultrasound
     \\
     ssTEM~\cite{ssTEM} & Neuron membranes, mitochondria, synapses and extracellular space & 20 & Microscopy
     \\
     STARE~\cite{STARE} & Blood vessels in retinal images (multi-annotator) & 20 & Optical camera 
     \\
     ToothSeg~\cite{ToothSeg} & Individual teeth & 598 & X-Ray
     \\
     VerSe~\cite{VerSe} & Individual vertebrae & 55 & CT
     \\
     WMH~\cite{WMH} & White matter hyper-intensities & 60 & Multimodal MRI 
     \\
     WORD~\cite{Word} & Abdominal organ segmentation & 120 & CT 
     \\
     \bottomrule
\end{tabular}
}
\end{table}

\clearpage
\section{Experimental Setup}
\label{appendix:setup}

\para{Training}
We use the Adam optimizer \cite{kingma2014adam} and train with a learning rate of $0.0001$ until convergence. We use a batch size of 8 for ScribblePrompt-UNet. For ScribblePrompt-SAM we use a batch size of 1, because of memory constraints.

\para{Task Diversity}
The final ScribblePrompt-UNet and ScribblePrompt-SAM models were trained with \mbox{$p_{synth}=0.5$}. \cref{tab:augmentations} shows the data augmentations we used, similar to the in-task augmentations from~\cite{universeg,tyche}.

\begin{table}
    \centering
    \caption{\textbf{Data augmentations during training.} For each example, an augmentation is sampled with probability $p$. We apply augmentations after (optional) synthetic label generation and before simulating user interactions.}
    \label{tab:augmentations}
    \rowcolors{2}{white}{gray!15}
    \begin{tabular}{lcc}
    \toprule
    \textbf{Augmentation} ~ & ~ $p$ ~ & ~ Parameters ~
    \\
    \midrule
    &  &  degrees $\in[0, 360]$ 
    \\
    \cellcolor{gray!15}  & \cellcolor{gray!15} & \cellcolor{gray!15} $\text{translation} \in [0, 0.2]$ 
    \\
    \multirow{-3}{*}{Random Affine} & \multirow{-3}{*}{ 0.5} & $\text{scale} \in [0.8, 1.1]$
    \\
     &  & $\text{brightness}\in[-0.1, 0.1]$ \\
     \multirow{-2}{*}{\cellcolor{white}{Brightness Contrast}} & \multirow{-2}{*}{\cellcolor{white}{0.5}} & \cellcolor{white} $\text{contrast} \in [0.8, 1.2]$ 
     \\
     \cellcolor{gray!15} & \cellcolor{gray!15} & \cellcolor{gray!15} $\sigma \in [0.1, 1.1]$ \\
    \multirow{-2}{*}{\cellcolor{gray!15}Gaussian Blur} & \multirow{-2}{*}{\cellcolor{gray!15} 0.5} & \cellcolor{gray!15} $k=5$ 
    \\
    \cellcolor{white} & \cellcolor{white} & \cellcolor{white} $\mu\in[0, 0.05]$
    \\
    \multirow{-2}{*}{\cellcolor{white}Gaussian Noise} & \multirow{-2}{*}{\cellcolor{white} 0.5} & \cellcolor{white} $\sigma \in [0, 0.05]$
    \\
    \cellcolor{gray!15} & \cellcolor{gray!15} & \cellcolor{gray!15} $\alpha\in[1, 2]$
    \\
    \multirow{-2}{*}{\cellcolor{gray!15}Elastic Transform} & \multirow{-2}{*}{\cellcolor{gray!15} 0.25} & \cellcolor{gray!15} $\sigma\in[6,8]$
    \\
    Sharpness & 0.5 & $\text{sharpness}=5$
    \\
    \cellcolor{gray!15}Horizontal Flip & \cellcolor{gray!15} 0.5 & \cellcolor{gray!15} None 
    \\
    Vertical Flip & \cellcolor{white} 0.5 & \cellcolor{white} None \\
    \bottomrule
    \end{tabular}
\end{table}

\para{SAM Baselines}
For baseline methods using the SAM architecture, we evaluate the models in both ``single mask'' and ``multi-mask'' mode. For each baseline method and interaction procedure, we selected the best performing mode based on the average Dice across the validation data and report final results on test data using that mode. In the results with simulated clicks and scribbles by dataset in \cref{appendix:simulated_clicks_scribbles}, we show results using both modes.
For ScribblePrompt-SAM and SAM-Med2D with adapter layers, multi-mask mode resulted in the highest Dice. For SAM-Med2D without adapter layers, we found multi-mask mode led to higher Dice for scribble inputs while single-mask mode led to higher Dice with click inputs. For SAM (ViT-b and ViT-h) and MedSAM, single-mask mode resulted in the higher Dice on average.

\clearpage
\section{Manual Scribbles}\label{appendix:manual_scribbles}

We provide additional setup details and visualizations for the manual scribbles evaluation in \cref{sec:manual_scribbles_experiment}.

\subsection{Setup}
\label{appendix:medscribble}

\para{MedScribble Dataset} We collected a diverse dataset of manual scribble annotations, which is available at \href{https://scribbleprompt.csail.mit.edu/data}{https://scribbleprompt.csail.mit.edu/data}. The MedScribble dataset contains annotations from 3 annotators for 64 image segmentation pairs. The examples were randomly selected from the validation split of 14 different datasets (7 training datasets and 7 validation datasets)~\cite{ACDC,AbdominalUS,BTCV,CAMUS,CHAOS_1,CHAOS_2,HipXRay,OASIS-data,OASIS-proc,OCTA500,PAXRay,PanDental,SCD,STARE,SpineWeb,WBC}.

For each task, the annotators were shown 5 training examples with the ground truth segmentation and instructed to draw positive scribbles on the region of interest and negative scribbles on the background for 3-5 new images (without seeing the ground truth segmentation). We collected the scribbles using a web app developed in Python using the Gradio library~\cite{gradio}. Two of the annotators used an iPad with stylus and one annotator used a laptop trackpad, to draw the scribbles. 

For the manual scribbles evaluation, we report results on a subset of MedScribble,  containing only examples from datasets unseen by ScribblePrompt during training. This subset contains 31 image-segmentation pairs (each with 3 sets annotations) covering 7 segmentations tasks from 7 different validation datasets~\cite{ACDC,SpineWeb,HipXRay,PanDental,SCD,WBC,BTCV}. The subset includes cardiac MRI, dental X-Ray, abdominal organ, spine vertebrae, and cell microscopy segmentation tasks.

\para{ACDC Scribbles Dataset} Like the other datasets we used, we split the ACDC dataset~\cite{ACDC} into 60\% train, 20\% validation and 20\% test by subject. We used the validation split for model selection for baseline methods (\eg single-mask vs. multi-mask mode for methods using the SAM architecture). We report results averaged across three labels on all slices for the test subjects. 

\para{MedSAM} We only evaluate MedSAM using bounding box prompts because it was fine-tuned exclusively with bounding box prompts and performs poorly with point inputs (\cref{fig:medsam_bad_examples}). We prompted MedSAM using a bounding box fit to the positive scribbles. For each dataset, we experimented with using the minimum enclosing bounding box or enlarging the box by 5 pixels in each direction and selected the settings that maximized Dice on the validation data. Using the minimum bounding box resulted in higher Dice scores for MedScribble and enlarging the bounding box resulted in in higher Dice scores for ACDC. 

\para{SAM} For methods using the SAM architecture (besides MedSAM), we converted the scribble masks to sets of positive and negative clicks for every non-zero pixel in the scribble masks.

\para{ScribblePrompt-UNet} For ScribblePrompt-UNet we found that blurring the scribble masks with a 3x3 Gaussian blur kernel with $\sigma=0.5$ prior to inference improved Dice scores, perhaps due to differences in the distribution of pixel values between the manually-collected scribbles and simulated scribbles during training. We also experimented with blurring the scribbles for ScribblePrompt-SAM and each of the baseline methods but it did not improve the Dice scores for any other methods.

\subsection{Results}

\begin{figure}
    \centering
    \includegraphics[width=\textwidth]{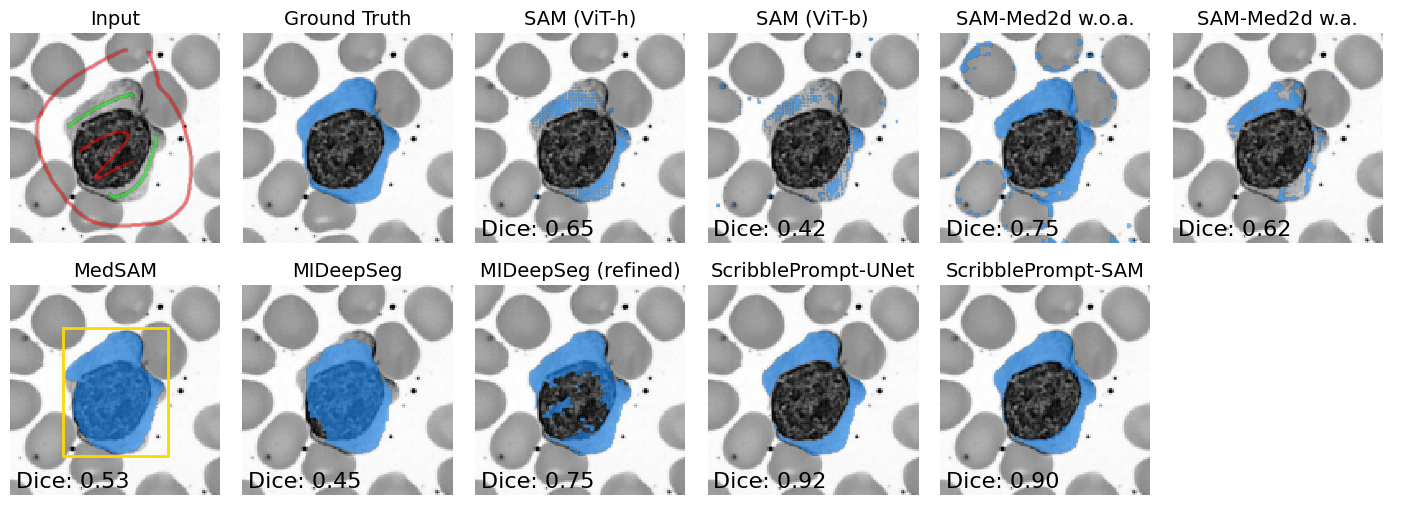}
    \includegraphics[width=\textwidth]{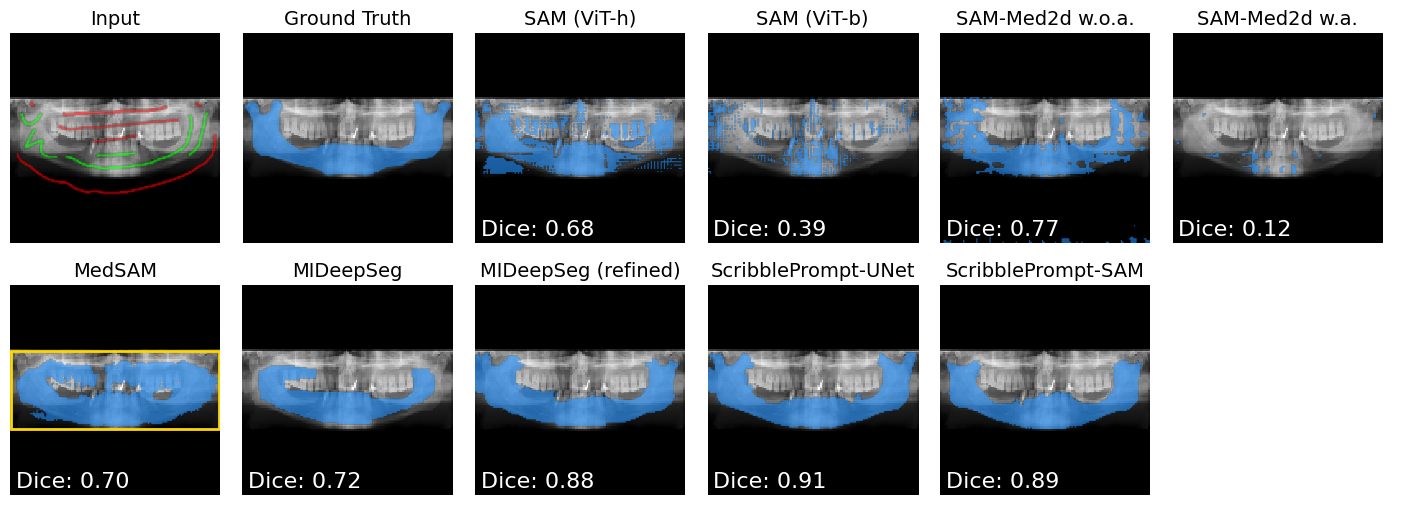}
    \includegraphics[width=\textwidth]{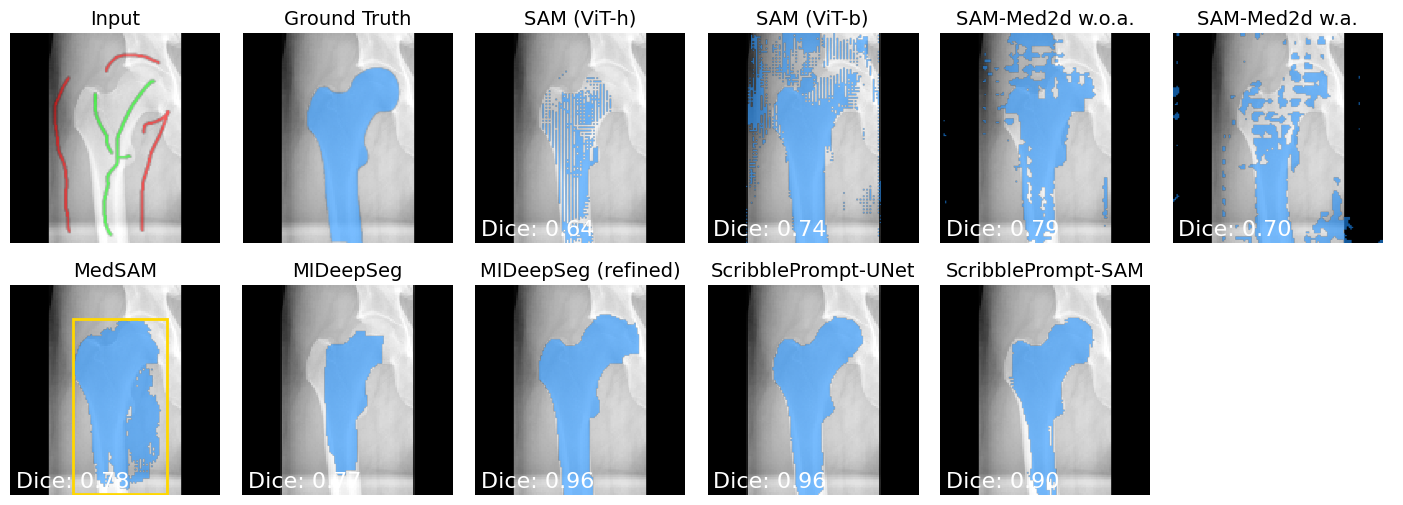}
    \includegraphics[width=\textwidth]{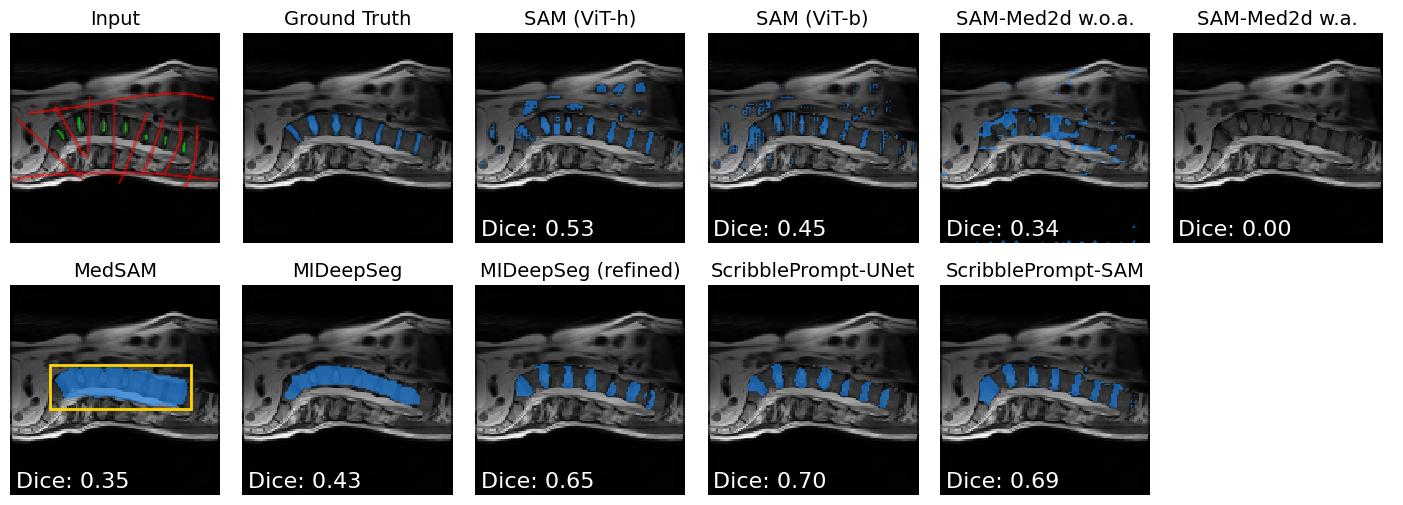}
    \caption{\textbf{Example predictions from MedScribble manual scribbles.} We evaluate on four examples from the MedScribble dataset. For each method, we show the \textcolor{eccvblue}{predicted segmentation} given a set of manually-collected \textcolor{ForestGreen}{positive} and \textcolor{red}{negative} scribbles as input. For MedSAM, we use a bounding box fit to the positive scribbles as the input.}
    \label{fig:medscribble_examples}
\end{figure}

\begin{figure}
    \centering
    \vspace{1cm}
    \includegraphics[width=\textwidth]{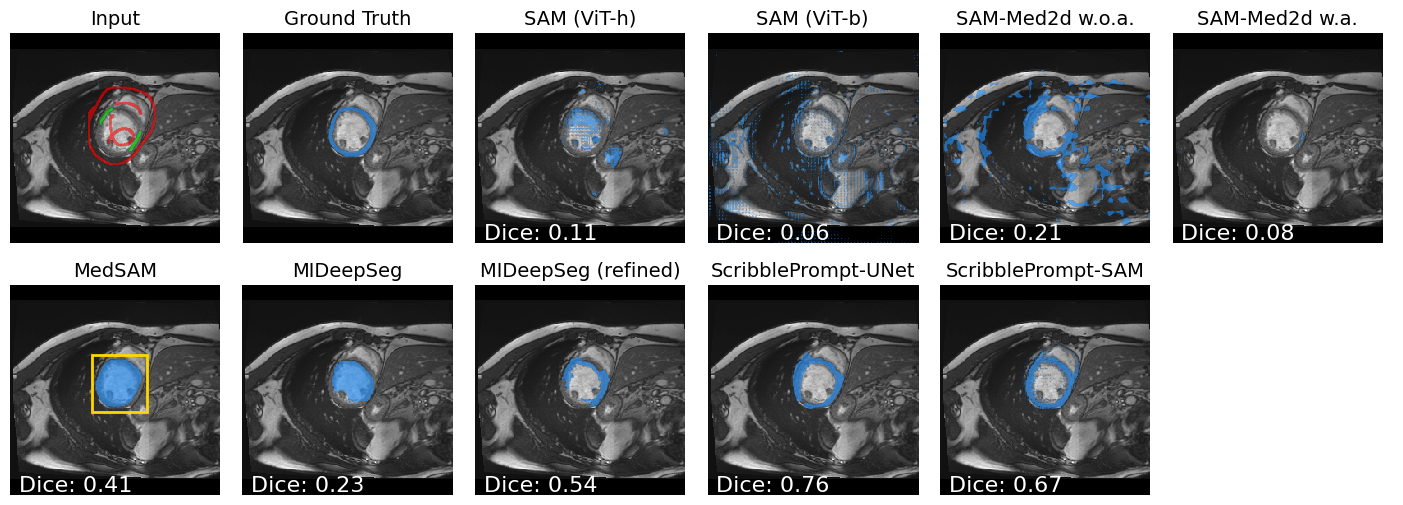}
    \includegraphics[width=\textwidth]{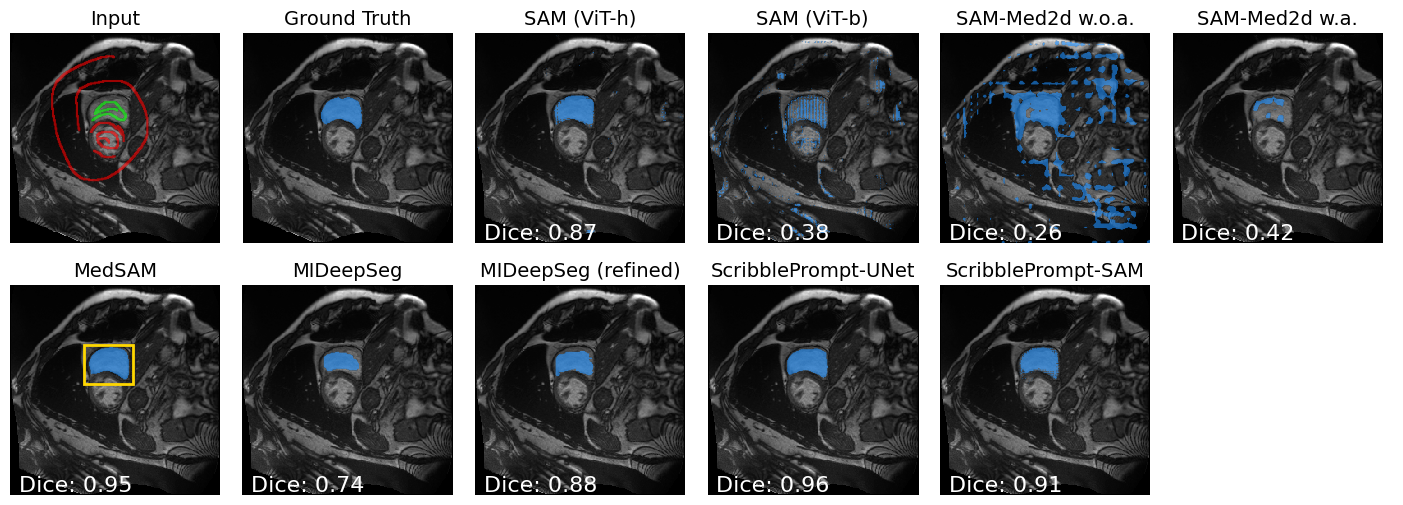}
    \includegraphics[width=\textwidth]{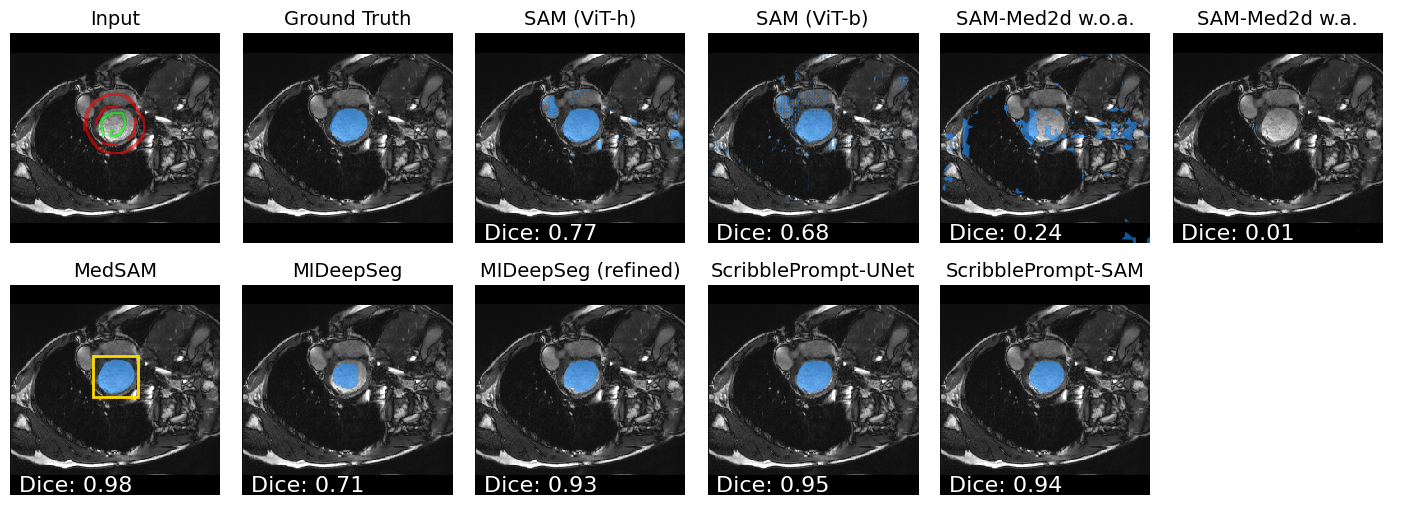}
    \caption{\textbf{Example predictions from ACDC manual scribbles.} We show examples for each label from the ACDC scribbles dataset~\cite{ACDC}. For each method, we show the \textcolor{eccvblue}{predicted segmentation} given a set of manually-collected \textcolor{ForestGreen}{positive} and \textcolor{red}{negative} scribbles as input. For MedSAM, we use a bounding box fit to the positive scribbles with 5 pixels added to each dimension as the input. Scribble thickenss is enlarged for visual clarity.}
    \label{fig:acdc_examples}
\end{figure}

\para{Visualizations}
\cref{fig:medscribble_examples} shows predictions for each method using examples from the MedScribble dataset. \cref{fig:acdc_examples} shows examples from the ACDC scribbles dataset.

\clearpage
\subsection{Comparison to Scribble-Supervised Learning}
\label{appendix:scribblesup}

We report preliminary results comparing ScribblePrompt to scribble-supervised learning. 
Scribble-supervised learning methods use scribble annotations as \emph{supervision} to train automatic segmentation models for predicting segmentation given only an input image~\cite{lin2016scribblesup, li_scribblevc_2023, zhang2022cyclemix, luo_scribble_2022, gotkowski2024embarrassingly}. These models are task-specific; a new model must be trained using scribble-supervised learning for each new task and training requires many scribble-annotated images from the same task to produce accurate results. 
In contrast, ScribblePrompt can perform new segmentation tasks at inference time without retraining, using scribbles as \textit{input}.

\para{Setup}
We compare ScribblePrompt-UNet to ScribFormer~\cite{li2024scribformer}, a recent state-of-the-art scribble-supervised learning method, on the ACDC scribbles dataset~\cite{ACDC}. Experiments reported in~\cite{li2024scribformer} show that ScribFormer's performance varies with the amount of training data, from 0.847 Dice given 14 training subjects to 0.894 Dice given 70 training subjects (and 15 validation subjects) from ACDC.

We evaluate each method given the same test data as in our manual scribbles evaluation: 20 subjects with scribble-annotations for three labels and background. For ScribFormer, we randomly partition the 20 test subjects into 80\% train and 20\% validation by subject, and train following~\cite{li2024scribformer}. We run inference for each model on all 20 test subjects, and report results averaged across the three labels for the 380 slices.

\para{Results}
\cref{tab:scribblesup} shows the difference in mean Dice between ScribblePrompt-UNet and ScribFormer is not statistically significant ($p=0.70$ with a paired t-test). Training ScribFormer required 2 hours using a Nvidia A100 GPU with 16 CPUs.

\begin{table}
    \centering
    \caption{\textbf{Comparison to scribble-supervised learning}. Mean Dice and HD95 with 95\% CI of predicted segmentations for ACDC ($n=1,140$).}
    \label{tab:scribblesup}
    \setlength{\tabcolsep}{0.5em}
    \begin{tabular}{lcc}
    \toprule
     & $\uparrow$ Dice Score & $\downarrow$ HD95 
    \\
    \midrule
    ScribFormer & $0.85 \pm 0.01$ & $4.05 \pm 0.99$
    \\
    \textbf{ScribblePrompt-UNet} & $0.84 \pm 0.01$ & $1.80 \pm 0.11$
    \\
    \bottomrule
    \end{tabular}
\end{table}

\para{Discussion}
Given limited scribble-annotated data from ACDC, ScribblePrompt-UNet predicts segmentations with similar Dice scores and lower HD95 compared to a scribble-supervised learning model trained on the data.

\clearpage
\section{Simulated Interactions}
\label{appendix:experiments}

We present additional results from the experiments in \cref{sec:simulated_experiment} with simulated interactions. 

\subsection{Bounding Boxes}
\label{appendix:bbox_results}

We evaluate models with simulated bounding box prompts.

\para{Setup}
We evaluate segmentation accuracy using Dice score after a single bounding box prompt. We simulate bounding boxes using the same procedure as used was used when training ScribblePrompt: we find the minimum enclosing bounding box for the ground truth label and then enlarge each dimension by $r \sim U[0,20]$ pixels to account for human error. We exclude MIDeepSeg~\cite{MIDeepSeg} from this evaluation because it is not designed to make predictions from a single bounding box input.

For methods using the SAM architecture, we apply the pixel normalization scheme in \cite{SAM} to images before inference. Upon further investigation, MedSAM~\cite{MedSAM} performed better with images rescaled to $[0,1]$; we report results for MedSAM with both normalization schemes.

\begin{figure}
    \centering
    \includegraphics[width=0.8\linewidth]{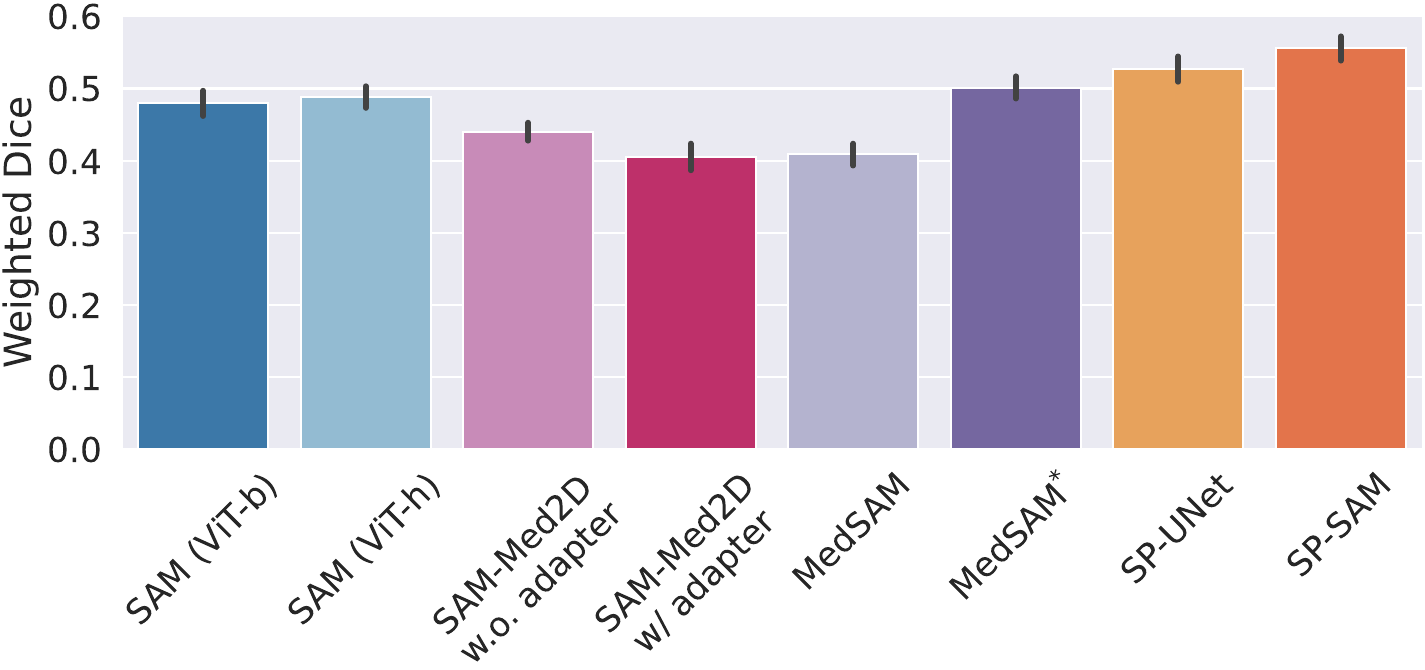}
    \caption{\textbf{Results with simulated bounding boxes}. Mean Dice on test data from 12 datasets with one simulated bounding box prompt, weighting each dataset equally. SP = ScribblePrompt. MedSAM$^*$ indicates MedSAM with input images re-scaled to $[0,1]$ instead of the pixel normalization from \cite{SAM}. Errorbars show 95\% CI from bootstrapping.}
    \label{fig:boxes}
\end{figure}

\begin{figure}
    \includegraphics[width=\linewidth]{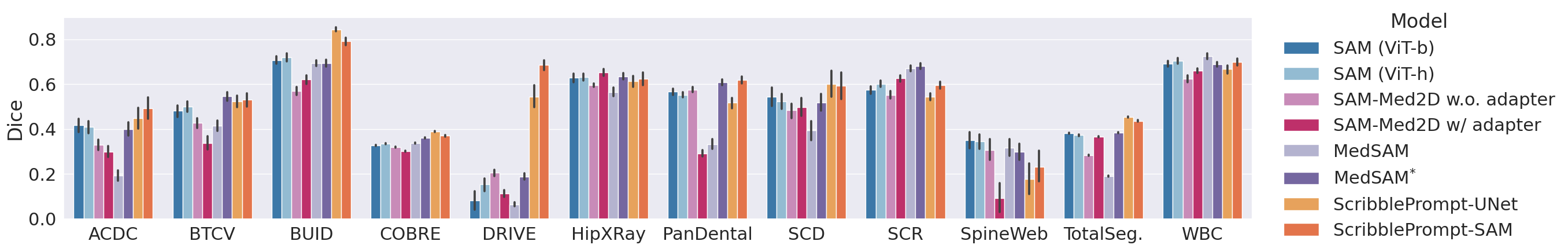}
    \caption{\textbf{Results with simulated bounding boxes by dataset.} Mean Dice after one simulated bounding box prompt. Among the evaluation datasets, bounding box prompts are the most effective for BUID, a breast ultrasound dataset. MedSAM$^*$ indicates MedSAM with input images re-scaled to $[0,1]$ instead of the pixel normalization from \cite{SAM}. Errorbars show 95\% CI from bootstrapping.}
    \label{fig:boxes_by_datasets}
\end{figure}

\para{Results}
\cref{fig:boxes} shows mean Dice after one bounding box prompt. \cref{fig:boxes_by_datasets} shows results by dataset. ScribblePrompt-SAM has the highest Dice on average after one bounding box prompt.

\begin{figure}
    \centering
    \includegraphics[width=0.8\linewidth]{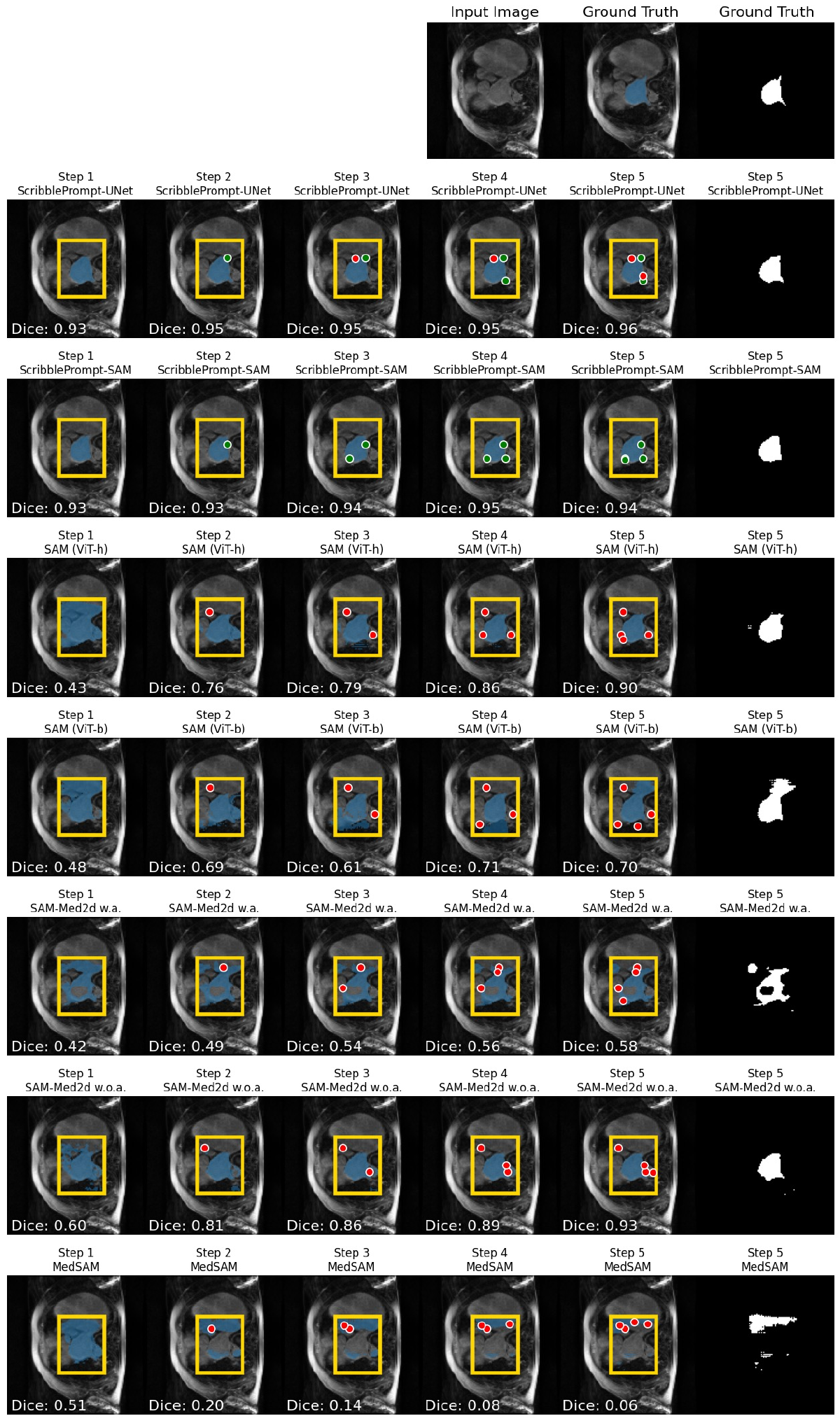}
    \caption{\textbf{Bounding box prompt with center correction clicks}. We simulate iterative interactive segmentation of the left ventricle in a cardiac MRI from the SCD dataset~\cite{SCD}. This label was seen during training but this dataset was not. ScribblePrompt models produce the highest dice predictions after a single bounding box prompt (first column) and are able to improve their predictions with additional corrections.}
    \label{fig:box_scd}
\end{figure}

\begin{figure}
    \centering
    \includegraphics[width=0.8\linewidth]{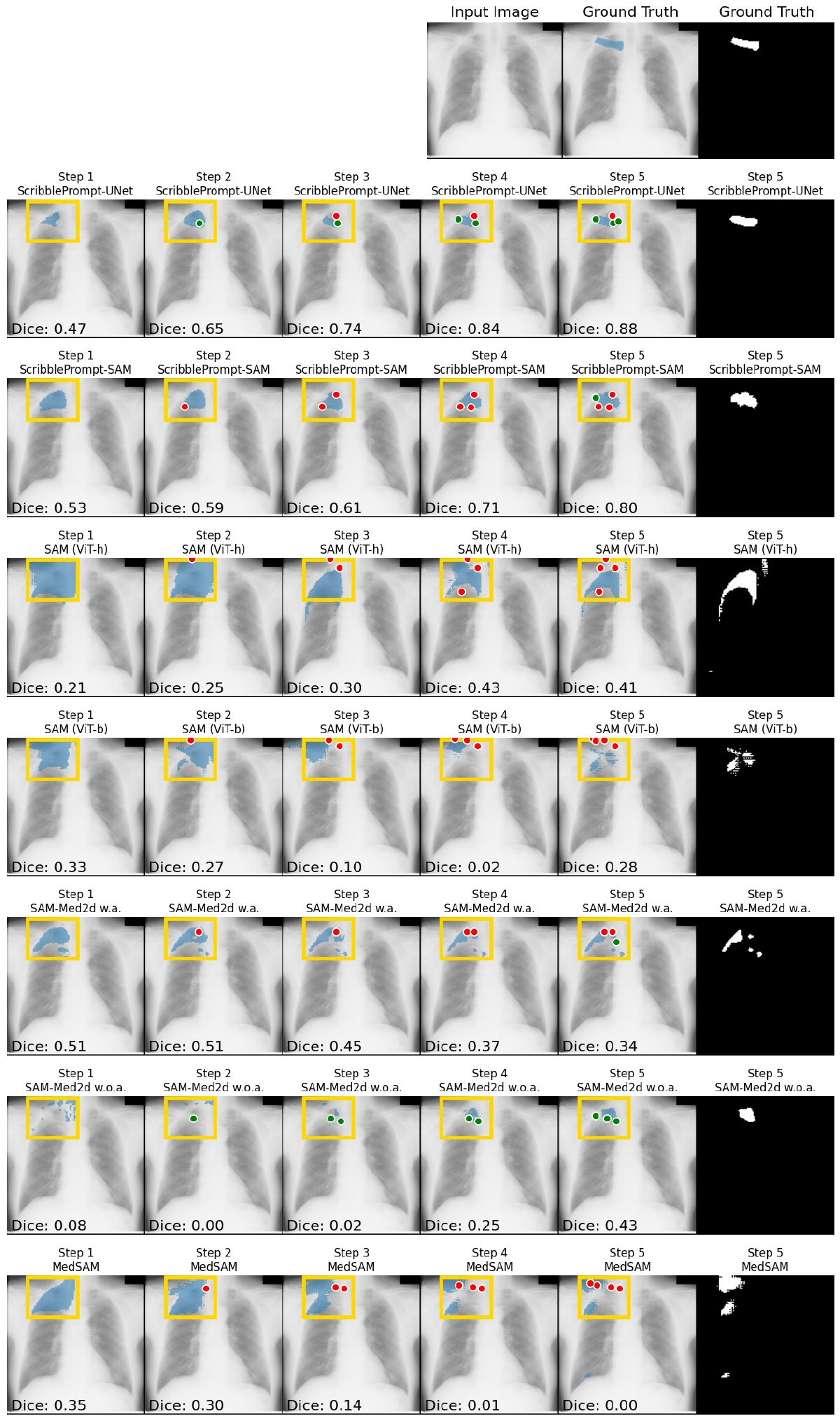}
    \caption{\textbf{Bounding box prompt with center correction clicks}. We show clavicle segmentation on an frontal chest X-Ray from the SCR dataset~\cite{SCR}. This dataset was completely held-out and this label was unseen during training. None of the methods are able to accurately segment the clavicle from a single bounding box prompt (first column). However, after a few correction clicks, ScribblePrompt-UNet and ScribblePrompt-SAM achieve 0.88 and 0.80 Dice, respectively.}
    \label{fig:box_scr}
\end{figure}

\para{Visualizations}
Due the ambiguity of many segmentation tasks, its often difficult to predict an accurate segmentation from a single bounding box prompt (\cref{fig:box_scr}). Although ScribblePrompt models produced the highest dice predictions from a single bounding box prompt in \cref{fig:boxes}, users may not be satisfied with this level of accuracy. Users can still achieve high Dice segmentations with ScribblePrompt by providing additional click and scribble interactions to correct the prediction. We visualize predictions for two examples in \cref{fig:box_scd} and \cref{fig:box_scr}, after a single bounding box prompt and after correction clicks. MedSAM has the highest mean Dice among the baselines after a single bounding box prompt (\cref{fig:boxes}), but its usability is limited because it cannot incorporate corrections. 

\clearpage
\subsection{Scribbles and Clicks}
\label{appendix:simulated_clicks_scribbles}

We provide additional setup details, baselines and results for the experiments with simulated scribbles and clicks presented in \cref{sec:simulated_experiment}.

\para{Setup} We evaluated each method following three scribble interaction procedures and three click interaction procedures. We provide details below on the MedSAM baseline and additional supervised baselines. 

\begin{figure}
    \centering
    \includegraphics[width=\linewidth]{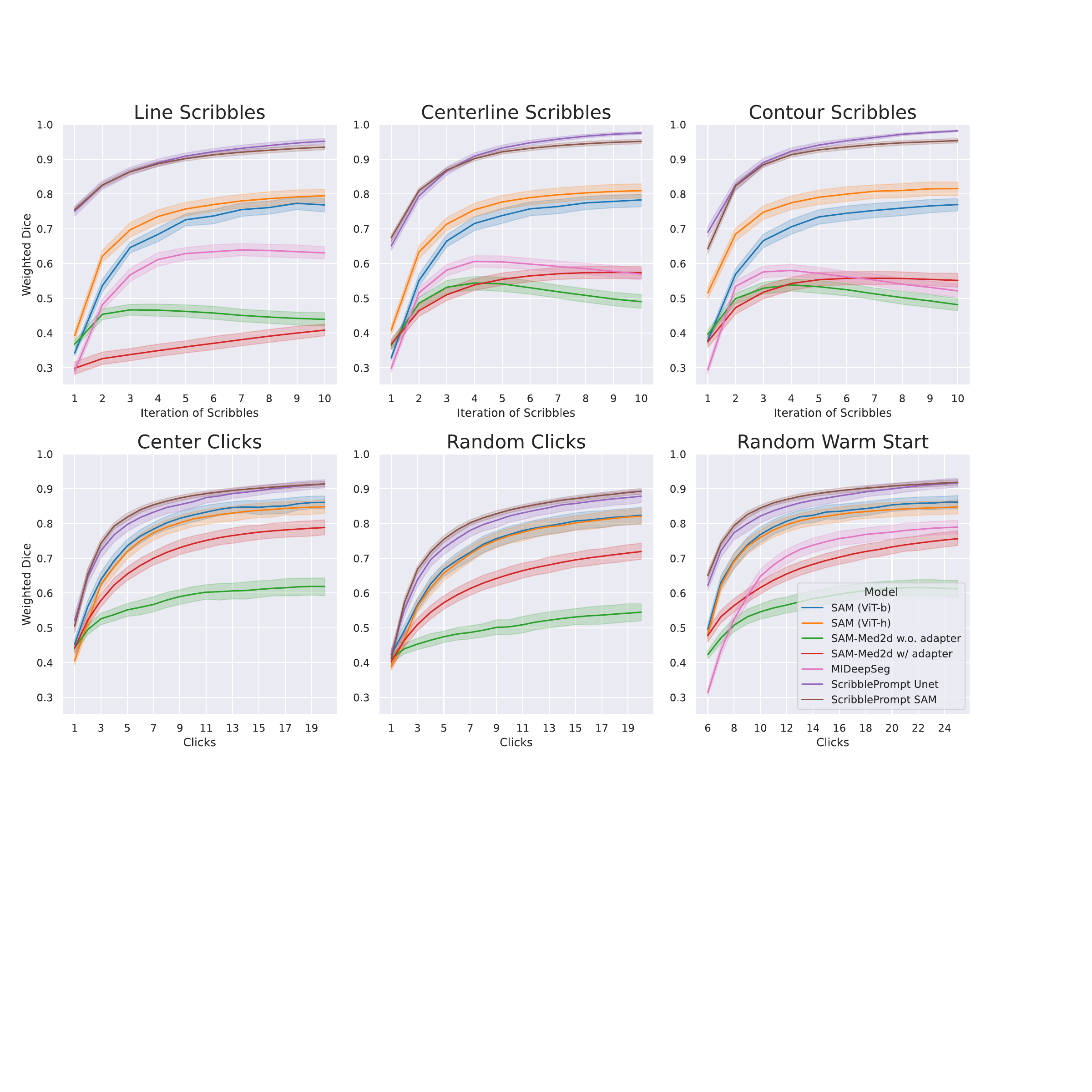}
    \caption{\textbf{Dice results with simulated scribbles and clicks}. We evaluate methods using three scribble procedures and three click procedures. We measure Dice averaged across twelve evaluation sets (the test splits of the nine validation and three test datasets), weighting each dataset equally. Shaded regions show 95\% CI from bootstrapping.}
    \label{fig:full_simulated_results}
\end{figure}

\begin{figure}
    \centering
    \includegraphics[width=\linewidth]{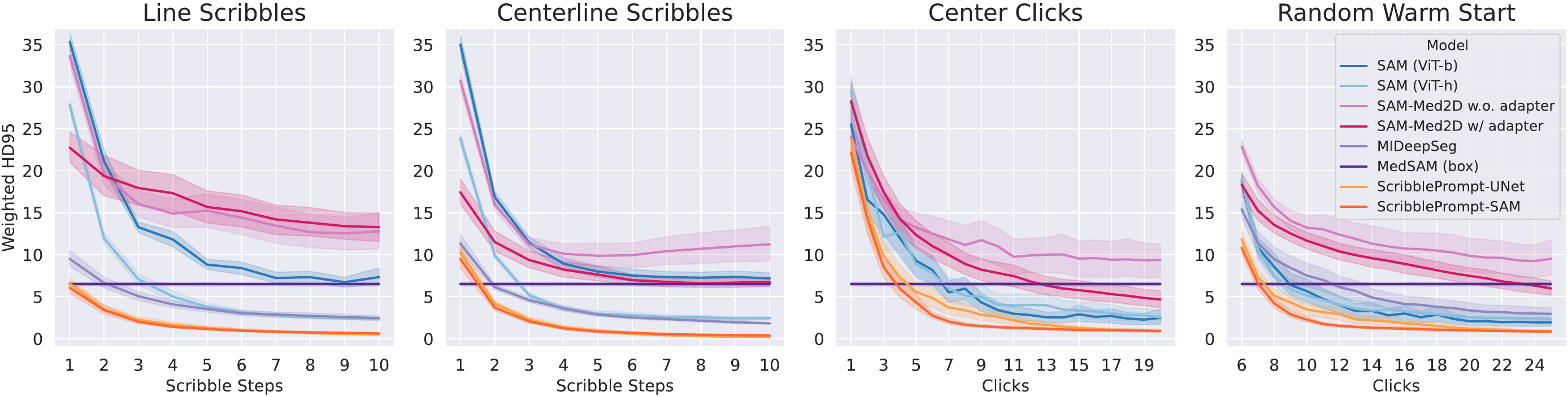}
    \caption{\textbf{HD95 results with simulated scribbles and clicks} We report HD95 for two scribble procedures and two click procedures. We measure HD95 averaged across twelve evaluation sets (the test splits of the nine validation and three test datasets), weighting each dataset equally. We exclude examples where the ground truth segmentation label was empty or the predicted segmentation was empty. Shaded regions show 95\% CI from bootstrapping.}
    \label{fig:full_simulated_results_hd}
\end{figure}

\begin{figure}
    \centering
    \includegraphics[width=\linewidth]{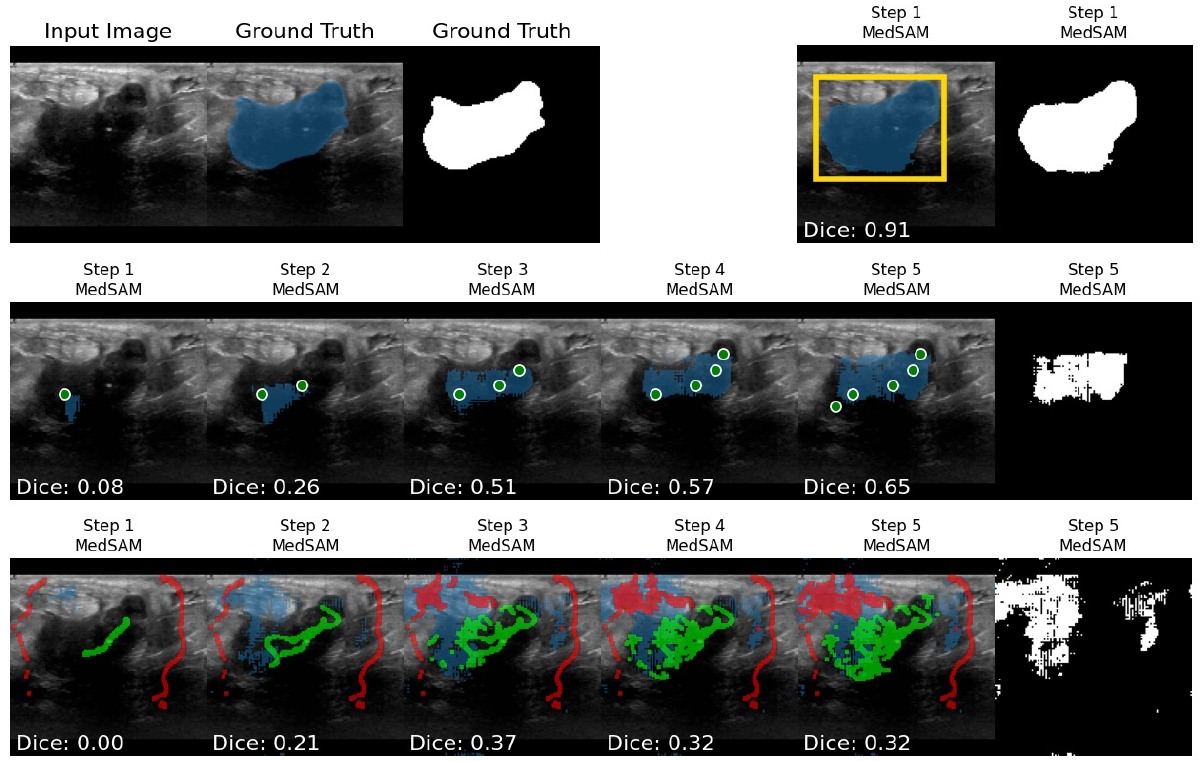}
    \caption{\textbf{MedSAM with bounding box, click, and scribble inputs.}
    We do not evaluate MedSAM with click and scribble inputs, which it was not trained for, because it produces poor segmentations with these inputs. Scribble thickness is enlarged for visual clarity.}
    \label{fig:medsam_bad_examples}
\end{figure}

\para{MedSAM} Since MedSAM~\cite{MedSAM} performs poorly with scribble and click prompts (\cref{fig:medsam_bad_examples}), we only evaluate it with bounding box prompts. We fit a bounding box to the ground truth segmentation and enlarged each dimension by $r \sim U[0,10]$ pixels, to match the amount of jitter used during training for MedSAM. We show the mean Dice of segmentations predicted by MedSAM from a single bounding box prompt as a horizontal line (\cref{fig:full_simulated_results}, \cref{fig:full_simulated_results_hd}) because MedSAM cannot incorporate corrections.

\begin{figure}
    \centering
    \includegraphics[width=0.9\linewidth]{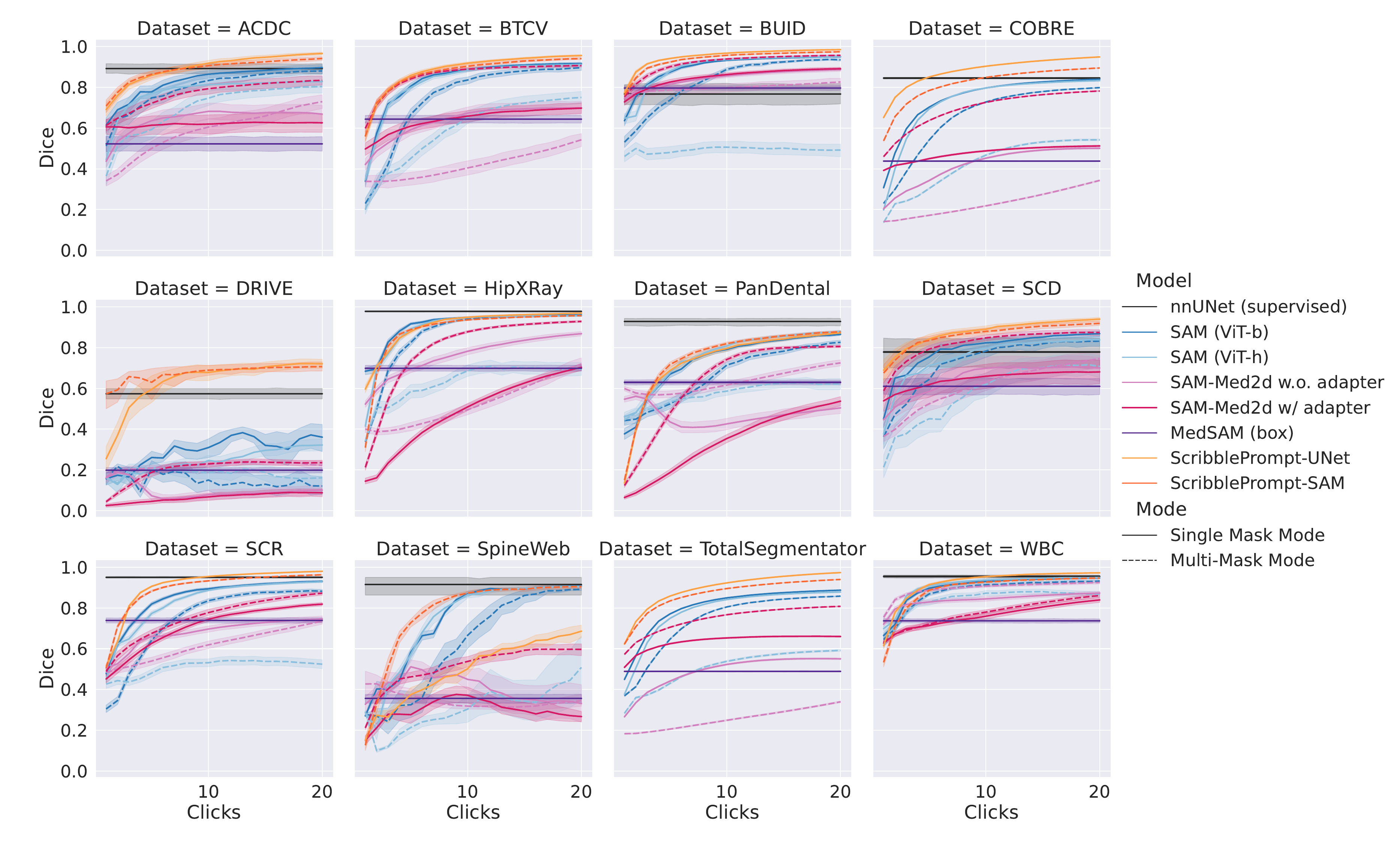}
    \vspace{-0.5\baselineskip}
    \caption{\textbf{Results by dataset with center clicks.} During the first step, one positive click is placed at the center of the largest component of the ground truth segmentation. In subsequent iterations, one (positive or negative) correction click is placed at the center of the largest component of the error region between the previous prediction and ground truth segmentation.}
    \label{fig:center_clicks_per_dataset}
\end{figure}

\begin{figure}
    \centering
    \includegraphics[width=0.9\linewidth]{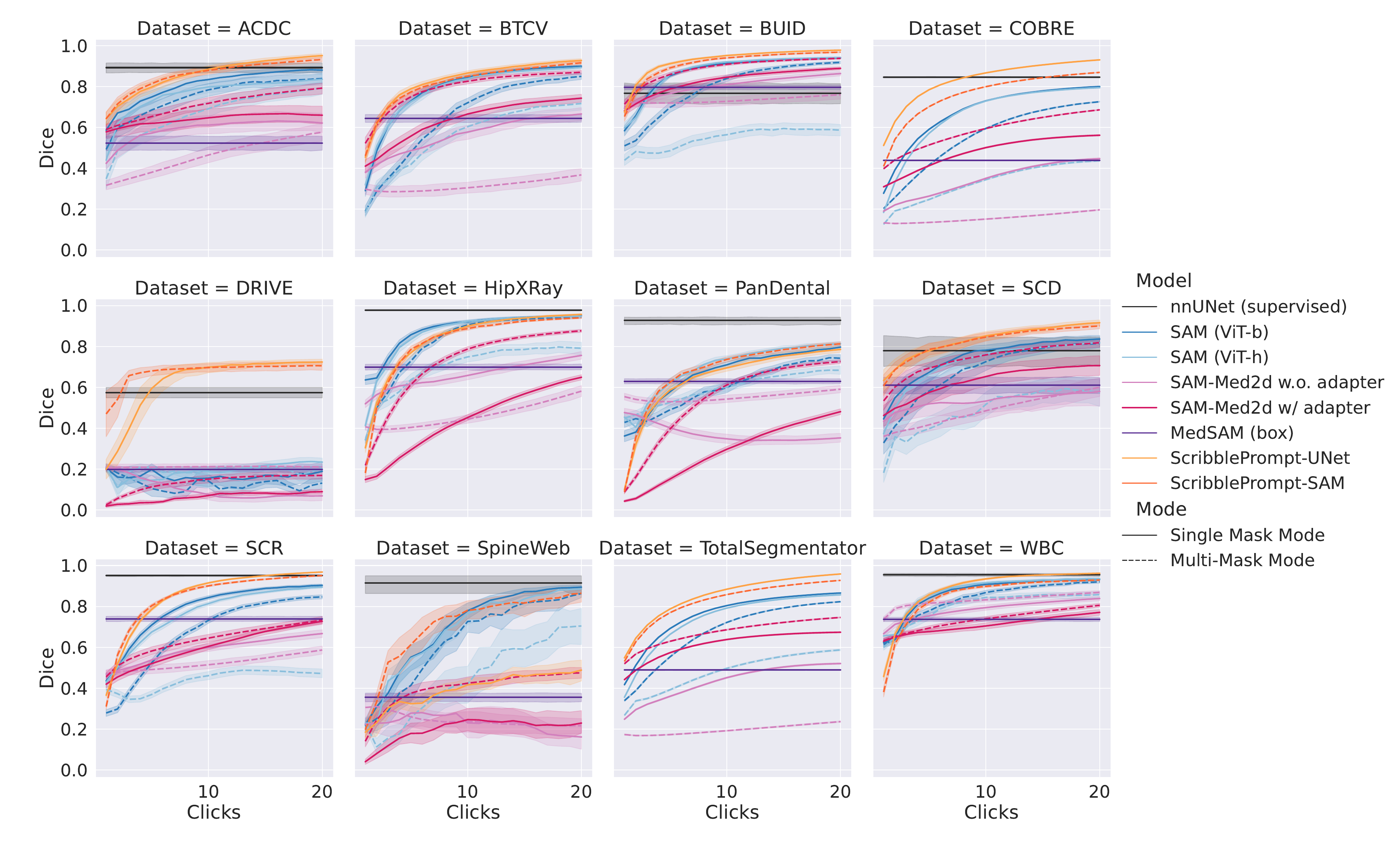}
    \vspace{-0.5\baselineskip}
    \caption{\textbf{Results by dataset with random clicks.} During the first step, one positive click is randomly sampled from the ground truth segmentation. In subsequent steps, one (positive or negative) correction click is randomly sampled from the error region between the previous prediction and ground truth segmentation.}
    \label{fig:random_clicks_per_dataset}
\end{figure}

\begin{figure}
    \centering
    \includegraphics[width=0.9\linewidth]{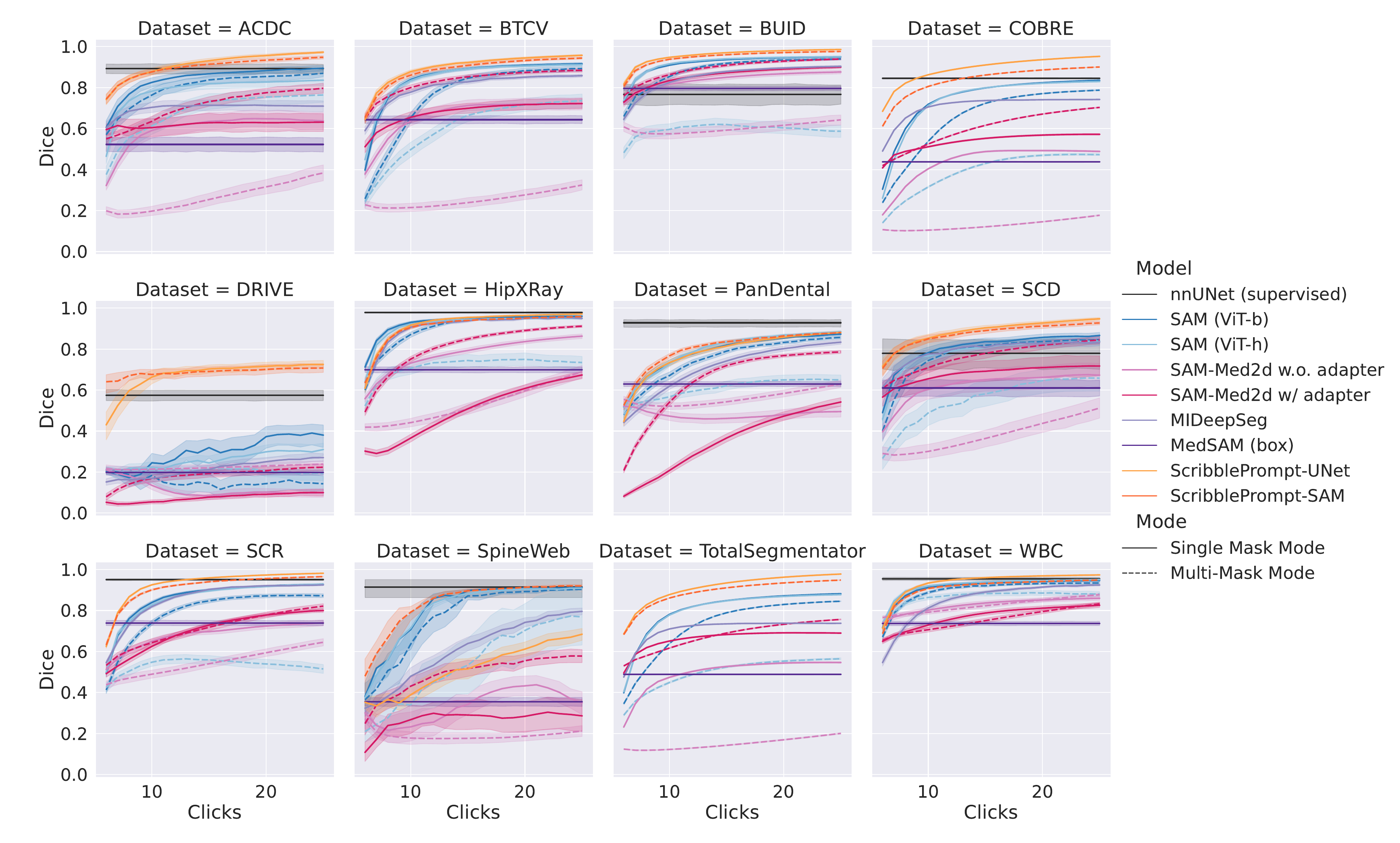}
    \vspace{-0.5\baselineskip}
    \caption{\textbf{Results by dataset with random warm start click procedure.} During the first step, three positive clicks are randomly sampled from the ground truth segmentation and three negative clicks are randomly sampled from the background. In subsequent steps, one (positive or negative) correction click is placed at the center of the largest component of the error region between the previous prediction and ground truth segmentation.}
    \label{fig:warm_start_per_dataset}
\end{figure}

\begin{figure}
    \centering
    \includegraphics[width=0.9\linewidth]{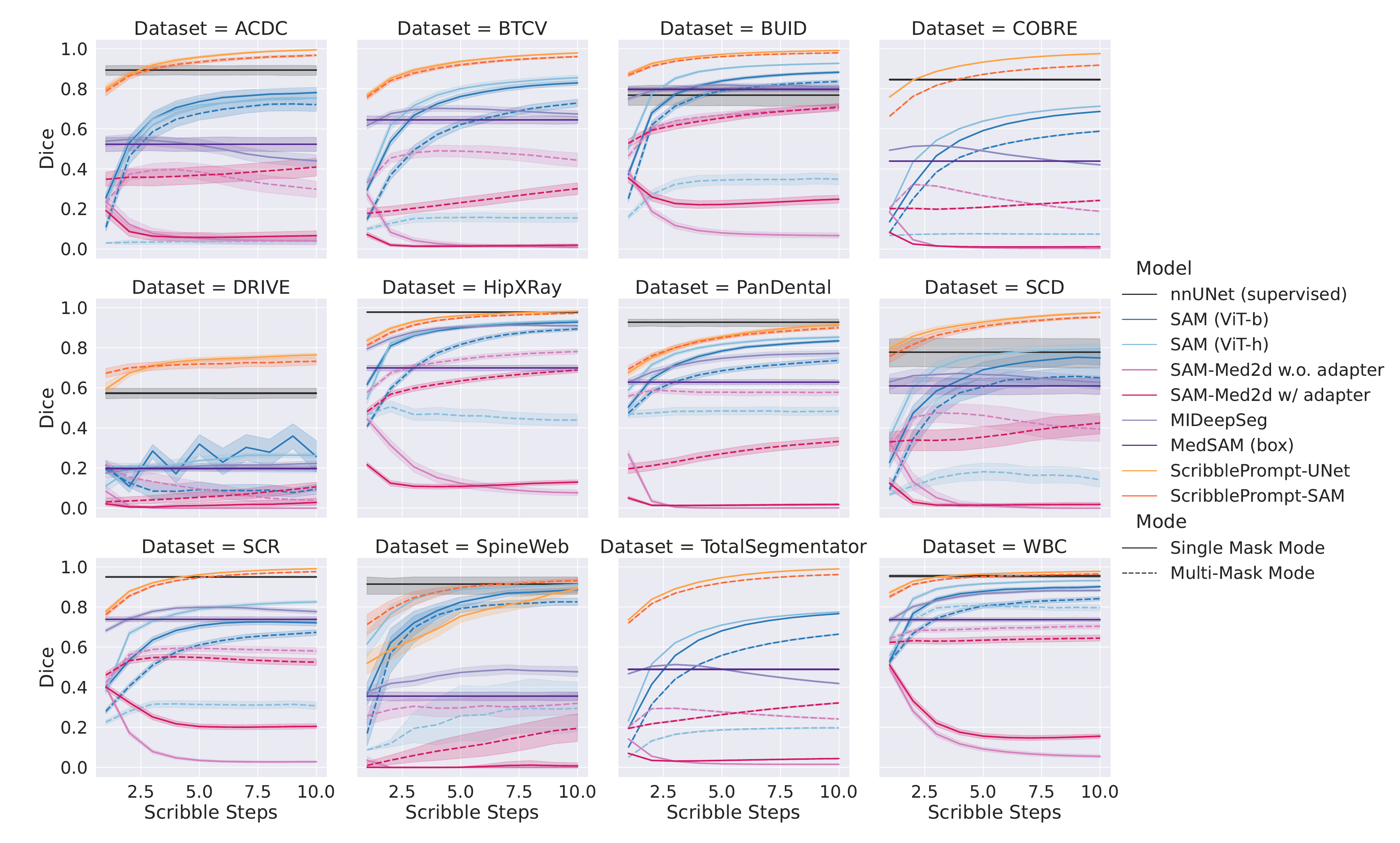}
    \vspace{-0.5\baselineskip}
    \caption{\textbf{Results by dataset with line scribbles.} During the first step we simulate three positive line scribbles and three negative line scribbles. In subsequent steps, we simulate one (positive or negative) correction line scribble based on the error region between the previous prediction and ground truth segmentation. Each line scribble covers a maximum of 128 pixels.}
    \label{fig:line_scribbles_per_dataset}
\end{figure}

\begin{figure}
    \centering
    \includegraphics[width=0.9\linewidth]{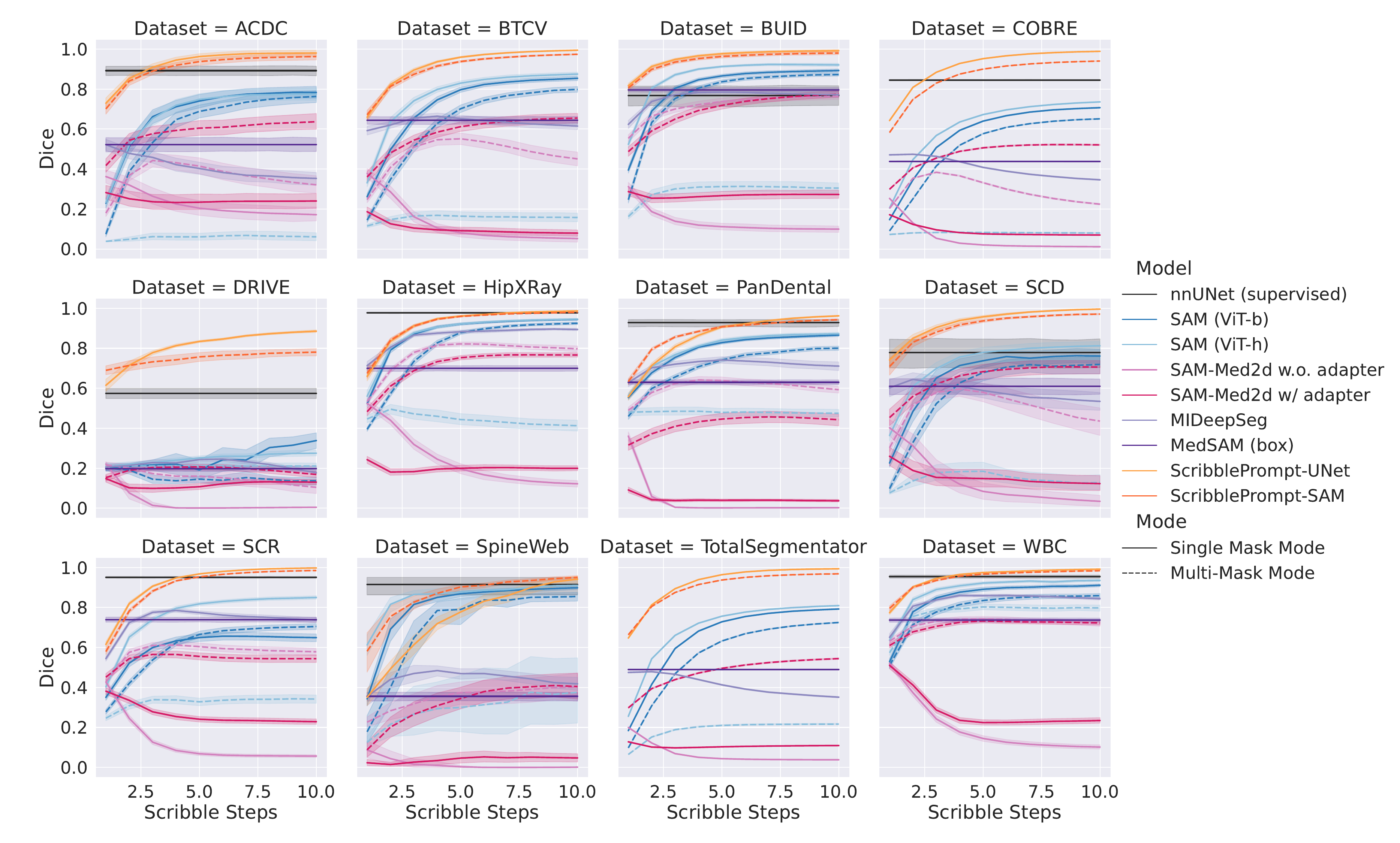}
    \vspace{-0.5\baselineskip}
    \caption{\textbf{Results by dataset with centerline scribbles.} During the first step, we simulate one positive and one negative centerline scribble. In subsequent steps, we simulate one (positive or negative) correction centerline scribbles based on the error region region between the previous prediction and ground truth segmentation. Each centerline scribble covers a maximum of 128 pixels.}
    \label{fig:centerline_scribbles_per_dataset}
\end{figure}

\begin{figure}
    \centering
    \includegraphics[width=0.9\linewidth]{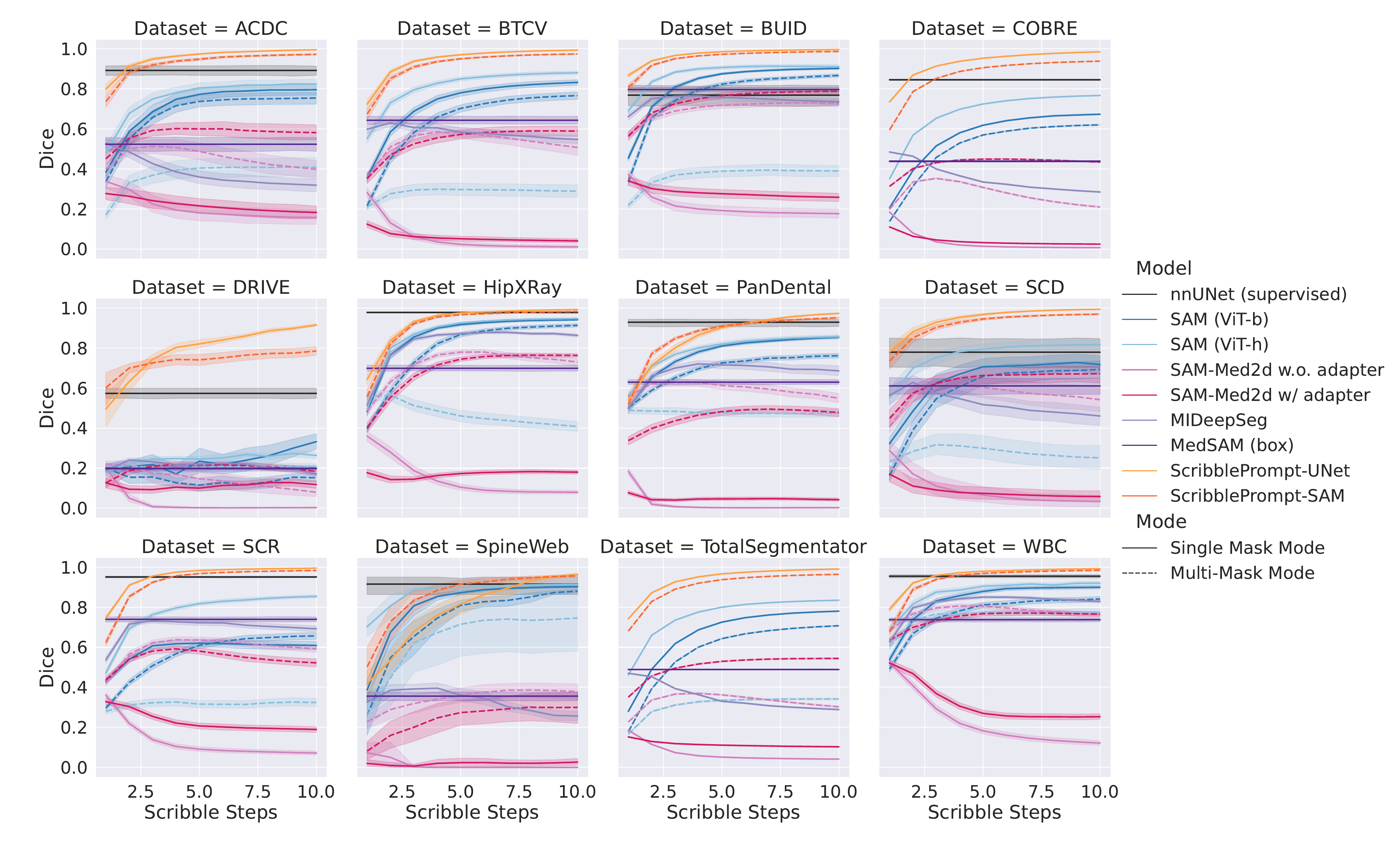}
    \vspace{-0.5\baselineskip}
    \caption{\textbf{Results by dataset with contour scribbles}. During the first step, we simulate one positive and one negative contour scribble based on the ground truth label. In subsequent steps, we simulate one (positive or negative) correction contour scribble based on the error region region between the previous prediction and ground truth segmentation.Each contour scribble covers a maximum of 128 pixels.}
    \label{fig:contour_scribbles_per_dataset}
\end{figure}

\para{Supervised Baselines} We trained fully-supervised task-specific nnUNets~\cite{isensee_nnunet_2021} for 10 of the evaluation datasets. We show the mean Dice of the segmentations predicted by the ensemble of nnUnets using horizontal lines in the results by dataset (\cref{fig:center_clicks_per_dataset}-\ref{fig:contour_scribbles_per_dataset}). 

\para{Results}
\cref{fig:full_simulated_results} shows Dice vs. steps of interaction for three simulated click-focused procedures and three simulated scribble-focused procedures. On average, ScribblePrompt-UNet and ScribblePrompt-SAM have the highest Dice among interactive methods at all steps for all of the simulated interaction procedures. For select interaction procedures we also show HD95 vs. steps of interaction (\cref{fig:full_simulated_results_hd}). ScribblePrompt-UNet and ScribblePrompt-SAM consistently achieve the lowest HD95. 

\para{Results by Dataset}
Figs. \ref{fig:center_clicks_per_dataset}, \ref{fig:random_clicks_per_dataset}, and \ref{fig:warm_start_per_dataset} show quantitative results by dataset for the click-focused interaction procedures. Figs. \ref{fig:line_scribbles_per_dataset}, \ref{fig:centerline_scribbles_per_dataset}, and \ref{fig:contour_scribbles_per_dataset} show quantitative results by dataset for scribble-focused interaction procedures. ScribblePrompt reaches (or surpasses) fully-supervised nnUNet performance for 5 unseen datasets within 1-3 centerline scribbles steps, and for 10 unseen datasets within 6 scribble steps (\cref{fig:centerline_scribbles_per_dataset}). 

\begin{figure}
    \centering
    \includegraphics[width=0.95\linewidth]{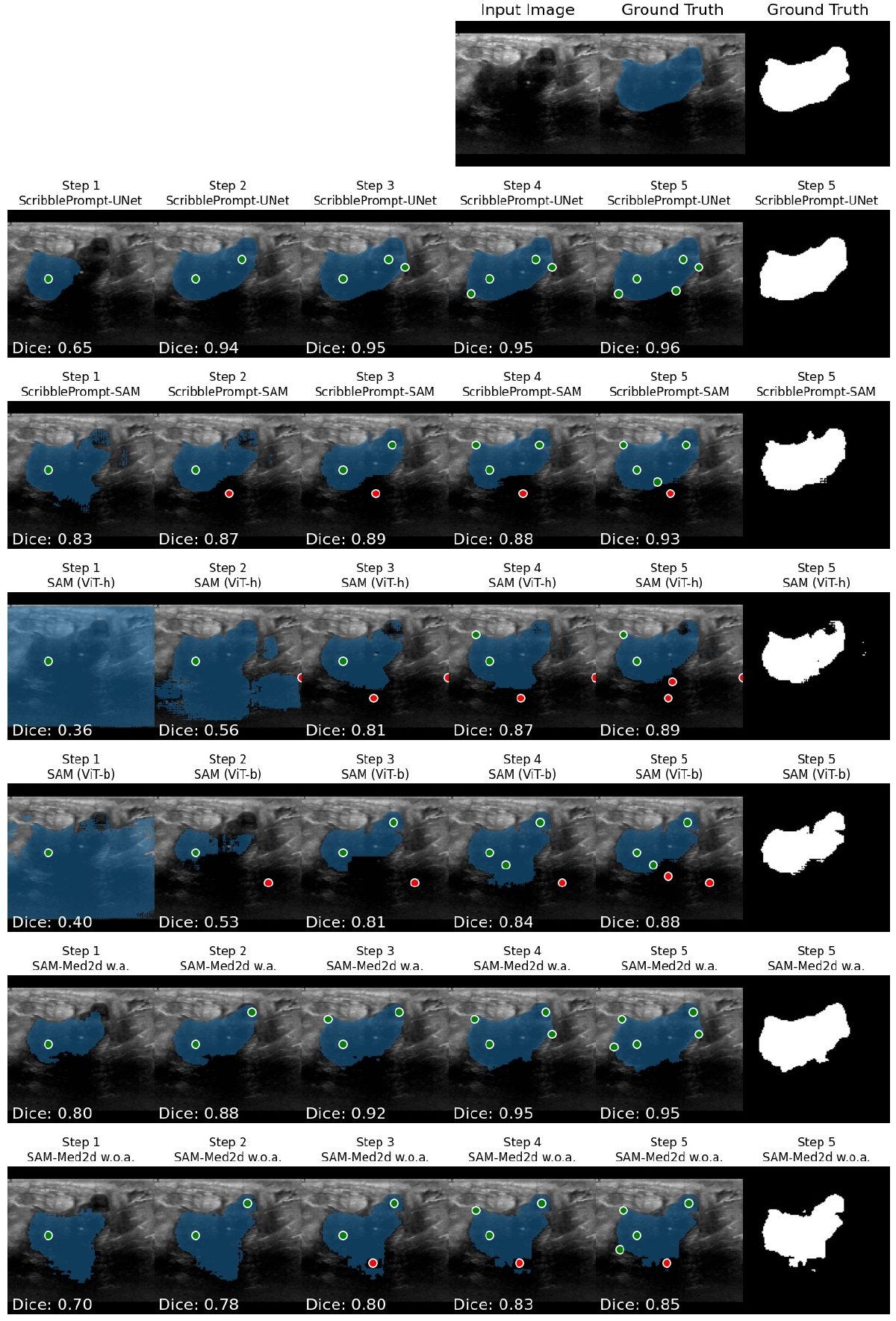}
    \vspace{-0.5\baselineskip}
    \caption{\textbf{Example predictions from center clicks}. We show an example of interactive segmentation of a malignant tumor in an Ultrasound image from the BUID~\cite{BUID} dataset. This dataset was unseen by ScribblePrompt models during training. We simulate an initial click in the center of the label followed by one correction click in the center of the error at each step. }
    \label{fig:example_center_clicks_buid}
\end{figure}

\begin{figure}
    \centering
    \includegraphics[width=0.95\linewidth]{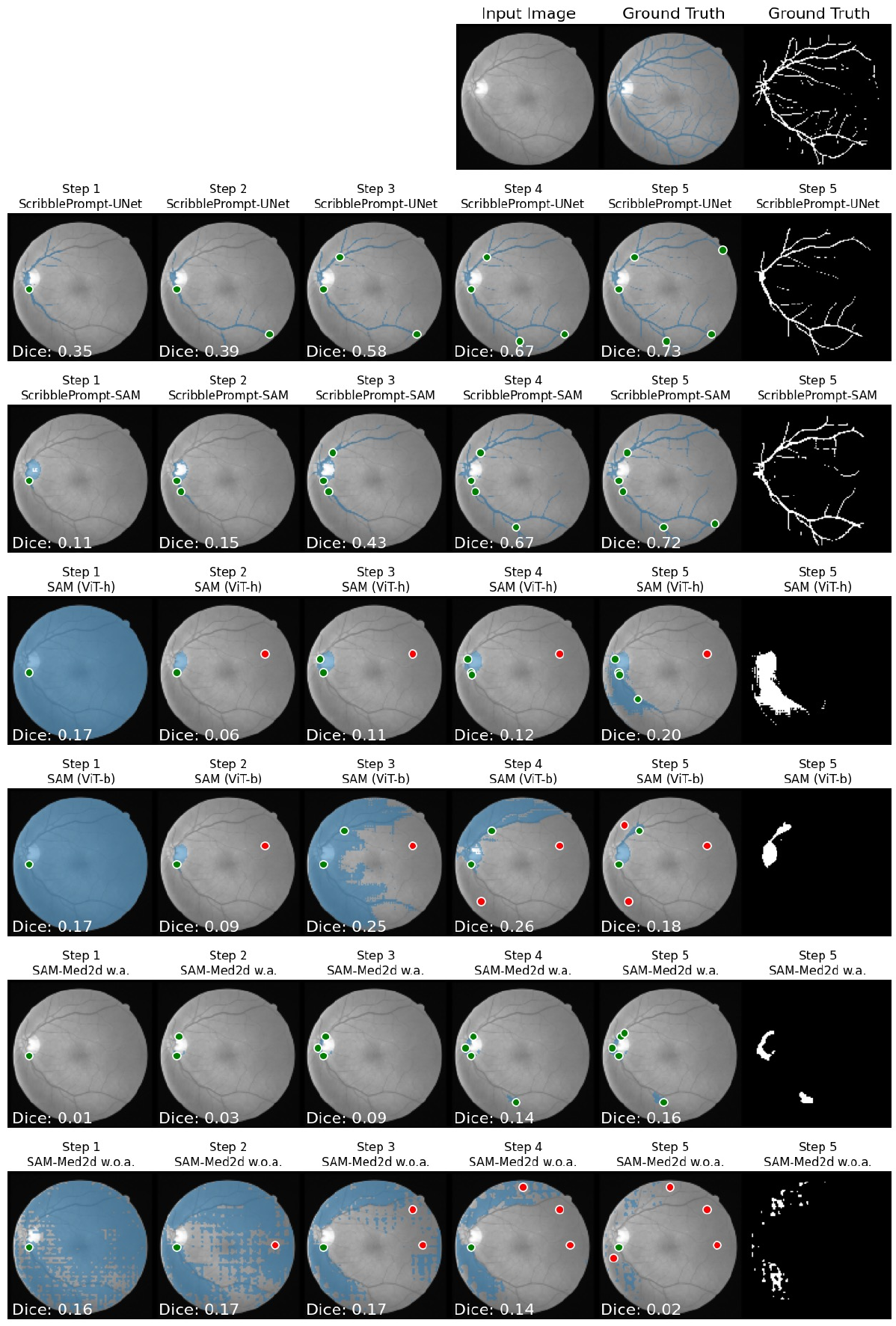}
    \vspace{-0.5\baselineskip}
    \caption{\textbf{Example predictions from center clicks}. We show an example of iterative interactive segmentation of retinal veins in a fundus photograph from the DRIVE dataset~\cite{DRIVE}. This dataset was unseen by ScribblePrompt models during training. The ScribblePrompt models are able to segment the retinal veins while baselines methods are not able to segment these thin structures.}
    \label{fig:example_center_clicks_drive}
\end{figure}

\begin{figure}
    \centering
    \includegraphics[width=0.85\linewidth]{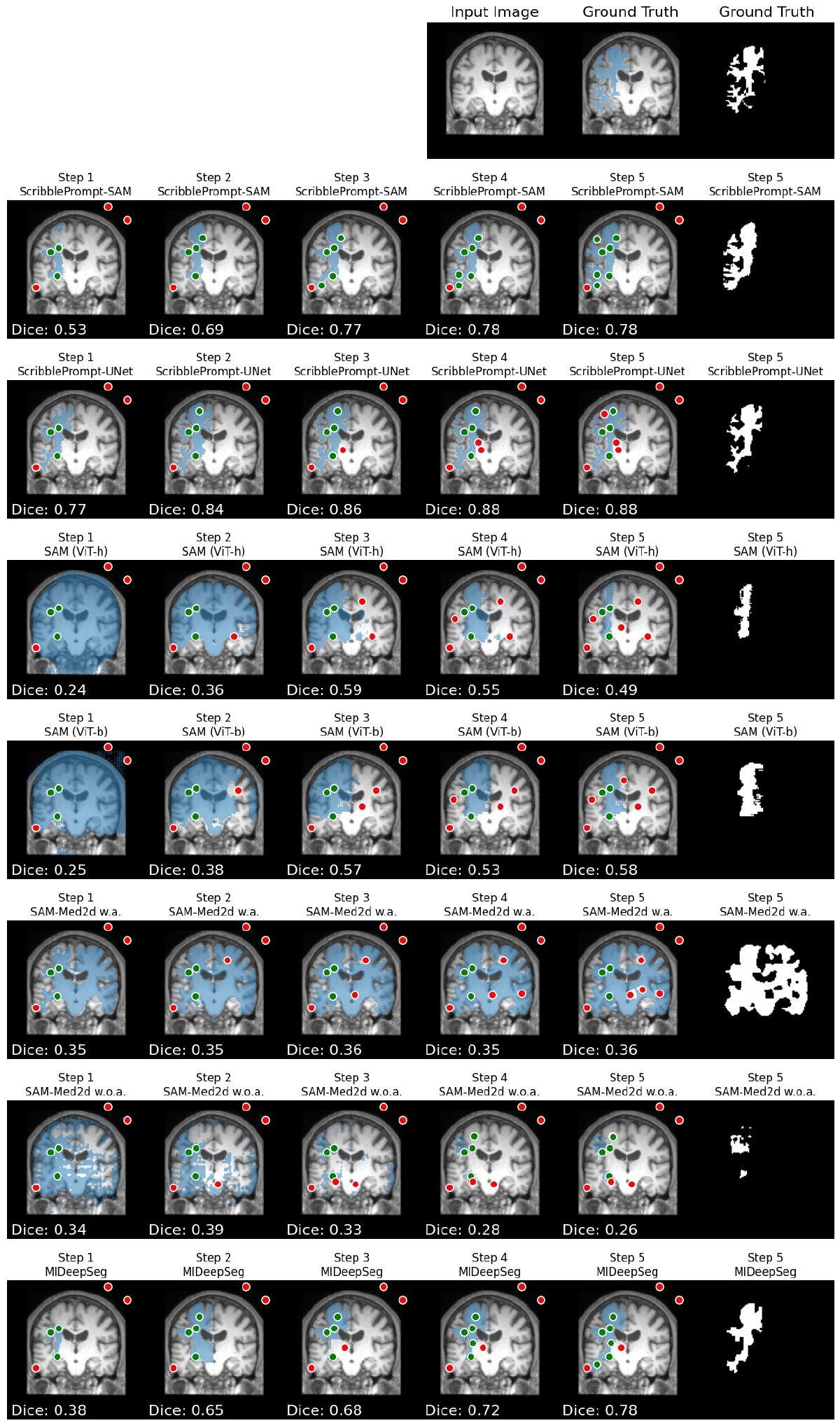}
    \vspace{-0.5\baselineskip}
    \caption{\textbf{Example predictions from random clicks and center correction clicks}. We show an example of white matter segmentation in a T1 brain MRI from the COBRE dataset~\cite{COBRE,fischl2012freesurfer,neurite}. This dataset was completely held-out from ScribblePrompt training and model selection. We simulate interactions following the warm start click protocol: we start with three positive and three negative random clicks, followed by one center correction click per step.}
    \label{fig:warm_start_cobre}
\end{figure}

\para{Visualizations} 
We show predictions for test examples from evaluation datasets unseen by ScribblePrompt during training. \cref{fig:example_center_clicks_buid}, \cref{fig:example_center_clicks_drive}, and \cref{fig:warm_start_cobre} show iterative predictions from each method using clicks. ScribblePrompt is able to segment large ambiguous objects (\cref{fig:example_center_clicks_buid}), as well as thin structures like vasculature (\cref{fig:example_center_clicks_drive}). For large and complex regions of interest such as white matter in brain MRI (\cref{fig:warm_start_cobre}), starting with a few random clicks at once is helpful.

\begin{figure}
    \centering
    \includegraphics[width=0.85\linewidth]{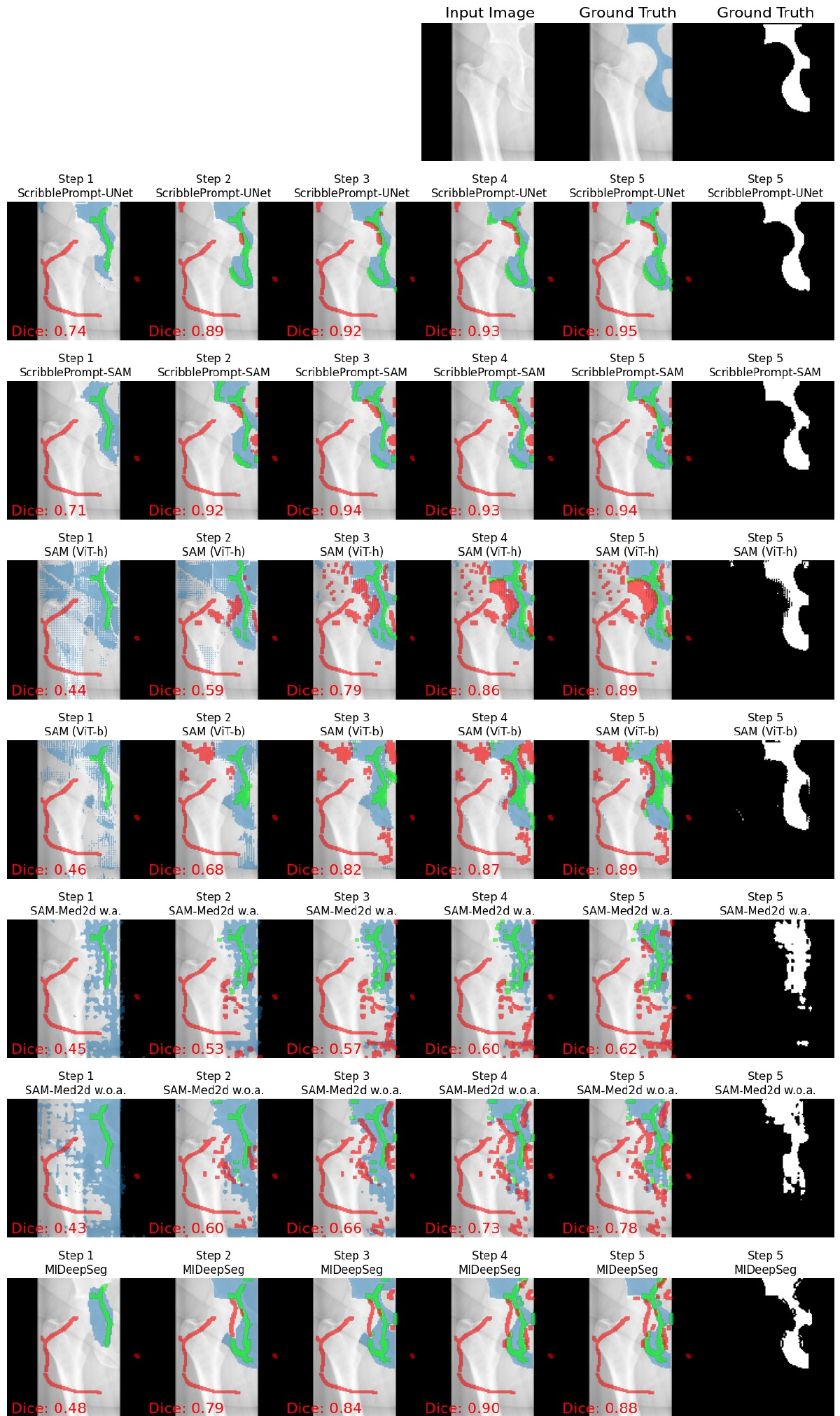}
    \vspace{-0.5\baselineskip}
    \caption{\textbf{Example predictions from centerline scribbles}. We simulate iterative interactive segmentation of the ilium in an X-Ray from the HipXRay dataset~\cite{HipXRay}. This dataset, label, and type of X-Ray was not seen by ScribblePrompt models during training. Correction scribbles were simulated separately for each method based on the error region of the previous prediction. ScribblePrompt models have the highest Dice predictions after 5 scribble steps. Scribble thickness is enlarged for visual clarity.
    }
    \label{fig:centerline_hipxray}
\end{figure}

\begin{figure}
    \centering
    \includegraphics[width=0.8\linewidth]{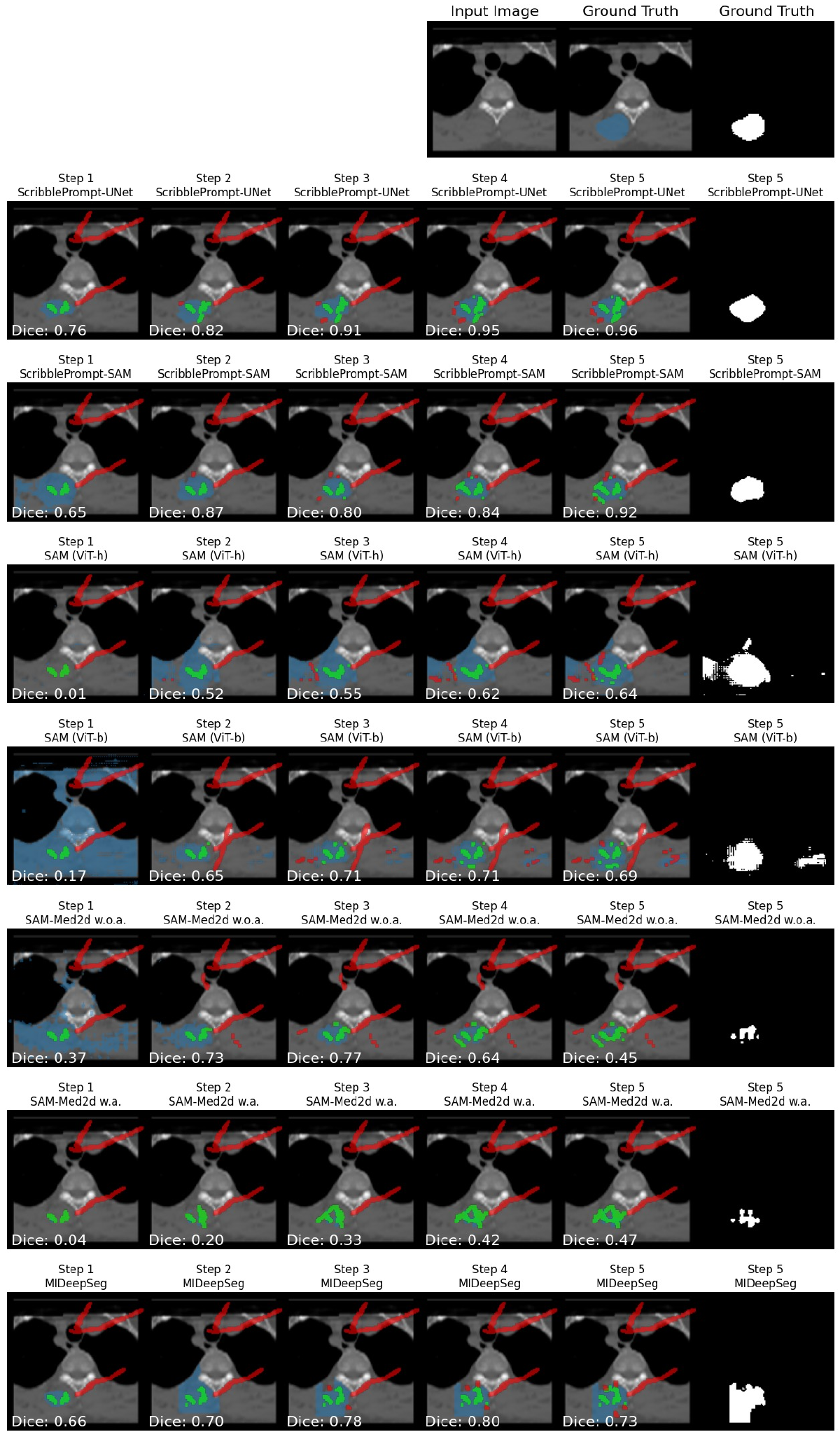}
    \vspace{-0.5\baselineskip}
    \caption{\textbf{Example predictions from line scribbles}. 
    We simulate iterative interactive segmentation of the left autochthon muscle in a CT from the TotalSegmentator dataset~\cite{TotalSegmentator}. This dataset was completely held-out and the label was unseen by ScribblePrompt models during training. This segmentation task is challenging because there is little contrast between the region of interest and surrounding tissue. ScribblePrompt models are able to accurately refine their predictions and a achieve Dice $\geq 0.92$ after 5 scribble steps. Scribble thickness is enlarged for visual clarity.
    }
    \label{fig:line_totalseg}
\end{figure}

\cref{fig:centerline_hipxray} and \cref{fig:line_totalseg}
show iterative interactive segmentation with centerline scribbles and line scribbles. ScribblePrompt is able to accurately segment labels unseen during training using scribbles.  

\clearpage
\section{User Study}
\label{appendix:user_study}

We conducted a user study comparing ScribblePrompt-UNet to SAM (ViT-b). We provide additional details on the user study design and implementation. 

\begin{figure}
    \centering
    \includegraphics[width=0.99\linewidth]{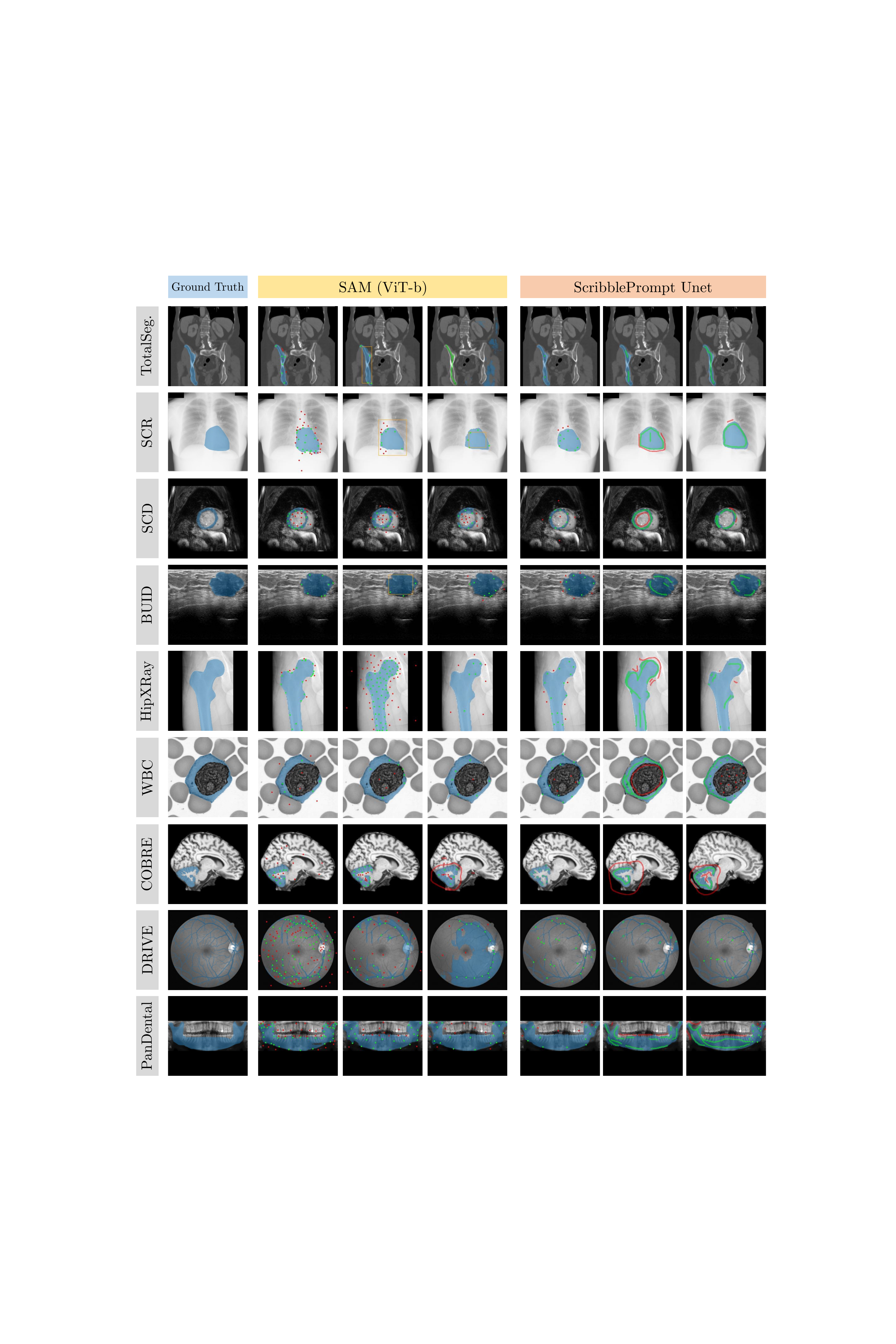}
    \caption{
    \textbf{Example segmentations and interactions from the user study}. 
    We show predictions with interactions provided by three study participants for each of the nine segmentation tasks in the user study. For each example, we visualize positive scribble and click inputs in \textcolor{ForestGreen}{green}, negative scribble and click inputs in \textcolor{red}{red}, bounding box inputs in \textcolor{Dandelion}{yellow}, and the predicted segmentation in \textcolor{eccvblue}{blue}. With SAM, study participants primarily used clicks. With ScribblePrompt, participants used a mix of scribbles and clicks. For the retinal vein segmentation task, participants preferred to use clicks with both models. Participants prompted SAM with denser clicks compared to ScribblePrompt. 
    }
    \label{fig:user_study_examples}
\end{figure}

\para{Study Design}
The goal of the user study was to compare ScribblePrompt to the best click-focused baseline method in terms of accuracy (Dice of the final segmentations), efficiency (time to achieve the desired segmentations) and user experience (perceived effort). 
Participants were given time to familiarize themselves with both models on a fixed set of practice images. Afterwards they used each model to segment a series of nine new test images from nine tasks that were not seen by the model during training (\cref{fig:user_study_examples}). 

The order in which participants used the models, and which image the users were assigned to segment with each model for each task was randomized. We randomly selected one training image per task to include in the set of practice images. We randomly selected two test images per task and randomized the assignment of each image to each model for each participants. Each participant segmented a total of 18 images during the study. The models were also annonymized (i.e., ``Model A'' and ``Model B''). We informed participants that one model was designed to be used with clicks and bounding boxes, while the other was designed for use with clicks, bounding boxes, and scribbles. 

For each segmentation task, the participants were shown the target segmentation and were asked to interact with the model until the predicted segmentation closely matched the target or they could no longer improve the prediction. We provided participants with the target segmentation to disentangle the cognitive process of identifying the region of interest from prompting the model to achieve the desired segmentation. 

\para{Study Participants}
Study participants were neuroimaging researchers at an academic hospital. Although the participants had prior experience with medical image segmentation, they did not necessarily have experience with the specific tasks and types of images used in the study.

We had a total of 29 participants with 16 participants completing all of the segmentations and the exit survey. We observed a higher attrition rate among participants who were assigned to use SAM first, even after being able to freely try out both models during the ``practice'' phase. Among the 13 participants assigned to use SAM first, 62\% did not finish all of their segmentations, compared with 31\% among the 16 assigned to use ScribblePrompt first. We report results on the 16 participants who completed all the segmentations and the exit survey. 

\para{Implementation}
Each participant used a web-based interface powered by a Nvidia Quatro RTX8000 GPU with 4 CPUs. Participants segmented the images at $256 \times 256$ resolution. The interface was developed in Python using the Gradio library~\cite{gradio}. The interface had a ``practice'' mode in which users could freely switch between the two models and images from the set of practice images. After experimenting with both models, users clicked a button to begin ``recorded activity'' mode in which users were led through performing specific segmentation tasks with specific models. Users provided positive/negative scribble inputs, positive/negative click inputs and/or bounding box inputs, and then clicked a button to receive a prediction from the model. 

\para{Survey Results}
Common factors that influenced participants preference for ScribblePrompt was being able to get accurate predictions from scribbles (``[ScribblePrompt] was more spatially smooth''), the model's responsiveness to a variety of inputs (``it landed on my desired predictions more easily''), and less perceived effort when using the model (``[ScribblePrompt] needed much less guidance''). Participants preferred using clicks and bounding boxes over scribbles with SAM, praising its ``snapiness'', the effectiveness of ``exclusion clicks'' and remarking it worked well for ``rigid structures''. However, participants also noted in some cases ``[SAM] did not respect object boundaries'', and for tasks such as retinal vein segmentation ``[SAM] required lots of clicks and still was not very accurate''.

\para{Visualizations}
We visualize some of the interactions used by study participants and the resulting predictions in \cref{fig:user_study_examples}. Study participants used denser clicks when prompting SAM compared to when prompting ScribblePrompt-UNet.

\clearpage
\section{Inference Runtime}
\label{appendix:gpu_runtime}

\para{Setup}
We measure inference time for a random input with a scribble covering 128 pixels. We report mean and standard deviation of inference time across 1,000 runs on a single CPU and on a Nvidia Quatro RTX8000 GPU.

\para{Results}
We show performance results in \cref{tab:runtime_cpu_gpu}. On a single CPU, \linebreak\mbox{ScribblePrompt-UNet} requires $0.27 \pm 0.04$ sec per prediction, enabling the model to be used even in low-resource environments. Prior work on interactive interfaces indicates that $<0.5$ sec latency is sufficient for cognitive tasks~\cite{liu2014effects}. ScribblePrompt-UNet is also faster than the baseline methods on a GPU. 

Its efficient fully-convolutional architecture gives ScribblePrompt-UNet low latency inference. With SAM, latency scales with the number of interactions because each point is encoded as a 256-dimensional vector embedding. For ScribblePrompt-UNet, clicks and scribbles are encoded in masks, so inference time (per prediction) is constant with the number of interactions. 

\begin{table}
    \centering
    \caption{\textbf{Performance Summary.} 
    We measure inference time separately on a single CPU and on an Nvidia Quatro RTX8000 GPU for a prediction with a random scribble input covering 128 pixels. We report mean and standard deviation across 1,000 runs. ScribblePrompt-SAM and MedSAM use the same architecture as SAM ViT-b. \firstone{Best} and \secondone{second best} are highlighted. 
    }
    \label{tab:runtime_cpu_gpu}
    \rowcolors{2}{white}{gray!15}
    \resizebox{0.99\textwidth}{!}{
    \setlength{\tabcolsep}{0.5em}
    \begin{tabular}{l|c|c|c|c}
    \toprule
    Architecture &  Param. & CPU Runtime (sec) & GPU Runtime (ms) & GPU Memory 
    \\
    \midrule
    SAM (ViT-h) & 641M & $130.79 \pm 7.96$ & $504.36 \pm 57.72$ & $21.912$ GB
    \\
    SAM (ViT-b) & 94M & $13.59 \pm 0.77$ & $133.85 \pm 24.26$ & $7.144$ GB
    \\
    SAM-Med2D w/ adapter & 271M & $1.23 \pm 0.07$ & $35.06 \pm 12.88$ & $1.489$ GB
    \\
    SAM-Med2D w.o. adapter & 91M & $0.63 \pm 0.02$ & \secondone{$24.86 \pm 9.56$} & $734$ MB
    \\    
    MIDeepSeg & 3M & \firstone{$0.08 \pm 0.02$} & $65.75 \pm 21.87$ & $11$ MB
    \\
    \midrule
    ScribblePrompt-UNet & 4M & \secondone{$0.27 \pm 0.04$} & \firstone{$1.96 \pm 0.20$} & $125$ MB \\
    \bottomrule
    \end{tabular}
    }
\end{table}

\clearpage
\section{Ablations}
\label{appendix:ablations}

We conduct two ablations of important ScribblePrompt design decisions: (1) synthetic label inputs used during training, and (2) types of prompts simulated during training. We report results on the validation splits of nine validation datasets that were unseen during training. 

\subsection{Synthetic Labels}
\label{appendix:ablation_synth}

\para{Setup}
We trained ScribblePrompt-UNet and ScribblePrompt-SAM with different values of $p_{synth}$, the probability of sampling a synthetic label. 

\begin{figure}
    \centering
    \includegraphics[width=\linewidth]{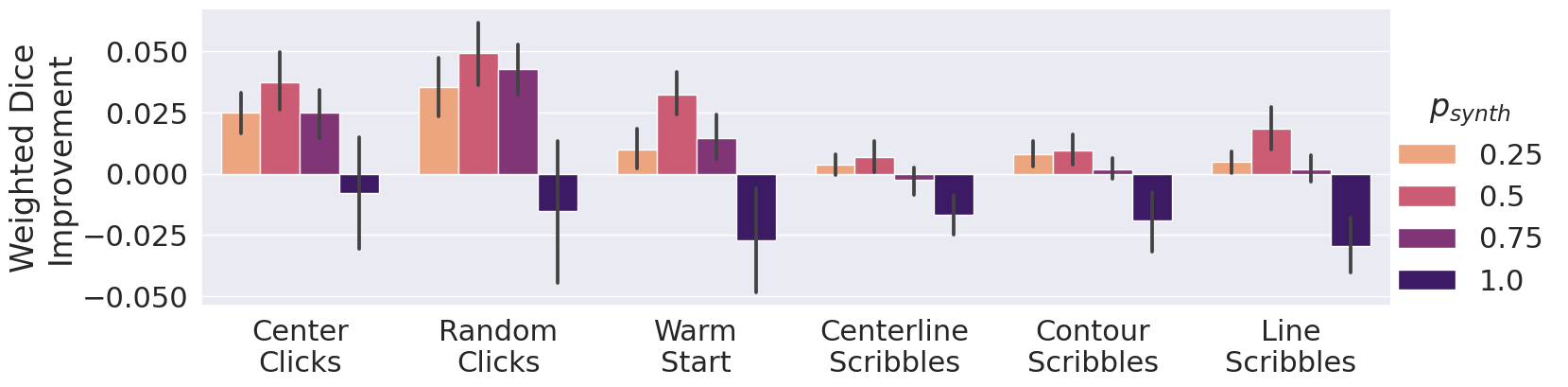}
    \caption{\textbf{Probability of synthetic labels during training for ScribblePrompt-UNet.} We report change in Dice relative to ScribblePrompt-UNet trained without any synthetic labels ($p_{synth}=0$). We show Dice after five steps of simulated interactions following six different (inference-time) interaction procedures. Errorbars show 95\% CI. }
    \label{fig:appendix_superpixel_ablation_unet}
\end{figure}

\begin{figure}
    \centering
    \includegraphics[width=0.9\linewidth]{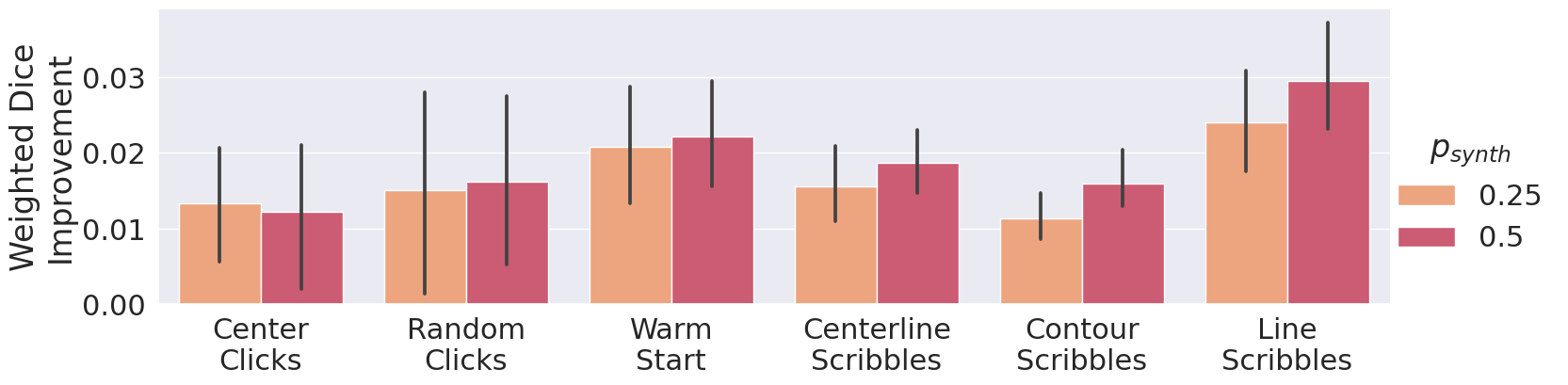}
    \caption{\textbf{Probability of synthetic labels during training for ScribblePrompt-SAM.} We report change in Dice relative to ScribblePrompt-SAM trained without any synthetic labels ($p_{synth}=0$). We show Dice after five steps of simulated interactions following six different (inference-time) interaction procedures. Errorbars show 95\% CI. }
    \label{fig:appendix_superpixel_ablation_sam}
\end{figure}

\para{Results}
Training with some synthetic labels improves both ScribblePrompt-UNet and ScribblePrompt-SAM's performance on validation data from nine (validation) datasets not seen during training (\cref{fig:appendix_superpixel_ablation_unet}, \ref{fig:appendix_superpixel_ablation_sam}). For both ScribblePrompt-UNet and ScribblePrompt-SAM, training with 50\% synthetic labels leads to the highest Dice on unseen datasets at inference time. 

\subsection{Prompt Types}
\label{appendix:ablation_prompt}

\para{Setup}
We evaluate ScribblePrompt-UNet models trained with different combinations of prompts, compared to the complete ScribblePrompt-UNet: 
\begin{itemize}
    \item \textbf{ScribblePrompt-UNet~(scribbles)} trained on boxes and scribbles.
    \item \textbf{ScribblePrompt-UNet~(clicks)} trained on boxes and clicks.
    \item \textbf{ScribblePrompt-UNet~(random clicks)} trained on boxes and random clicks. 
\end{itemize}

\begin{figure}
    \centering
    \includegraphics[width=\linewidth]{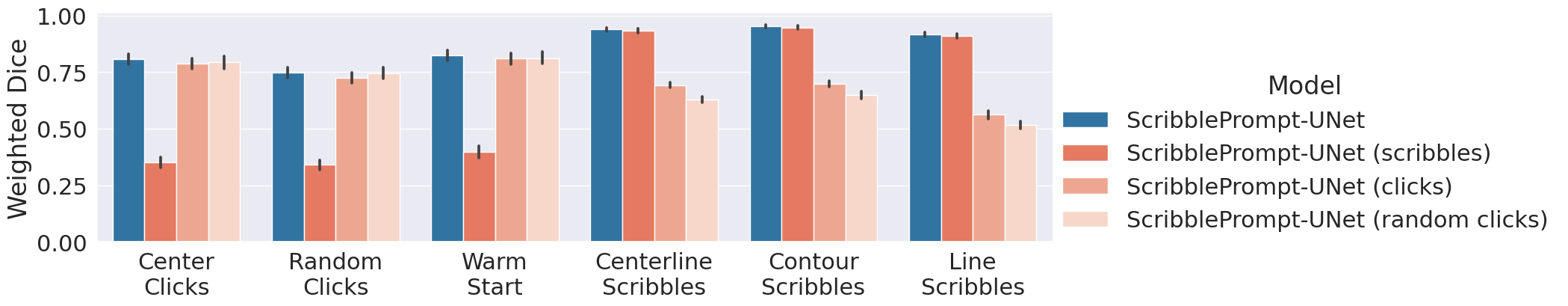}
    \caption{\textbf{Ablation of interactions during training}. We report Dice after five steps of simulated interactions following six inference-time interaction procedures. Error bars show 95\% CI from bootstrapping.}
    \label{fig:full_prompt_ablation}
\end{figure}

\para{Results}
\cref{fig:full_prompt_ablation} shows results for six different inference-time interaction procedures. 
ScribblePrompt-UNet trained with scribbles, clicks, and bounding boxes predicts segmentations more accurately than do ablated versions of ScribblePrompt-UNet. 

\end{document}